\documentclass{article}

\usepackage{PRIMEarxiv}

\usepackage[utf8]{inputenc} % allow utf-8 input
\usepackage[T1]{fontenc}    % use 8-bit T1 fonts
\usepackage{hyperref}       % hyperlinks
\usepackage{url}            % simple URL typesetting
\usepackage{booktabs}       % professional-quality tables
\usepackage{amsfonts}       % blackboard math symbols
\usepackage{nicefrac}       % compact symbols for 1/2, etc.
\usepackage{microtype}      % microtypography
\usepackage{lipsum}
\usepackage{fancyhdr}       % header
\usepackage{graphicx}       % graphics

\usepackage{kotex}      
\usepackage{natbib}
\usepackage{multirow}
\usepackage{array}     % 열 정렬 조금 더 유연하게
\usepackage{makecell} 
\usepackage{adjustbox}
\usepackage{tabularx}
\usepackage[table]{xcolor}
\usepackage{colortbl}
\usepackage{pgf}
\usepackage{tikz}        % 또는 \usepackage{pgfplots}
\usepackage{pgfplots}
\pgfplotsset{compat=1.18}  % pgfplots 쓰면 이 줄도 같이
\usepackage{subcaption}
\usepackage{tikz}
\usetikzlibrary{positioning,arrows.meta,shapes.misc,fit}
\usepackage{amssymb}   % \checkmark
\usepackage{pifont}    % \xmark
\usepackage{wrapfig}
\usepackage{amsmath}
\usepackage{float}   % preamble에 추가
\usepackage{}
\usepackage{placeins}  % 프리앰블에 추가

\usepackage{natbib}
\graphicspath{{media/}}     % organize your images and other figures under media/ folder

\title{\textbf{Evaluating Large Language Models on the 2026 Korean CSAT Mathematics Exam: \\ Measuring Mathematical Ability in a Zero–Data-Leakage Setting}}

\author{
  Goun Pyeon\thanks{These authors contributed equally.}, 
  Inbum Heo\footnotemark[1], 
  Jeesu Jung\footnotemark[1], 
  Taewook Hwang\footnotemark[1],
  Hyuk Namgoong, \\
  \textbf{Hyein Seo,
  Yerim Han,
  Eunbin Kim,
  Hyeonseok Kang,
  Sangkeun Jung\thanks{Corresponding author.}} \\
  Department of Computer Science \& Engineering \\
  Chungnam National University \\
  \texttt{\{vusrhdns714, inbum10222, jisu.jung5, taewook5295, hyuk199, }\\ \texttt{hyenee97, namu.rim2, eunbinkim777, dnfldjaak11, hugmanskj\}@gmail.com}
}

\begin{document}
\maketitle

% \tableofcontents
% \newpage

\begin{abstract}

This study systematically evaluated the mathematical reasoning capabilities of Large Language Models (LLMs) using the 2026 Korean College Scholastic Ability Test (CSAT) Mathematics section, ensuring a completely \textbf{contamination-free} evaluation environment. To address data leakage issues in existing benchmarks, we digitized all 46 questions (22 common and 24 elective) within \textbf{two hours} of the exam's public release, eliminating any possibility of inclusion in model training data. We conducted comprehensive evaluations of 24 state-of-the-art LLMs across varying input modalities (Text-only, Image-only, Text+Figure) and prompt languages (Korean, English).
The GPT-5 family models achieved perfect scores (100 points) under a limited set of language–modality configurations, while Grok 4, Qwen 3 235B, and Gemini 2.5 pro also scored above 97 points. Notably, gpt-oss-20B achieved 95.7 points despite its relatively small size, demonstrating high cost-effectiveness. Problem-specific analysis revealed Calculus as the weakest domain with significant performance degradation on 4-point high-difficulty problems. Text input consistently outperformed image input, while prompt language effects varied by model scale.
In reasoning enhancement experiments with GPT-5 series, increased reasoning intensity improved performance (82.6→100 points) but quadrupled token usage and drastically reduced efficiency, suggesting that models with minimal reasoning may be more practical. This research contributes: (1) implementation of a completely unexposed evaluation environment, (2) a standardized digitization pipeline that converts human-targeted exam materials into LLM-ready evaluation data, and (3) a practical evaluation perspective integrating performance, cost, and time considerations. Detailed results and model comparisons are available at the 2026 Korean CSAT LLM Evaluation Leaderboard; \url{https://isoft.cnu.ac.kr/csat2026/}.

\end{abstract}

% keywords can be removed

\section{Introduction}
\label{sec:intro}

The mathematical reasoning abilities of Large Language Models (LLMs) have become a central indicator for assessing a model’s logical thinking and problem-solving skills. Because mathematical problems demand clear answers and structured reasoning, they are widely used as benchmarks for evaluating a model’s true reasoning capabilities. However, with the rapid performance gains of recent models, traditional evaluation approaches have increasingly struggled to accurately distinguish an LLM’s genuine mathematical competence.

In particular, given that LLMs are trained on vast web-scale corpora, it is difficult to verify whether a model has been exposed—directly or indirectly—to publicly available benchmark datasets or closely related materials~\citep{min-etal-2022-rethinking, dodge-etal-2021-documenting}. Even when a benchmark itself is not included in training data, paraphrased, synthetic, or structurally transformed versions of benchmark problems may still have been used in training or inference.

Such \textbf{potential contamination} to benchmark data can distort measured performance, making it difficult to determine whether a model solves problems using generalized mathematical reasoning or merely reproduces patterns memorized during training. In other words, an LLM may perform well on a public benchmark simply because it has been implicitly optimized for those questions. Recent studies suggest that benchmark contamination may have occurred in the training data of major LLMs such as GPT-4 and Claude, raising the possibility that high scores may reflect \emph{memorization or pattern replication} rather than true reasoning ability~\citep{balloccu2024leak, deng2024investigating}. Thus, an evaluation dataset that is completely free from any risk of data leakage is essential for validating an LLM’s genuine mathematical understanding and problem-solving ability.

In this context, the Mathematics section of the Korean College Scholastic Ability Test (CSAT) provides a uniquely leakage-free evaluation environment. The CSAT is a national-level exam created annually by the government, and because test items remain undisclosed until the exam ends, rapid digitization and evaluation immediately after release ensures that no LLM could have incorporated similar problems into its training data. Moreover, with 46 questions covering core domains such as \textit{Probability and Statistics}, \textit{Calculus}, and \textit{Geometry}, the CSAT Mathematics exam enables comprehensive measurement of an LLM’s practical mathematical proficiency across the entire secondary curriculum.

This study establishes a new evaluation framework that systematically assesses the mathematical reasoning capabilities of state-of-the-art LLMs under a fully contamination-free setting, using the Mathematics section of the 2026 CSAT. By digitizing and preprocessing all questions within two hours of their public release, we constructed a \emph{Contamination-Free CSAT benchmark} that could not appear in the training data of any model. Beyond simple accuracy comparison, our framework varies input modality, prompting language, reasoning configuration, and solving time, enabling a multidimensional analysis of how LLMs interpret and solve real exam problems.

This work focuses on the following four research questions (RQs):

\begin{itemize}
\item \textbf{RQ1. What are the mathematical problem-solving abilities of LLMs on the 2026 CSAT?}

\item \textbf{RQ2. Which types of CSAT Mathematics problems are most challenging for LLMs?}

\item \textbf{RQ3. How does LLM mathematical performance vary across different input conditions?}

\item \textbf{RQ4. What impact does varying the strength of reasoning have on model performance?}
\end{itemize}

% Our results show that although the GPT-5 family achieved scores above 90 points—with GPT-5 Codex being the only model to obtain a perfect score—many models displayed notably poor performance on Calculus questions and high-difficulty 4-point problems, especially short-answer problems.
Our results show that although several GPT-5 models achieved perfect scores under specific language–modality combinations, the GPT-5 family was the only group to reach such performance levels. In contrast, many other models exhibited notably poor performance on Calculus questions and on high-difficulty 4-point items, particularly short-answer problems.

Text-only input consistently outperformed image-based input, and increasing the reasoning strength of GPT-5 improved accuracy from 82.6 to 100 points, but with a 4–5× increase in token usage, sharply reducing efficiency. Notably, the small-scale gpt-oss-20B achieved 95.7 points at the lowest cost (\$0.01), suggesting that model size is not a determining factor in mathematical performance.

The structure of this paper is as follows.
Section~\ref{sec:2_realated_work} reviews prior work on math benchmarks and LLM evaluation.
Section~\ref{sec:3_framework} analyzes the CSAT problem structure and introduces our CSAT-Math evaluation framework, including the data construction pipeline and prompt design.
Section~\ref{sec:4_experiment_setup} describes the evaluated models and experimental setup.
Section~\ref{sec:5_experiment} presents experiments and results corresponding to the four research questions (RQ1–RQ4).
Finally, Section~\ref{sec:6_discussion} discusses key findings, Section~\ref{sec:7_limitation} outlines the main limitations, and Section~\ref{sec:8_conclusion} concludes with future research directions.
\section{Related Work}
\label{sec:2_realated_work}

\subsection{Math Benchmarks for LLMs}

A wide range of benchmarks has recently been proposed to evaluate the mathematical reasoning abilities of large language models (LLMs). These benchmarks can be broadly categorized into \textbf{(1) benchmarks specifically constructed for LLM performance evaluation} and \textbf{(2) benchmarks based on real standardized exam problems}.

The first category—LLM-specific math benchmarks—consists of datasets designed to precisely measure a model’s computational and reasoning abilities. GSM8K~\citep{cobbe2021gsm8k} evaluates basic numerical reasoning and short chain-of-thought (CoT) capabilities using grade-school–level word problems. CHAMP~\citep{mao2024champ} includes hints and intermediate reasoning steps to analyze how much a model depends on external information, while GSM-Plus~\citep{li2024gsmplus} introduces paraphrasing and numerical perturbation to test robustness under distribution shifts.  
More recently, multimodal benchmarks such as MathVista~\citep{lu2024mathvista} and MATH-Vision~\citep{wang2024mathvision} incorporate visual information—figures, diagrams, and mathematical notation—to evaluate vision–language integrated reasoning.

The second category evaluates LLMs using real exam problems. International standardized tests such as the SAT, GRE, AMC, AIME, and IMO are commonly used, and a representative math benchmark\footnote{\url{https://artificialanalysis.ai/evaluations/aime-2025}} for LLMs relies on AIME problems. Benchmarks such as MATH~\citep{hendrycks2021math}, MathBench~\citep{liu2024mathbench}, OlymMATH~\citep{sun2025challenging}, and U-MATH~\citep{chernyshev2024u} include problems drawn from real examinations and assess conceptual understanding and logical structuring.  
While these benchmark suites expand beyond simple accuracy—incorporating reasoning coherence and explanation quality—they remain predominantly focused on English-language, Western exam contexts.

In contrast, our work systematically evaluates LLM math performance in a non-English educational setting using the Korean CSAT Mathematics exam. The CSAT is a nationwide high-stakes university entrance exam taken annually by over 500,000 students and is regarded as a \textbf{national standard exam with high validity, reliability, and carefully balanced difficulty}.  
By using all 46 questions—22 from the common curriculum and 24 from elective subjects—we conduct a \textbf{comprehensive evaluation} covering geometry, calculus, and probability and statistics.

\subsection{Data Leakage in LLM Evaluation}

\textbf{Data leakage} poses a critical threat to the reliability of LLM performance evaluation and has been a central topic of discussion in math and language benchmark research.

\citet{xu2024benchmarking} showed that major math benchmarks such as GSM8K and MATH appear in the training data of modern LLMs, reporting contamination across 31 models. \citet{hidayat2025simulating} quantified contamination rates in large-scale benchmarks including MMLU and HellaSwag, while \citet{balunovic2025matharena} demonstrated that performance drops substantially when LLMs are evaluated on MathArena, a private mathematics competition dataset unavailable during training.  
Together, these findings suggest that widely used benchmarks may fail to measure true reasoning ability, as models may exploit memorized patterns rather than demonstrate genuine generalization.

To eliminate this risk, our study \textbf{collects and processes the 2026 CSAT Mathematics exam immediately after public release} and conducts model evaluation right away, ensuring that no LLM could have been exposed to the data beforehand.  
This procedure guarantees a fully contamination-free setting, \textbf{blocking any possibility of prior contamination and enabling faithful measurement of mathematical reasoning using real, previously unseen exam problems}.

\subsection{Prior Work on Input Modality and Prompting Language}

LLM performance is influenced not only by model architecture but also by \textbf{input modality} and \textbf{prompting language}. Early LLMs were optimized for text-only input, but recent advances have produced \textbf{multimodal LLMs} capable of jointly processing images, diagrams, and mathematical notation. This has driven interest in evaluating models on visually grounded mathematics tasks~\citep{lu2024mathvista, wang2024mathvision}.

Prompting language is another key factor. Research on multilingual prompting shows that the chain-of-thought structure, reasoning trajectory, and solution strategy can vary depending on the language in which the problem is presented. Several studies report that English prompts often provide greater stability and accuracy~\citep{park2025cross, zhou2025landscape}.

Building on this literature, we analyze how input modality (text, image, text+visual information) and prompting language (Korean, English) affect model performance in a realistic exam setting. By crossing these factors, we evaluate and compare LLM mathematical reasoning from multiple perspectives.

\subsection{Prior Work on LLM Reasoning and Reasoning Evaluation}

Research on \textbf{reasoning} in LLMs focuses on how to elicit, regulate, and evaluate the \textbf{reasoning process} a model performs before generating a final answer. Zero-shot Chain-of-Thought (CoT)~\citep{kojima2022zeroshot} demonstrated that simple prompts such as “Let’s think step by step” can induce multi-step reasoning without fine-tuning. \citet{wang2022self} proposed generating multiple reasoning paths in parallel and selecting the final answer using majority voting, improving stability and consistency.

A growing body of work examines how to scale test-time reasoning. \citet{snell2024testtimecompute} showed that increasing reasoning depth and compute—known as \textbf{test-time scaling}—can yield stronger performance. \citet{zhou2025landscape} visualized various reasoning types and cognitive trajectories, enabling qualitative assessment of reasoning quality. These studies collectively highlight that the relationship between reasoning length and accuracy is not linear: excessively long reasoning may accumulate errors and degrade performance.

Our work quantitatively evaluates these insights in a real exam setting. By systematically varying the reasoning mode (e.g., \texttt{reasoning\_effort}) on the same CSAT problems, we analyze the \emph{trade-off} between reasoning strength, performance, and computational cost.
\section{Korean CSAT-Math Evaluation Framework for LLMs}
\label{sec:3_framework}

\subsection{Overview of the CSAT}
\label{sec:problem_composition}

The Korean College Scholastic Ability Test (CSAT; Korean: 수능) is a nationally administered, high-stakes standardized examination produced by the Korea Institute for Curriculum and Evaluation (KICE). It serves as the most authoritative assessment used for university admissions in Korea and evaluates whether students possess the academic abilities required for higher education\footnote{\url{https://suneung.re.kr/sub/info.do?m=0101&s=suneung}}.

The CSAT is administered once per year and taken by roughly half a million students nationwide—for example, 554,174 examinees in the 2026 CSAT. Because of its societal significance, the exam has substantial nationwide impact. For example, during the 2026 CSAT held on November 13, 2025, the morning commute for office workers was delayed to support test-takers, and all aircraft takeoffs and landings were suspended during the English listening section to minimize noise\footnote{\url{https://www.yna.co.kr/view/AKR20251111058900003}}.

Due to its importance, the CSAT is constructed under extremely strict security protocols to prevent leakage. All exam writers and reviewers are isolated in a confidential facility beginning approximately 40 days before the exam, with complete prohibition of external communication and electronic devices\footnote{\url{https://www.news1.kr/society/education/5973136}}.  
Because the problem-writing, review, printing, and distribution processes are fully secured end-to-end, the CSAT is widely considered a \textbf{near-impossible-to-leak examination}.

The CSAT consists of five subject areas: Korean, Mathematics, English, Korean History, and elective Social/Science/Vocational studies.  
This work focuses exclusively on the \textbf{Mathematics section}.

Since 2022, the CSAT Mathematics section has consisted of 22 common-subject questions (Math I and Math II) and 8 elective questions selected from Probability and Statistics, Calculus, or Geometry, for a total of 30 questions to be solved in 100 minutes.

In this study, to evaluate overall mathematical competence and cross-subject reasoning, we include \textbf{all 46 questions}—22 common-subject problems and all 24 elective-subject problems from the three elective domains—regardless of the single-subject choice that human test-takers make.

Table~\ref{tab:csat_structure} summarizes the complete structure, problem types, and scoring scheme.

\begin{table}[!t]
\centering
\vspace{-0.5em}
\caption{Structure and Question Types of the CSAT Mathematics Section}
\label{tab:csat_structure}
\vspace{0.7em}
\begin{tabular}{c c c ccc c}
 &  & \textbf{Common} & \multicolumn{3}{c}{\textbf{Elective}} & \multirow{2}{*}{\textbf{Total}}\\
\cmidrule{1-6}
 \multicolumn{2}{c}{\textbf{Subject}} & Math I + Math II & Probability & Calculus & Geometry & \\
\midrule
 \multicolumn{2}{c}{\textbf{Num. of Questions}} & 11 + 11 & 8 & 8 & 8 & 46 \\
\midrule
 \multirow{2}{*}{\textbf{Type}} & Multiple-choice & 15 & 6 & 6 & 6 & 33 \\
 & Short-answer & 7 & 2 & 2 & 2 & 13 \\
\midrule
 \multirow{3}{*}{\textbf{Score}} & 2-point problems & 2 & 1 & 1 & 1 & 5 \\
 & 3-point problems & 6 & 4 & 4 & 4 & 18 \\
 & 4-point problems & 14 & 3 & 3 & 3 & 23 \\
\midrule
 \multicolumn{2}{c}{\textbf{Total Score}} & 74 & 26 & 26 & 26 & 152 \\
\end{tabular}
\end{table}

\begin{itemize}
    \item \textbf{Common Subject (22 problems)}: Math I (11 problems) and Math II (11 problems), taken by all test-takers.
    \item \textbf{Elective Subject (8 problems)}: Students select one among Probability and Statistics, Calculus, and Geometry.
    \item \textbf{Question Types}: Multiple-choice and short-answer.
    \begin{itemize}
        \item \textbf{Short-answer}: Students directly provide an integer between 0 and 999.
        \item \textbf{Multiple-choice}: Five answer options (1–5) are provided.
    \end{itemize}
    \item \textbf{Scoring}: Each problem is worth 2, 3, or 4 points depending on required reasoning complexity.
\end{itemize}

According to the official 2026 CSAT Mathematics curriculum guide\footnote{\url{https://www.suneung.re.kr/boardCnts/fileDown.do?fileSeq=ead4803d0df8cda516056b38fef4c7ae}}, each subject covers the following areas:
\begin{itemize}
    \item \textbf{Common Subjects}
    \begin{itemize}
        \item \textbf{Math I}: Exponential and logarithmic functions, trigonometric functions, sequences
        \item \textbf{Math II}: Limits and continuity, differentiation, integration
    \end{itemize}
    \item \textbf{Elective Subjects}
    \begin{itemize}
        \item \textbf{Probability and Statistics}: Counting, probability, statistical inference
        \item \textbf{Calculus}: Limits of sequences, differential calculus, integral calculus
        \item \textbf{Geometry}: Conic sections, plane vectors, three-dimensional geometry and coordinates
    \end{itemize}
\end{itemize}

\begin{figure}[t]
    \centering
    \includegraphics[width=0.8\linewidth]{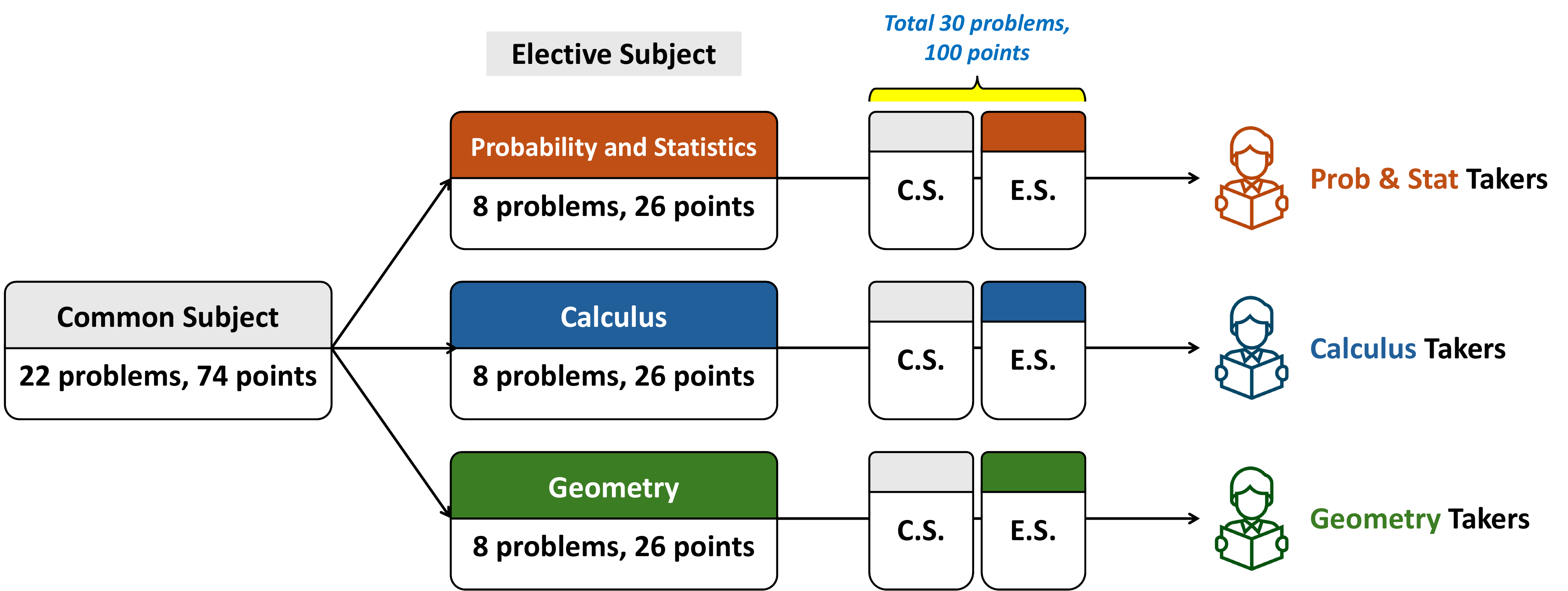}
    \caption{Overall structure of the CSAT Mathematics Section (Common (C.S.) vs. Elective Subjects (E.S.))}
    \label{fig:math3}
\end{figure}

Because the CSAT Mathematics exam has a systematically designed structure—including well-defined scope, problem types, and difficulty—it is particularly suited for evaluating LLM mathematical reasoning. The exam follows a consistent progression from common to elective subjects, as visualized in Figure~\ref{fig:math3}.

\subsection{Data Construction Pipeline}
\label{sec:pipeline}

\begin{figure}[!t]
    \centering
    \includegraphics[width=0.95\linewidth]{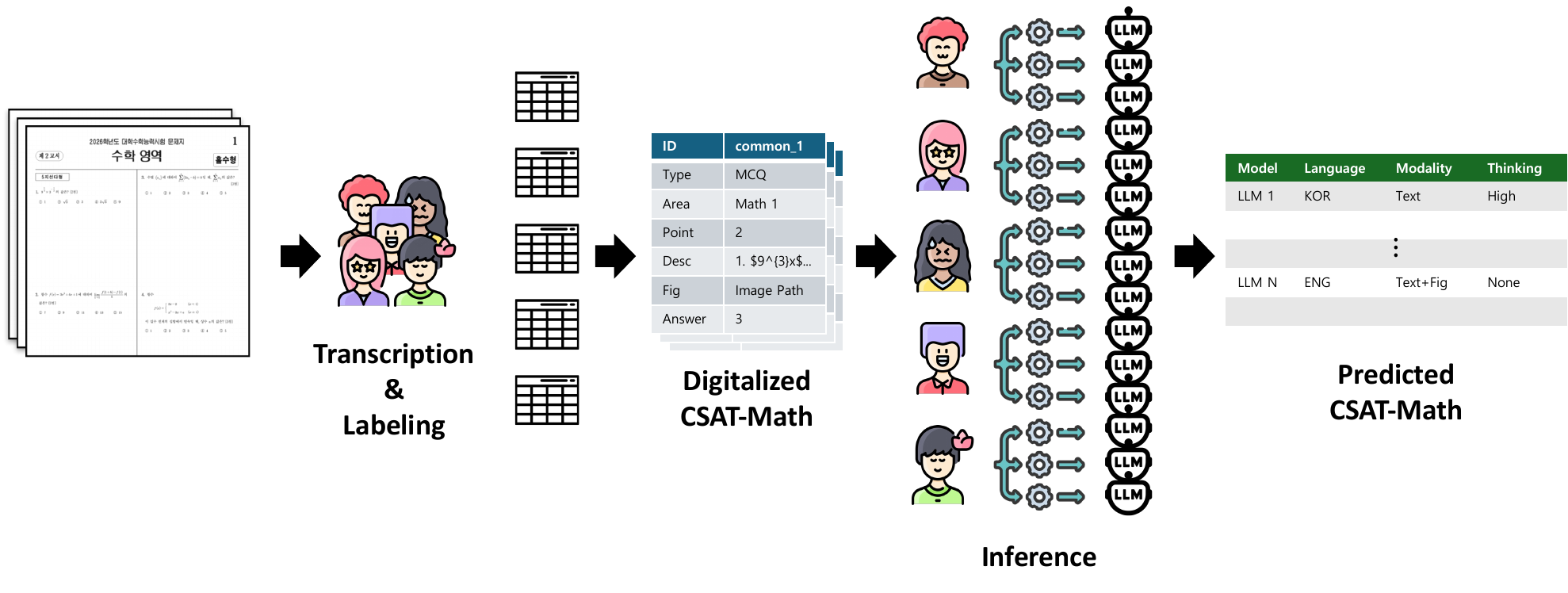}
    \caption{Manual dataset construction (transcription and labeling) and LLM evaluation pipeline}
    \label{fig:pipeline}
\end{figure}

Our evaluation framework is designed to \textbf{completely eliminate} the possibility of prior contamination during model training. To achieve this, we constructed the evaluation dataset and began model evaluation within the first two hours after the official release of the exam.  
A total of seven researchers (six annotators and one reviewer) contributed to the manual dataset construction, and eight participated in the LLM evaluation and monitoring process.  
Figure~\ref{fig:pipeline} illustrates the full pipeline executed immediately after exam release.

The dataset construction process proceeds as follows:
\begin{enumerate}
    \item Annotators receive the Digitalized CSAT-Math template and their assigned question IDs.
    \item Immediately after the official exam release\footnote{Released at 14:10 KST via the KICE website: \url{https://cdn2.kice.re.kr/suneung-26/index.html}}, each annotator extracts all required information for their assigned questions.
    \item A reviewer aggregates all entries and checks for errors and consistency.
    \item A unified Digitalized CSAT-Math dataset is finalized.
\end{enumerate}

\subsection{Structure of the 2026 CSAT Math Question Paper}

The actual question booklet distributed to students consists of:
\begin{itemize}
    \item Subject designation: Common and elective subjects (Probability, Calculus, Geometry)
    \item Question type header: Displayed at the start of each type section
    \item Problem statement: Question description with problem number and score
    \item Scoring: 2-, 3-, or 4-point values
    \item Visual materials: Diagrams, graphs, or figures accompanying the problem
    \item Tables: Used in a small number of questions (e.g., $z$-score tables in Probability/Statistics)
    \item Answer choices: Provided for multiple-choice questions (1–5)
\end{itemize}

\subsubsection{Digitalized CSAT-Math Data Format}

Annotators input structured information into the Digitalized CSAT-Math dataset as follows:

\begin{itemize}
    \item \texttt{prob\_id}: Unique problem ID.  
    Example: \texttt{2026\_odd\_common\_11} refers to the 11th common-subject problem in the odd-form 2026 exam; \texttt{2026\_odd\_probability\_30} refers to the 30th problem in the Probability and Statistics section.
    \item \texttt{prob\_type}: Question type (multiple-choice or short-answer)
    \item \texttt{prob\_area}: Subject area (Math I, Math II, Probability, Calculus, Geometry)
    \item \texttt{prob\_point}: Score value (2, 3, or 4 points)
    \item \texttt{prob\_desc}: Problem text converted into Markdown + \LaTeX
    \item \texttt{prob\_img\_path}: Cropped image of the full problem
    \item \texttt{prob\_fig\_img\_path}: Cropped image of figures or diagrams only
    \item \texttt{answer}: Correct answer (1–5 for multiple-choice; integer 0–999 for short-answer)
\end{itemize}

From 2022 to 2026, \texttt{prob\_id}, \texttt{prob\_type}, and \texttt{prob\_point} remain fixed for each problem index. For example, problems 1–22 are always common-subject questions, and problems 23–30 always correspond to the three elective subject sets.  
Therefore, annotators were instructed to input only the five variable fields:  
\texttt{prob\_area}, \texttt{prob\_desc}, \texttt{prob\_img\_path}, \texttt{prob\_fig\_img\_path}, and \texttt{answer}.  

\subsection{Digitalized CSAT-Math Dataset Construction}

The goal of the dataset construction process is twofold:  
(1) to \textbf{preserve the original structure and mathematical expressions of the exam problems as faithfully as possible}, and  
(2) to \textbf{standardize the data into a consistent format suitable for direct LLM input}.

A total of seven annotators participated in building the Digitalized CSAT-Math dataset.  
Among the 46 questions, four annotators each handled 9 questions and one annotator handled 10 questions, producing the fields \texttt{prob\_desc}, \texttt{answer}, and \texttt{prob\_area}. Additionally, one annotator exclusively extracted all problem and figure images.

\begin{figure}[htbp]
    \centering
    \includegraphics[width=0.9\linewidth]{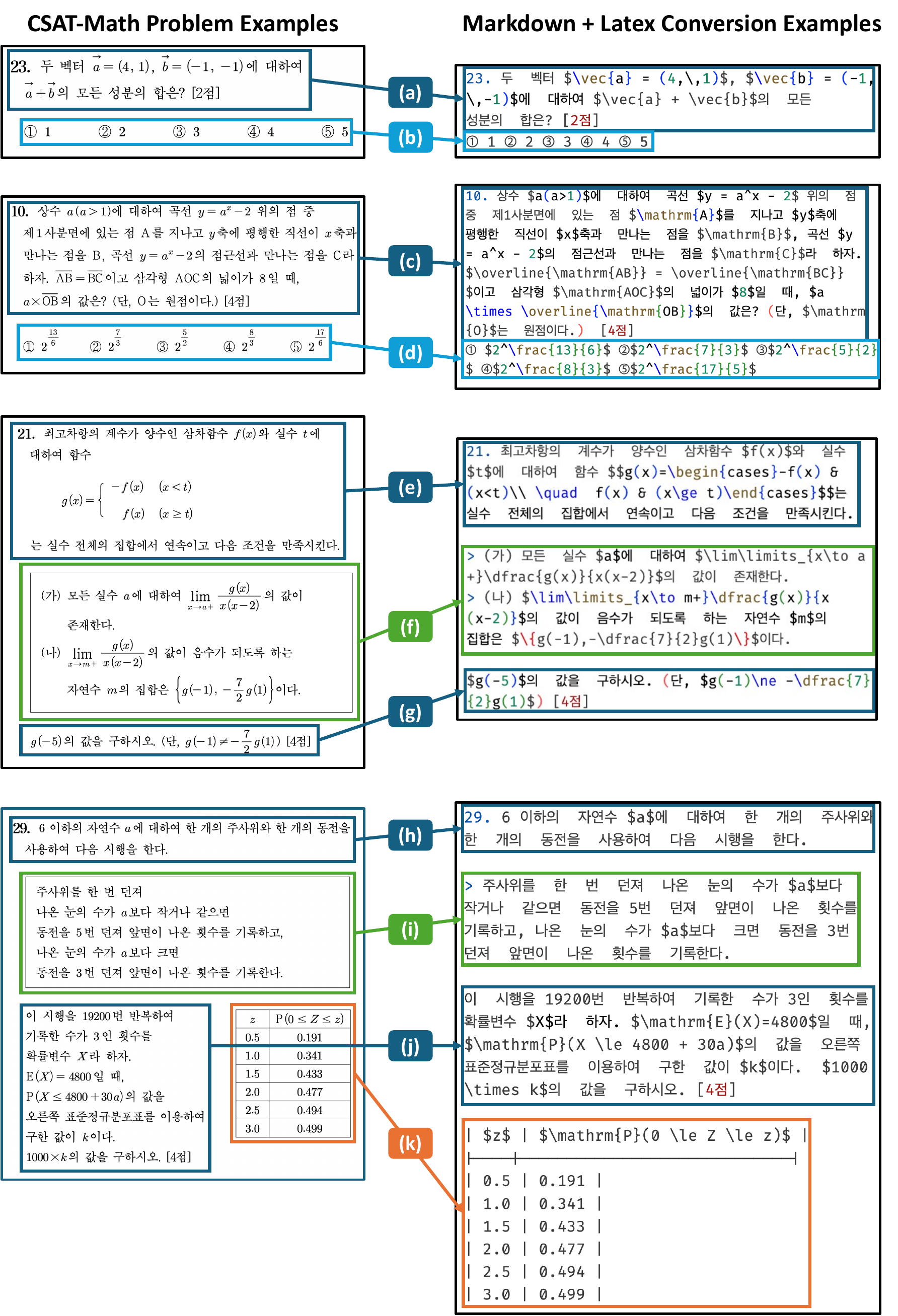}
    \caption{Example of a real CSAT problem and its Digitalized CSAT-Math textual representation}
    \label{fig:textualized_mapped}
\end{figure}

\subsubsection{Text Construction for Each Problem}
\label{sec:dataset}

To enable LLM-based evaluation using PDF CSAT problems, all questions were fully transcribed into LLM-friendly text. Each problem’s stem, choices, score, tables, and mathematical expressions were encoded into the \texttt{prob\_desc} field of the Digitalized CSAT-Math format.  
The \texttt{prob\_desc} field represents a textual reconstruction of the “human-readable problem sheet.”

\paragraph{Extraction of Structured Content Using Markdown}

Markdown provides a lightweight syntax for expressing structured document elements using only plain text. For example, “\texttt{>}” represents a quotation block, and tables can be written using the syntax “\texttt{| a | b |}”.

CSAT mathematics problems contain various structural elements such as boxed passages (Figure~\ref{fig:textualized_mapped} f, i), multiple-choice options (Figure~\ref{fig:textualized_mapped} b, d), and occasional tables (Figure~\ref{fig:textualized_mapped} k).  
When naively converted to plain text, such structures often lose their meaning or readability.  
By contrast, Markdown naturally preserves these structures without information loss.

General paragraphs were written as plain text (Figure~\ref{fig:textualized_mapped} a, c, h), while boxed “보기” sections were represented as block quotes using “\texttt{>}” (Figure~\ref{fig:textualized_mapped} f, i).  
Using Markdown allows the reconstructed text to closely resemble the original layout, and can be visually rendered to match the PDF problem sheet.

\paragraph{Mathematical Expressions Using \LaTeX}

All mathematical expressions were converted into \LaTeX, ensuring clear boundaries between text and math and allowing LLMs to interpret expressions accurately.  
Inline formulas were written as \texttt{\$...\$} (Figure~\ref{fig:textualized_mapped} f, g, j), and block-level expressions were written as \texttt{\$\$...\$\$} (Figure~\ref{fig:textualized_mapped} e).

We reproduced the printed formulas in the CSAT as closely as possible:  
non-italic text (e.g., $\mathrm{P}$) was expressed using \texttt{\textbackslash mathrm\{\}}, vectors and segment lengths were written using \texttt{\textbackslash vec\{AB\}} and \texttt{\textbackslash overline\{AB\}} respectively, and operators such as limits or summations used \texttt{\textbackslash limits} to ensure upper/lower bounds appear above and below as in the original problem sheet.

\paragraph{Representation of Problem Components (Number, Score, Answer Choices)}

Problem numbers, scoring labels (e.g., “[4점]”), and answer choices were transcribed exactly as printed.  
Multiple-choice options were preserved using original circled numerals (①–⑤), e.g., “① 2  ② 3  ③ 4  ④ 5  ⑤ 6”.  
This ensured that the Digitalized CSAT-Math text closely mirrored the actual problem layout.

\subsubsection{Metadata Construction for Each Problem}

We define \textbf{metadata} as information that is required for evaluation and analysis but is not explicitly visible to LLMs in the original exam sheet.  
This includes the true \textit{subject area} of each problem and its \textit{correct answer.}

\paragraph{Actual Subject Area per Problem}

As described in Section~\ref{sec:problem_composition}, CSAT Mathematics consists of Math I, Math II, and one of three elective subjects.  
However, within the 1–22 common-subject questions, the subject source (Math I vs. Math II) is not indicated in the problem sheet and must be manually identified.

Because this information is not officially provided, we manually annotated each problem in the 1–22 range as belonging to Math I or Math II, constructing the \texttt{prob\_area} field accordingly.

\paragraph{Ground-Truth Answers}

The \texttt{answer} field holds the correct answer for each problem.  
Multiple-choice answers were stored as integers corresponding to their options (e.g., ① $\rightarrow$ \texttt{1}, ⑤ $\rightarrow$ \texttt{5}).  
Short-answer problems were recorded as integers within the 0–999 range.  
Expressing all answers in a unified numerical form enables consistent scoring across both multiple-choice and short-answer problems.

\begin{figure}[!t]
    \centering
    \includegraphics[width=\linewidth]{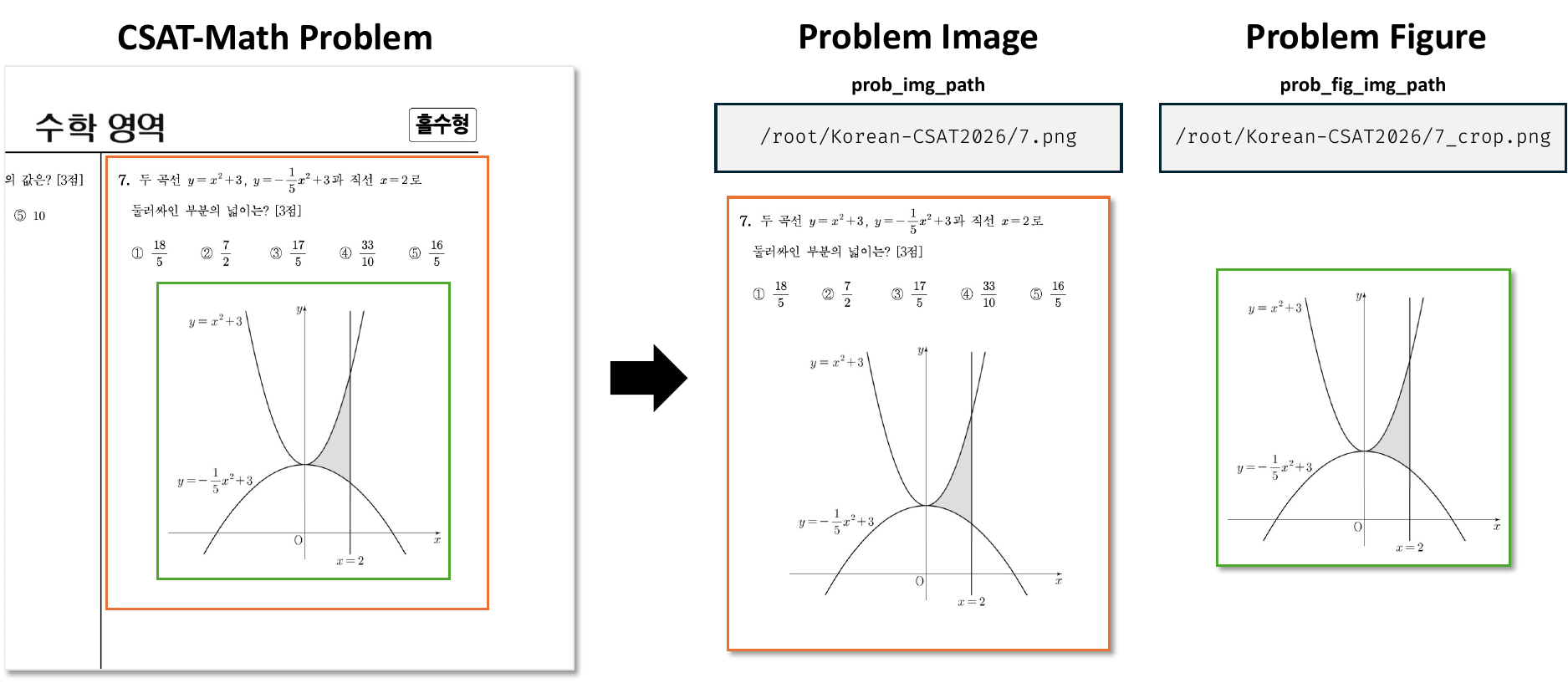}
    \caption{Examples of extracted CSAT problem images}
    \vspace{-1em}
    \label{fig:prob_img_example}
\end{figure}

\subsubsection{Image Data Construction}

\paragraph{Image Extraction}

Many CSAT math questions contain diagrams or visual elements.  
In the 2026 exam, 8 of the 46 questions included such visual components.  
To analyze the effect of visual information on LLM mathematical reasoning, we extracted two types of images from the original PDF:

\begin{itemize}
    \item \textbf{Problem Image}: Full problem region (\texttt{prob\_img\_path})
    \item \textbf{Problem Figure}: Only diagrams/graphs/visual elements (\texttt{prob\_fig\_img\_path})
\end{itemize}

Figure~\ref{fig:prob_img_example} shows examples of these extracted image components.

\subsubsection{Quality Assurance}

Annotators and reviewers rendered each problem using Markdown editors (Obsidian\footnote{\url{https://obsidian.md/}} or StackEdit\footnote{\url{https://stackedit.io/}}) to verify visual alignment with the original PDF problem sheet.  
They checked Markdown+\LaTeX rendering accuracy—including formula formatting, line breaks, symbols, and layout.  
A final pass by the supervisor ensured consistency and correctness across all entries.

\begin{figure}[!t]
    \centering
    \includegraphics[width=0.7\linewidth]{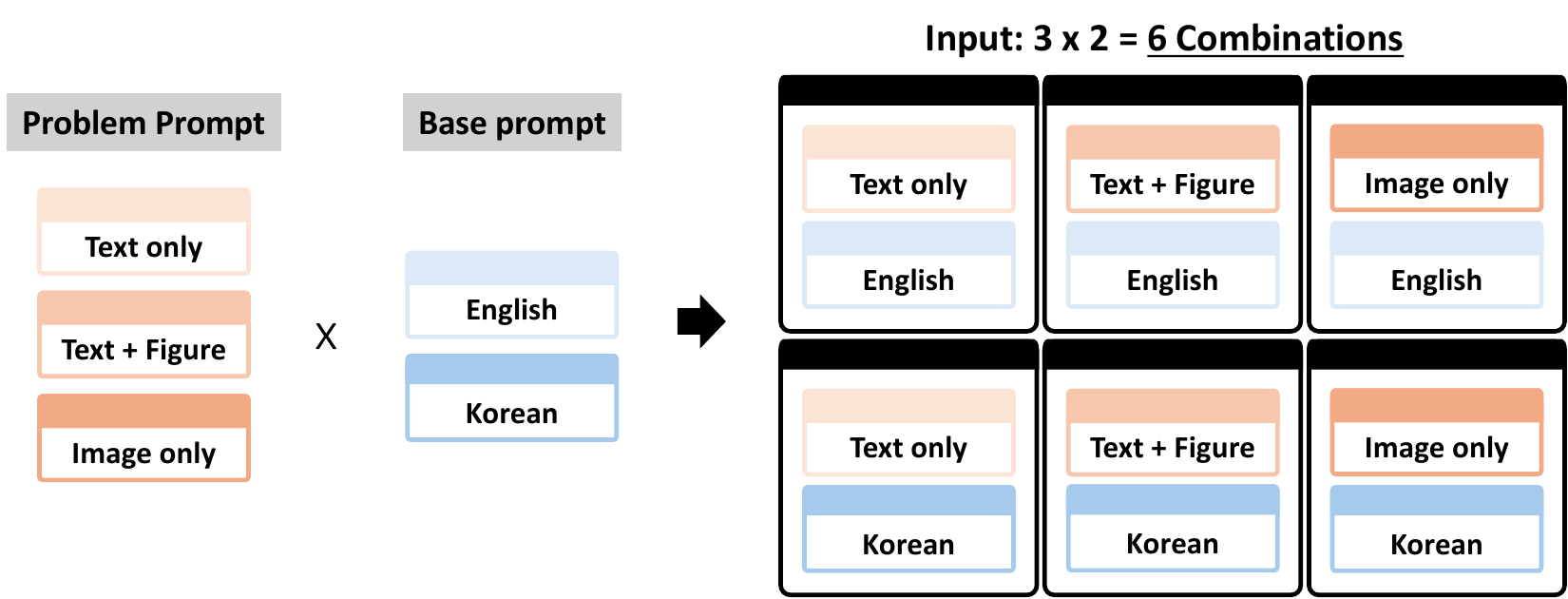}
    \caption{Example of prompt composition}
    \vspace{-1em}
    \label{fig:prompt_skeleton}
\end{figure}

\subsection{Prompt Template for Problem Solving}
\label{sec:prompt_base}

LLM prompts consist of two components:  
(1) the \textbf{problem}, containing the CSAT math content, and  
(2) the \textbf{instruction}, which guides the model on how to solve the problem in a CSAT-appropriate manner.

To evaluate both multimodal and multilingual capabilities, we created multiple combinations of problem input modalities and instruction languages.  
There are three input modalities and two instruction-language settings, resulting in six total prompt conditions.  
Figure~\ref{fig:prompt_skeleton} illustrates these combinations.

\begin{figure}[htbp]
    \centering
    \includegraphics[width=\linewidth]{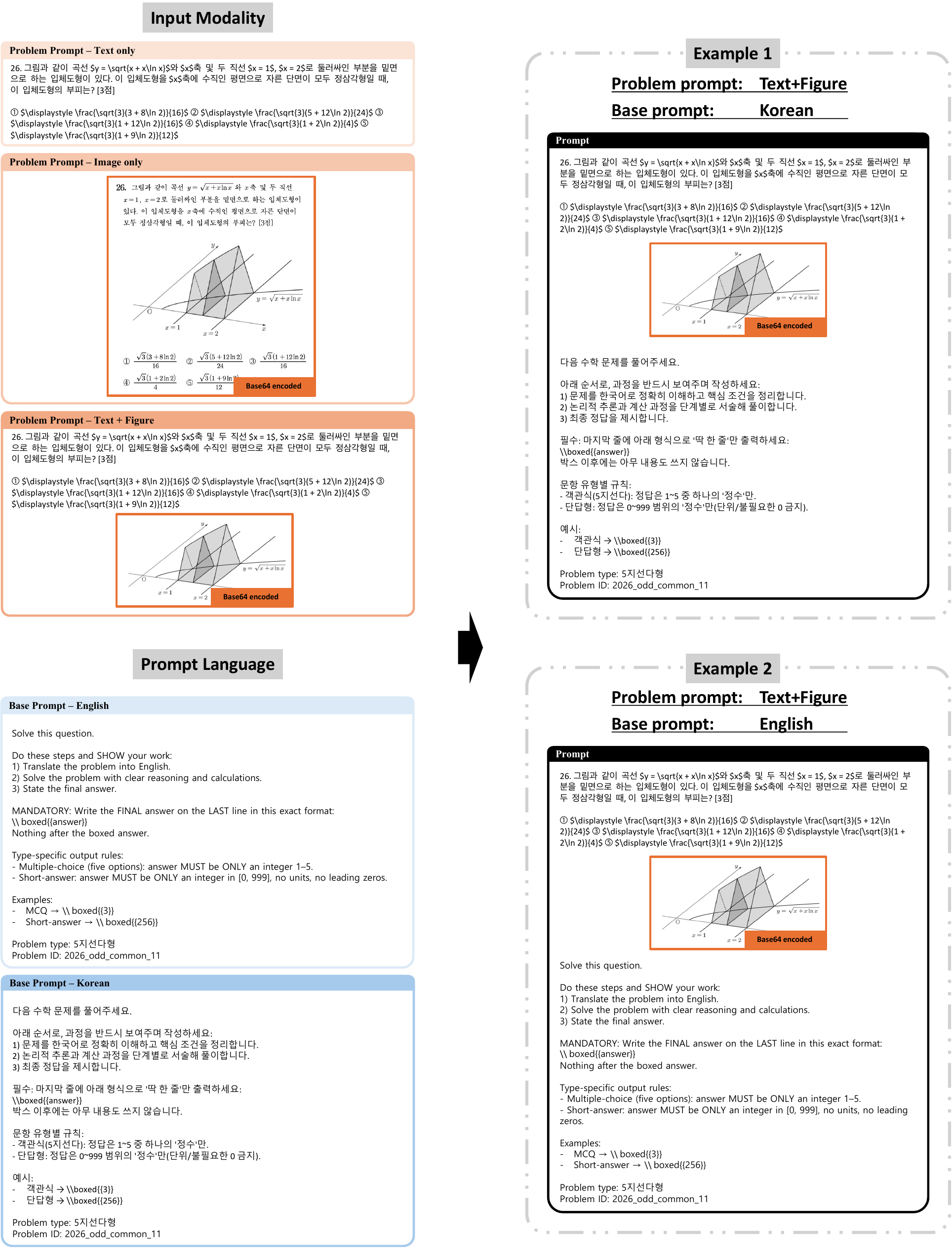}
    \caption{Example of a full prompt}
    \label{fig:prompt_whole}
\end{figure}

\paragraph{Input Modalities}

Because CSAT questions may include both textual and visual elements, we evaluate LLMs under three input formats:

\begin{itemize}
    \item \textbf{Text only}: The problem is provided solely as text (\texttt{prob\_desc}).  
    All diagrams or figures are omitted. This modality is supported by all models.
    
    \item \textbf{Image only}: The model receives the full problem as an image (Problem Image) along with the instruction.  
    This setting mirrors actual test-taking conditions and evaluates the model’s OCR and visual reasoning capabilities.  
    Text-only LLMs are excluded from this condition.
    
    \item \textbf{Text + Figure}: The textual portion is provided as \texttt{prob\_desc}, while diagrams or graphs are separately provided as images (Problem Figure).  
    This hybrid setting requires both text understanding and visual interpretation.  
    Text-only models are excluded.
\end{itemize}

\paragraph{Prompting Language}

To analyze multilingual prompting behavior, we use two instruction-language conditions:

\begin{itemize}
    \item \textbf{Korean}: Both the problem and the instruction are presented in Korean.
    \item \textbf{English}: The problem text remains in Korean, but the instruction is given in English.
\end{itemize}

Combining three modalities and two languages yields six conditions (Figure~\ref{fig:prompt_skeleton}).  
Text-only models are evaluated under the two text-based conditions, while multimodal models are evaluated under all six.  
Examples of full prompt configurations are shown in Figure~\ref{fig:prompt_whole}.
\section{CSAT-Math LLM Experiment Design}
\label{sec:4_experiment_setup}

In this section, we describe the overall experimental design.  
Specifically, we (1) detail the list of LLMs used in the evaluation and their execution environment, and  
(2) explain the experimental goals and procedures aligned with the four core research questions (RQs).

\subsection{Model List}

We target high-performing LLMs capable of solving CSAT-level math problems, while also considering practical deployability.  
To this end, we restrict our evaluation to models satisfying the following criteria:

\begin{itemize}
    \item \textbf{Performance}:  
    To pre-screen mathematical capabilities, we only consider the 24 models highlighted on the AIME 2025 benchmark homepage\footnote{\url{https://artificialanalysis.ai/evaluations/aime-2025}} as of October 5, 2025.  
    This allows us to compare detailed behaviors on the CSAT among models that already demonstrate strong performance and recognition on a challenging competition-style math benchmark.
    
    \item \textbf{Accessibility}:  
    For experimental convenience and reproducibility, we only include models that were directly accessible via OpenRouter as of October 5, 2025.  
    Using a unified API interface simplifies our evaluation pipeline and enables future researchers to reproduce or extend our setup without changing the underlying infrastructure.
    
    \item \textbf{Release time}:  
    To maintain a \textbf{zero-leakage} evaluation environment, we only include models that were released \textbf{before} the 2026 CSAT date (November 13, 2025).  
    Models released or updated after the exam date are excluded, as they may potentially have been trained or tuned on post-exam content.
    
    \item \textbf{Diversity}:  
    To avoid bias toward specific model families, we balance \textbf{proprietary models} (e.g., GPT, Claude, Gemini) and \textbf{open-weight models} (e.g., Qwen, Llama, DeepSeek).  
    We also include models with varying parameter scales (from billions to hundreds of billions) and architectural designs (standard LLMs vs. reasoning-focused variants) to analyze performance differences as a function of family and size under a shared evaluation setting.
\end{itemize}

Table~\ref{tab:model_setting} summarizes the final set of models, including provider, API pricing, parameter scale, and whether a built-in reasoning/thinking mode is supported.

\begin{table}[!t]
\vspace{-1em}
\centering
\caption{List of LLMs, providers, prices, model scales, and Reasoning/Thinking support. “1M Input” and “1M Output” denote the price per 1 million input/output tokens, respectively, as specified on OpenRouter by each model provider. Model scale is reported in terms of active parameters; when not publicly available, we estimate it based on prior documentation and reported performance (marked with $\sim$$^{\ast}$). Reasoning support is marked with \checkmark if available and “--” otherwise.}
\vspace{0.7em}
\label{tab:model_setting}
\begin{adjustbox}{width=\textwidth}
\begin{tabular}{c lll rr cc}
\toprule
\textbf{Type} & & \textbf{Model} & \textbf{Model Provider} & \textbf{1M Input} & \textbf{1M Output} & \textbf{Scale} & \textbf{Reasoning Support} \\
\midrule
\multirow{12}{*}{Closed}
& \multirow{4}{*}{\includegraphics[height=0.8em]{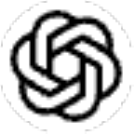} GPT}
& GPT-5 Codex & openai & \$1.25 & \$10.00 & $\sim$635B$^{\ast}$ & -- \\
& & GPT-5 & openai & \$1.25 & \$10.00 & $\sim$330B$^{\ast}$ & \checkmark \\
& & GPT-5-mini & openai & \$0.25 & \$2.00 & $\sim$27B$^{\ast}$ &  \checkmark \\
& & GPT-5-nano & openai & \$0.05 & \$0.40 & $\sim$15B$^{\ast}$ &  \checkmark \\
\cmidrule{2-8}
& \multirow{2}{*}{\includegraphics[height=0.8em]{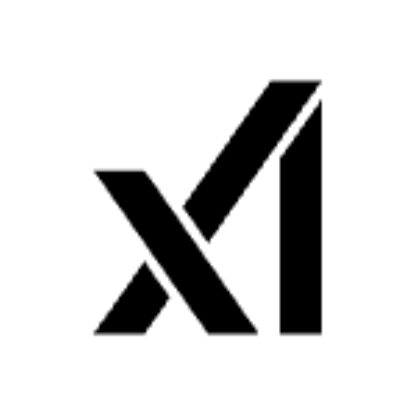} Grok}
& Grok 4& xai  & \$6.00 & \$30.00 & $\sim$213B$^{\ast}$ & \checkmark \\
& & Grok 4 Fast & xai & \$0.40 & \$1.00 & $\sim$40B$^{\ast}$ & -- \\
\cmidrule{2-8}
& \multirow{3}{*}{\includegraphics[height=0.8em]{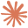} Claude}
& Claude 4.5 Sonnet & anthropic & \$6.00 & \$22.50 & $\sim$100B$^{\ast}$ & \checkmark \\
& & Claude 4.5 Haiku & anthropic & \$1.00 & \$5.00 & $\sim$30B$^{\ast}$ & \checkmark \\
& & Claude 4.1 Opus & anthropic & \$15.00 & \$75.00 & $\sim$400B$^{\ast}$ & \checkmark \\
\cmidrule{2-8}
& \multirow{2}{*}{\includegraphics[height=0.8em]{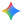} Gemini}
& Gemini 2.5 Pro & google-vertex/global & \$2.50 & \$15.00 & $\sim$128B$^{\ast}$ & \checkmark \\
& & Gemini 2.5 Flash & google-vertex/global & \$0.30 & \$2.50\$ & $\sim$27B$^{\ast}$ & -- \\
\midrule
\multirow{13}{*}{Open}
& \multirow{1}{*}{\includegraphics[height=0.8em]{icon/openai.pdf} GPT}
& gpt-oss 20B & hyperbolic & \$0.04 & \$0.04 & 3.6B & \checkmark \\
\cmidrule{2-8}
& \multirow{2}{*}{\includegraphics[height=0.8em]{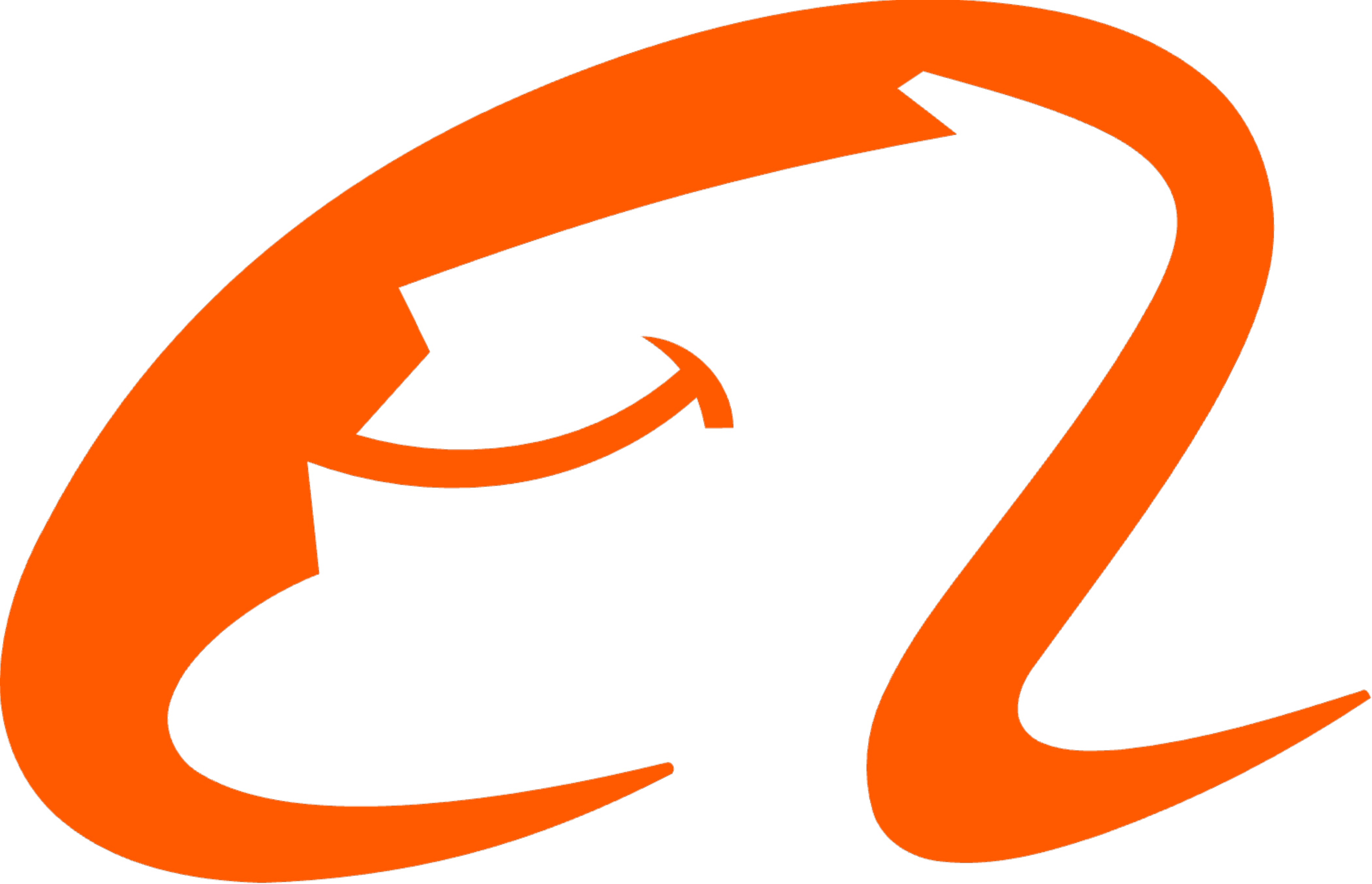} Qwen}
& Qwen3 235B A22B & siliconflow/fp8 & \$0.30 & \$1.50 & 22B & -- \\
& & Qwen3 235B A22B Thinking 2507 & siliconflow/fp8 & \$0.45 & \$3.50 & 22B & \checkmark \\
\cmidrule{2-8}
& \multirow{2}{*}{\includegraphics[height=0.8em]{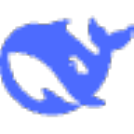} Deepseek}
& Deepseek R1 0528 & novita/fp8 & \$0.27 & \$0.41 & $\sim$37B$^{\ast}$ & \checkmark \\
& & Deepseek V3.2 Exp & novita/fp8 & \$0.70 & \$2.50 & $\sim$37B$^{\ast}$ & -- \\
\cmidrule{2-8}
& \multirow{1}{*}{\includegraphics[height=0.8em]{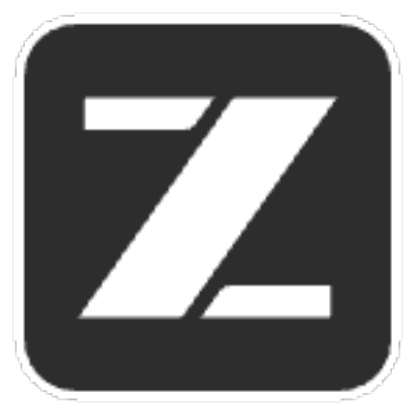} GLM}
& GLM 4.6 & z-ai & \$0.60 & \$2.20 & 32B & -- \\
\cmidrule{2-8}
& \multirow{2}{*}{\includegraphics[height=0.8em]{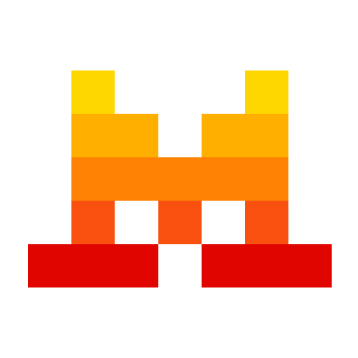} Magistral}
& Magistral Medium 2506 & mistral& \$2.00 & \$5.00  & 45B & -- \\
& & Magistral Medium 2506 Thinking & mistral & \$2.00 & \$5.00 & 45B & \checkmark \\
\cmidrule{2-8}
& \multirow{1}{*}{\includegraphics[height=0.8em]{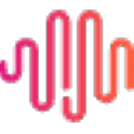} MiniMax}
& MiniMax M2 & minimax & \$0.30 & \$1.20 & $\sim$10B$^{\ast}$ & -- \\
\cmidrule{2-8}
& \multirow{2}{*}{\includegraphics[height=0.8em]{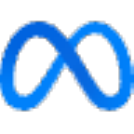} Llama}
& Llama 4 Maverick & friendli & \$0.20 & \$0.60 & 17B & -- \\
& & Llama 3.3 Nemotron Super 49B V1.5 & deepinfra/fp8 & \$0.10 & \$0.40 & 49B & -- \\
\cmidrule{2-8}
& \multirow{2}{*}{\includegraphics[height=0.8em]{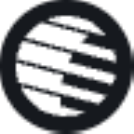} Kimi}
& Kimi K2 0905 & moonshotai & \$0.60 & \$2.50 & 32B & -- \\
& & Kimi K2 0905 Thinking & moonshotai & \$0.60 & \$2.50 & 32B & \checkmark \\
\bottomrule
\end{tabular}
\end{adjustbox}
\vspace{-1em}
\end{table}

\subsection{Configuring Model Reasoning}
\label{sec:Reasoning-model}

Modern LLMs increasingly move beyond simple pattern matching and incorporate built-in \textbf{reasoning} mechanisms that analyze problem structure and generate intermediate thoughts.  
In this work, we define reasoning as \textbf{“the internal thinking or multi-step inference process a model performs before producing a final answer”}.  
Typically, such reasoning processes consume separate reasoning tokens, distinct from the visible output tokens.

While both proprietary and open-weight models have begun to expose some reasoning functionality, these are often available only through dedicated “reasoning models” or with limited, coarse-grained options. Fine-grained control over reasoning strength is rarely supported.  
Therefore, for systematic comparison, we conduct a dedicated reasoning study only on GPT-5, which allows us to control both reasoning strength and explanation length.

GPT-5 exposes two key configuration knobs:
(1) \texttt{Reasoning\_Effort}: internal reasoning strength (amount of reasoning tokens), and  
(2) \texttt{Text\_Verbosity}: length of the final explanation.  
These two parameters are independent: for example, one can use high reasoning effort with a concise explanation (high effort + low verbosity), or minimal reasoning with a long explanation (minimal effort + high verbosity).

\paragraph{Reasoning\_Effort}

GPT-5 provides four discrete levels for internal reasoning:

\begin{itemize}
    \item \texttt{minimal}: minimizes reasoning tokens; fastest responses
    \item \texttt{low}: basic reasoning; sufficient for simpler problems
    \item \texttt{medium} (default): balanced reasoning; stable for typical math problems
    \item \texttt{high}: deep reasoning; suitable for difficult, complex problems
\end{itemize}

\paragraph{Text\_Verbosity}

\texttt{Text\_Verbosity} controls the length of the final response, independently of \texttt{Reasoning\_Effort}:

\begin{itemize}
    \item \texttt{low}: concise answers; may omit detailed reasoning steps
    \item \texttt{medium} (default): standard explanatory length
    \item \texttt{high}: detailed, extended explanations
\end{itemize}

\paragraph{\texttt{Reasoning\_Effort} × \texttt{Text\_Verbosity} Combinations}

Combining the four effort levels with the three verbosity levels yields 12 distinct configurations.  
Table~\ref{tab:Reasoning_verbosity} summarizes their qualitative characteristics.  
We treat each configuration as a separate experimental condition and compare their behavior on the CSAT.

\begin{table}[t]
\centering
\setlength{\tabcolsep}{4pt}
\caption{Qualitative characteristics of GPT-5 Reasoning Effort × Text Verbosity combinations}
\vspace{0.7em}
\label{tab:Reasoning_verbosity}
\begin{tabular}{c c l}
\toprule
\textbf{Effort} & \textbf{Verbosity} & \multicolumn{1}{c}{\textbf{Characteristics}} \\
\midrule
\midrule
\multirow{3}{*}{minimal} & low    & Minimal reasoning × short output \\
 & medium & Minimal reasoning × medium-length output \\
 & high   & Minimal reasoning × long output \\
\midrule
\multirow{3}{*}{low}  & low    & Low reasoning × short output \\
  & medium & Low reasoning × medium-length output \\
  & high   & Low reasoning × long output \\
\midrule
\multirow{3}{*}{medium} & low    & Medium reasoning × short output \\
 & medium & Medium reasoning × medium-length output (default) \\
 & high   & Medium reasoning × long output \\
\midrule
\multirow{3}{*}{high} & low    & High reasoning × short output \\
 & medium & High reasoning × medium-length output \\
 & high   & High reasoning ×long output \\
\bottomrule
\end{tabular}
\vspace{-1em}
\end{table}

\subsection{API and Runtime Environment}

All experiments were conducted via the OpenRouter API\footnote{\url{https://openrouter.ai}}, using the following fixed hyperparameters for consistent evaluation:

\begin{itemize}
    \item \textbf{Temperature}: 0.0 \\
    Temperature controls the randomness in token sampling. Higher values increase diversity in generated outputs, while values closer to 0 bias the model toward the most likely tokens, yielding more deterministic behavior.  
    Since our primary goal is to analyze performance differences across input modalities (Text-only, Image-only, Text+Figure) and prompting languages (Korean, English), we fix Temperature at 0.0 to minimize randomness-induced variance.
    
    \item \textbf{Top-\textit{p}}: 0.0 \\
    Top-\textit{p} controls nucleus sampling by restricting candidate tokens to a cumulative probability mass of \textit{p}. Larger values introduce more variability; values near 0 strongly favor the most probable token.  
    We set Top-\textit{p} to 0.0 to eliminate sampling variability and ensure deterministic outputs.
    
    \item \textbf{Max tokens}: Unconstrained, up to each model’s maximum setting.
    
    \item \textbf{Retry policy}:  
    On API failure, we retry up to 5 times with an exponential backoff strategy\footnote{\url{https://en.wikipedia.org/wiki/Exponential_backoff}} (e.g., 5s → 10s → 20s → 40s …), ensuring robust completion of large-scale experiments.
    
    \item \textbf{Timeout}: None.
\end{itemize}

Experiments were performed by eight researchers using personal PCs and servers. The environment is as follows:

\begin{itemize}
    \item \textbf{OS}: Ubuntu 22.04 LTS or Windows 10
    \item \textbf{Language/runtime}: Python 3.10
    \item \textbf{Key libraries}:
    \begin{itemize}
        \item \texttt{openai==2.8.0}: OpenRouter-compatible API calls and response handling
        \item \texttt{pandas==2.3.3}: Structuring results and generating leaderboards
    \end{itemize}
    \item \textbf{Parallelization}: Up to 16 processes in parallel.  
    Each process is run via a single Python command, solving all 46 questions sequentially for a specific model–input-condition pair, orchestrated via shell scripts.
\end{itemize}

\subsection{LLM Response Logging and Predicted CSAT-Math Format}
\label{sec:outputs}

The problem-solving script automatically logs all information at the question level for every API call. Logged fields include:

\begin{itemize}
    \item Full API request
    \item Full model response
    \item Extracted final answer
    \item Actual response time per problem (latency, KST)
    \item API token usage (input / output / total)
    \item Error type and number of retries (if any)
\end{itemize}

The \textbf{Predicted CSAT-Math format} is defined as follows:

\begin{itemize}
    \item \texttt{prob\_id}: Unique problem ID
    \item \texttt{prob\_type}: Problem type [multiple-choice or short-answer]
    \item \texttt{prob\_area}: Subject area [Math I, Math II, Probability and Statistics, Calculus, Geometry]
    \item \texttt{answer}: Ground-truth answer [①–⑤ for multiple-choice, 0–999 integer for short-answer]
    \item \texttt{model\_index}: Internal model index
    \item \texttt{model\_id}: Model identifier for OpenRouter requests (e.g., \texttt{openai/gpt-5-codex})
    \item \texttt{provider}: Serving provider (e.g., \texttt{novita/fp8}), which can affect quantization and pricing
    \item \texttt{content\_mode}: Input modality [\texttt{text\_only}, \texttt{image\_only}, or \texttt{text\_fig}]
    \item \texttt{language}: Prompt language [\texttt{ko} (Korean) or \texttt{en} (English)]
    \item \texttt{variant\_tag}: Tag used for naming log files
    \item \texttt{start\_time\_kst}: Start time (KST) of problem solving (e.g., 2025-11-13 15:52:28.894)
    \item \texttt{end\_time\_kst}: End time (KST) of problem solving (e.g., 2025-11-13 15:56:20.391)
    \item \texttt{model\_answer}: Parsed answer extracted from the model output.  
    The instruction requires that the final answer be enclosed in \texttt{\textbackslash boxed\{\}}, and we parse the content inside the last \texttt{\textbackslash boxed\{\}}.
    \item \texttt{elapsed\_sec}: Time taken to solve the problem, computed from \texttt{start\_time\_kst} and \texttt{end\_time\_kst} (in seconds with millisecond precision)
    \item \texttt{prompt\_tokens}: Number of input tokens used for the prompt
    \item \texttt{completion\_tokens}: Number of tokens generated by the model
    \item \texttt{total\_tokens}: Sum of input and output tokens, representing total token usage per problem
    \item \texttt{attempt\_count}: Number of attempts; 1 if successful on the first try, 0 if failed due to unrecovered errors
    \item \texttt{error}: Logged error type, if any
    \item \texttt{request\_body}: Additional options passed in the model call (e.g., \texttt{{"Reasoning\_effort": "high", "text\_verbosity": "low"}})
\end{itemize}

Using this logs, we analyze correlations between performance and latency, modality-specific behavior, and stability across model families and sizes.

\subsection{Goals of the CSAT-Math Experiments}

To answer the four research questions defined in Section~\ref{sec:intro} (RQ1–RQ4), we use the LLM response logs and the Predicted CSAT-Math data described in Section~\ref{sec:outputs}.  
By systematically varying input conditions, problem characteristics, and reasoning configurations, we examine LLM mathematical problem-solving ability from multiple angles.

\begin{itemize}
\item \textbf{RQ1}: Compare overall performance of recent LLMs and derive a ranking based on total score and accuracy over all 46 problems.

\item \textbf{RQ2}: Analyze which concepts and item types exhibit concentrated errors, considering subject area (Math I/II, Calculus, Geometry, Probability and Statistics), score weight (2/3/4 points), problem format (multiple-choice vs. short-answer), and structural/difficulty factors.

\item \textbf{RQ3}: Compare how input modality and prompting language affect accuracy, error patterns, and latency.

% Because CSAT problems combine text, notation, and figures, compare how input modality and prompting language affect accuracy, error patterns, and latency.

\item \textbf{RQ4}: Quantitatively evaluate how changes in reasoning strength affect performance, cost, and latency.
\end{itemize}

This evaluation framework cross-combines multiple axes—(i) model family and size, (ii) prompting language, (iii) input modality, and (iv) reasoning configuration—to measure LLM math capabilities in a fine-grained manner and to isolate the experimental conditions relevant to each RQ.

\subsection{Overall Experimental Configuration}

Table~\ref{tab:experiment_design} summarizes our experimental design.  
We evaluate 24 models under all combinations of \textbf{prompt language (2) × input modality (3)}, and additionally run a reasoning ablation for the GPT-5 family over all \texttt{Reasoning\_Effort} and \texttt{Text\_Verbosity} combinations (4 × 3 = 12).  
This multi-faceted setup enables a precise analysis of math performance, modality sensitivity, language effects, and the trade-off between reasoning strength, cost, and latency.

\begin{table}[t]
\centering
\caption{Summary of experimental settings. We evaluate 24 models across language and input-modality conditions, and conduct additional Reasoning configuration experiments (Reasoning\_Effort × Text\_Verbosity) for GPT-5.}
\vspace{0.7em}
\label{tab:experiment_design}
\setlength{\tabcolsep}{8pt}
\renewcommand{\arraystretch}{1.25}
\begin{tabular}{l c l}
\toprule
\textbf{Axis} & \textbf{\# Levels} & \textbf{Levels} \\
\midrule
Model               & 24 & LLM and multimodal models (Table~\ref{tab:model_setting}) \\
Language            & 2  & Korean, English \\
Input modality      & 3  & Text, Image, Text+Figure \\
\midrule
GPT-5 Reasoning\_Effort    & 4  & Minimal, Low, Medium, High \\
GPT-5 Text\_Verbosity & 3  & Low, Medium, High \\
\midrule
Base combinations   & $24 \times 2 \times 3$ & Conditions shared by all models \\
GPT-5 extra combos  & $4 \times 3$           & Effort–Verbosity combinations \\
\bottomrule
\end{tabular}
\end{table}

\subsection{Evaluation Metrics}

We compute evaluation metrics over the full Korean-CSAT-Math benchmark under a unified scoring scheme. 
Although human test-takers answer only one elective subject—solving 30 problems in total (22 common + 8 elective)—we require LLMs to answer all three elective sets, yielding 46 problems overall.
% Although human test-takers answer only one elective subject, we require LLMs to answer all three elective sets, yielding a total of 46 problems.

\begin{itemize}
    \item \textbf{Score}: Sum of the point values of correctly answered problems.  
    For example, if a model correctly answers five 2-point problems, fourteen 3-point problems, and seventeen 4-point problems, the score is $(2\times5) + (3\times17) + (4\times18) = 133$.
    
    \item \textbf{Normalized Score}: Total score normalized to a 100-point scale, where 152 is the maximum (Table~\ref{tab:csat_structure}: 74 points for common + 26 each for Probability, Calculus, and Geometry).  
    The formula is $\dfrac{\text{score}\times100}{152}$.  
    In the above example, $\dfrac{133\times100}{152} = 87.5$.
    
    \item \textbf{Latency}: Time taken for the model to generate a response, measured from API invocation until the full response is received.  
    We compute this using \texttt{elapsed\_sec} in the Predicted CSAT-Math logs and report it in seconds (with millisecond precision).
    
    \item \textbf{Input Token Usage}: Number of tokens consumed by the prompt.
    
    \item \textbf{Output Token Usage}: Number of tokens generated in the completion.
    
    \item \textbf{Total Token Usage}: Sum of input and output tokens (the \texttt{total\_tokens} field), representing total token consumption per problem.
    
    \item \textbf{Cost (\$)}: Per-problem inference cost, computed from OpenRouter’s pricing policy for each model (Table~\ref{tab:model_setting}).  
    Costs are calculated based on both input and output token usage.  
    Note that even open-weight models (e.g., gpt-oss-20B) incur API costs when accessed via OpenRouter, although they may be freely usable outside this setting.  
    We apply a consistent OpenRouter-based formula to all models.
\end{itemize}
\section{LLM Experiment Results}
\label{sec:5_experiment}

%---------------------------------------------------------------------------------------------
% RQ0. Which models currently solve CSAT-level math problems best?
%---------------------------------------------------------------------------------------------

\subsection{LLM Score Leaderboard (Text-only / Korean)}
\label{sec:leaderboard}

The overall performance of each model under the proposed Korean-CSAT LLM evaluation framework is summarized in Table~\ref{tab:score} in Appendix~\ref{sec:appendix_whole_performance}.  
More detailed results and the latest leaderboard are available on the official website\footnote{\url{https://isoft.cnu.ac.kr/csat2026/}}.  
Below, we present key findings observed under this specific configuration.

The main analysis in this subsection focuses on the condition \textbf{input modality: Text-only, prompt language: Korean}.  
This choice reflects the presence of models that do not support image input and the fact that the target benchmark consists of Korean CSAT problems.

\paragraph{Model-wise performance analysis}

\begin{figure}[!t]
    \centering
    \includegraphics[width=\linewidth]{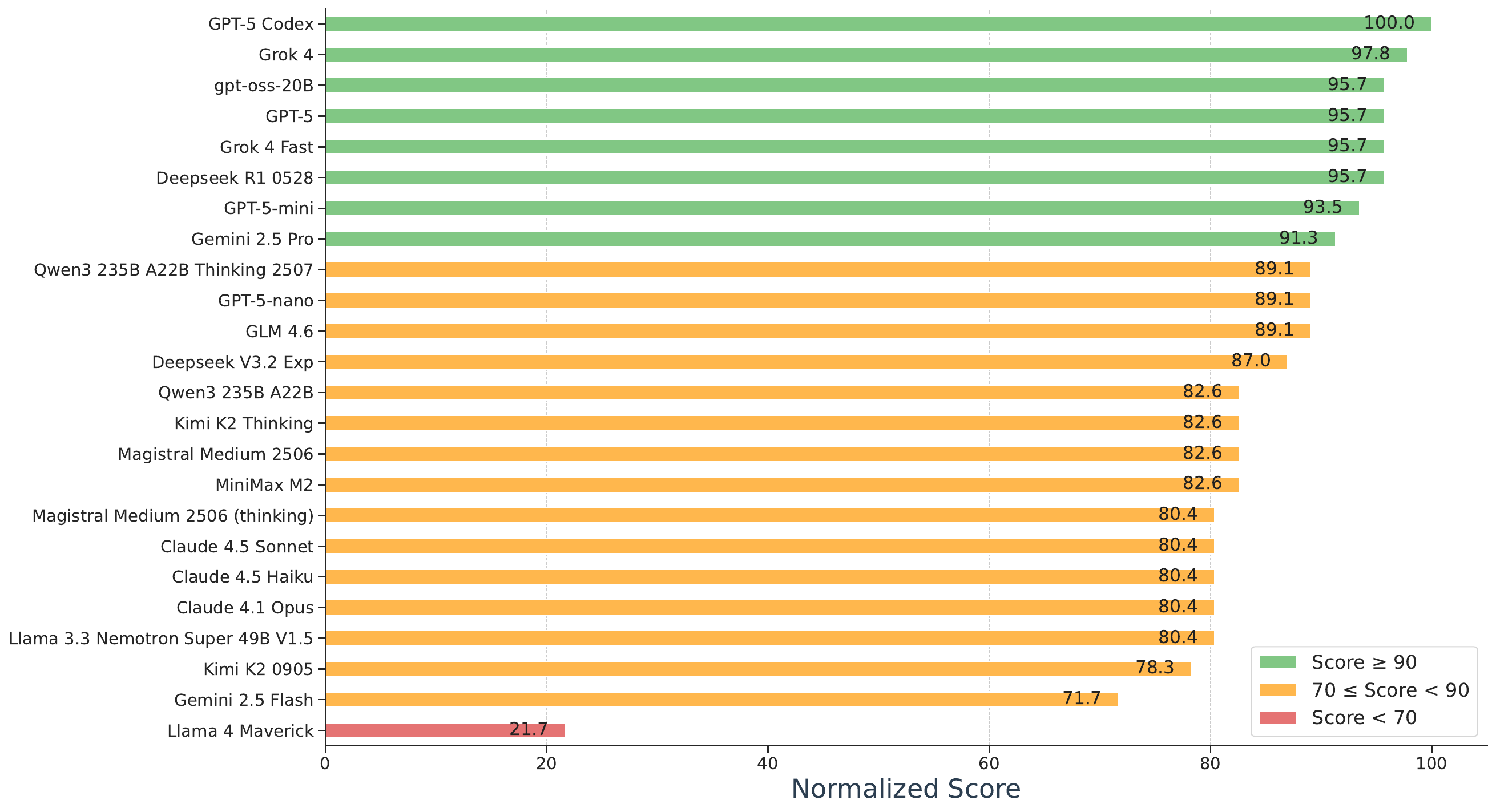}
    \caption{Model-wise performance comparison (input modality: Text-only, prompt language: Korean)}
    \label{fig:RQ0_leaderborad}
\end{figure}

As shown in Figure~\ref{fig:RQ0_leaderborad}, \textbf{GPT-5 Codex} is the only model that achieved a perfect normalized score of 100 in the main analysis.
Other GPT-5 family models (GPT-5: 95.7, GPT-5-mini: 93.5), Grok 4 models (Grok 4: 97.8, Grok 4 Fast: 95.7), and Gemini 2.5 Pro (91.3) also surpassed 90 points, demonstrating strong performance among recent proprietary models.  
Notably, \textbf{gpt-oss-20B achieves performance comparable to the top tier despite being a relatively lightweight open-weight model}.

Most of the remaining models scored in the 70--90 range.  
For example, Qwen3 235B A22B Thinking 2507 and GLM 4.6 achieved scores close to 90, while DeepSeek V3.2 Exp and Kimi K2 Thinking performed in the 80-point range.  
In contrast, the Claude family and Gemini 2.5 Flash underperformed relative to other proprietary models, and Llama 4 Maverick ranked last with a score of 21.7.  
These results indicate substantial variance in mathematical problem-solving capability across LLMs.

\paragraph{Model-wise latency analysis}

Human test-takers are given 100 minutes to solve 30 problems, whereas our setup requires LLMs to solve all 46 problems.  
To account for the increased number of questions, we scale the time budget to 152 minutes and evaluate models under both \textbf{unlimited-time} and \textbf{time-limited} settings.

\paragraph{(1) Unlimited-time performance}

\begin{figure}[!t]
    \centering
    \includegraphics[width=\linewidth]{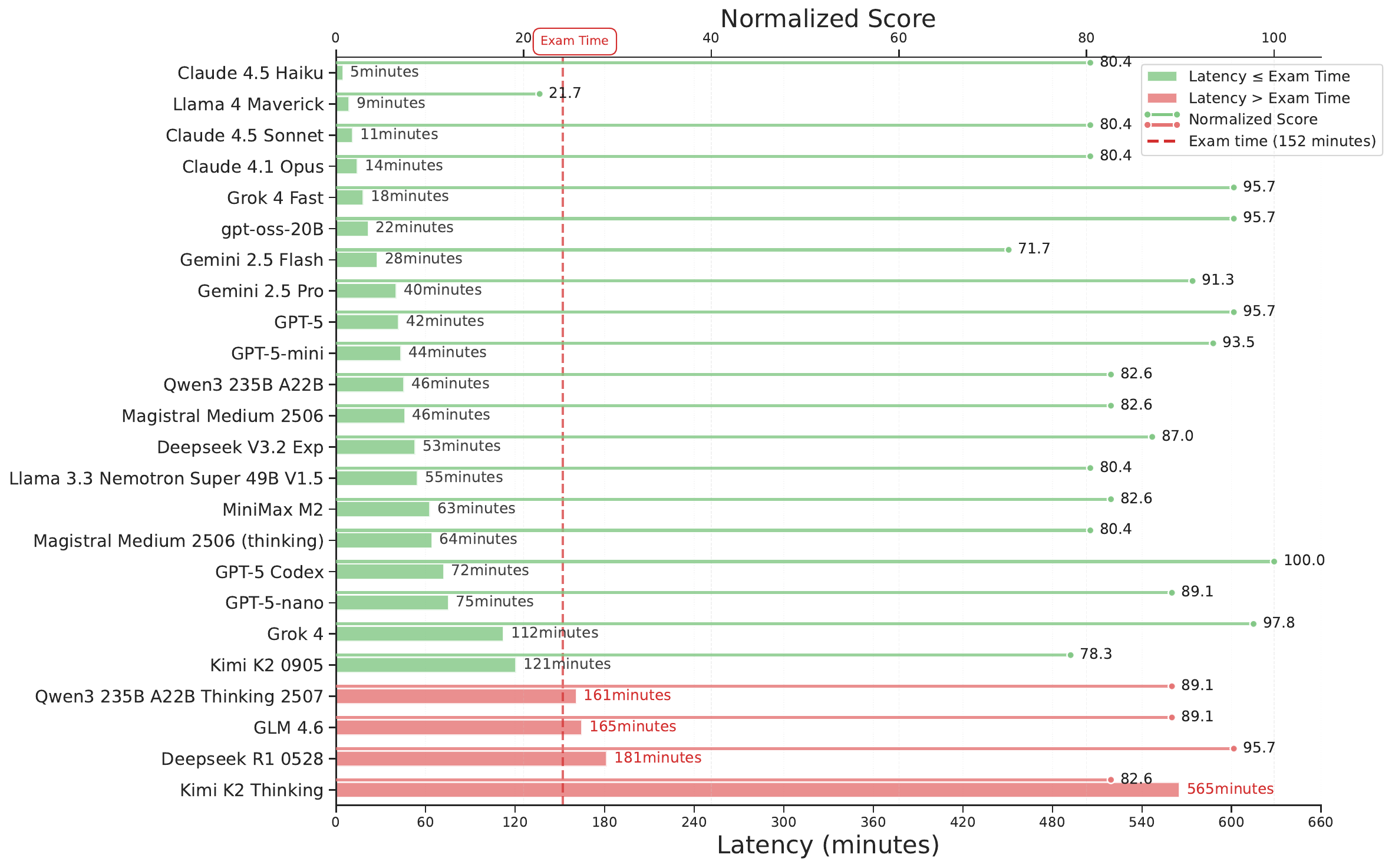}
    \caption{Latency vs. score by model (Text-only, Korean)}
    \label{fig:RQ0_timechart}
\end{figure}

Figure~\ref{fig:RQ0_timechart} plots model latency and performance under the Text-only, Korean prompt condition.  
A notable observation is that most models solved all 46 problems within 100 minutes, i.e., within the human time budget for only 30 problems.  
Even GPT-5 Codex, the top-performing model, completed all questions in 72 minutes.

Figure~\ref{fig:RQ0_timechart} shows the latency–score distribution under unlimited reasoning, Text-only input, and Korean prompts.

\begin{figure}[!t]
    \centering
    \includegraphics[width=\linewidth]{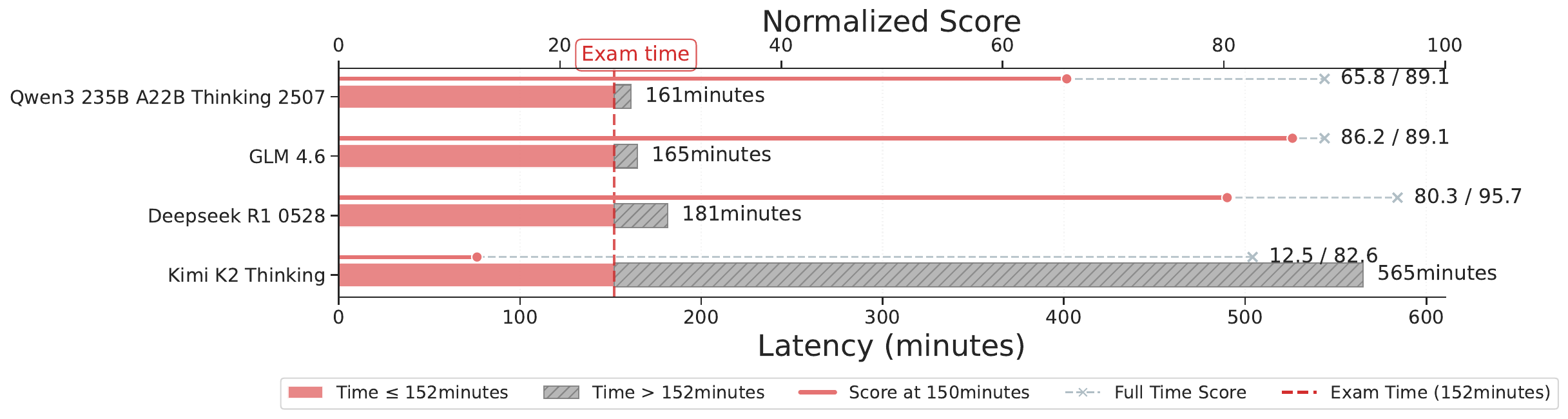}
    \caption{Time-limited performance of models exceeding the 152-minute budget}
    \label{fig:RQ0_timechart_over}
\end{figure}

\begin{table}[!t]
\centering
\caption{
Correlation analysis between latency and score.  
We report Pearson, Spearman, and Kendall correlations, as well as a simple OLS regression.  
All correlation coefficients are small and statistically insignificant ($p>0.05$).
}
\vspace{0.7em}
\label{tab:correlation_latency_score}
\begin{tabular}{lccc}
\toprule
\textbf{Method} & \textbf{Correlation} & \textbf{$p$-value} & \textbf{N} \\
\midrule
\midrule
Pearson & $0.123$ & $5.68\times 10^{-1}$ & 24 \\
Spearman & $0.291$ & $1.68\times 10^{-1}$ & 24 \\
Kendall~$\tau$ & $0.186$ & $2.19\times 10^{-1}$ & 24 \\
\midrule
Log-latency Pearson & $0.377$ & $6.98\times 10^{-2}$ & 24 \\
\midrule
OLS slope ($\beta_1$) & $0.016$ & $5.68\times 10^{-1}$ & 24 \\
OLS $R^2$ & $0.015$ & -- & 24 \\
\bottomrule
\end{tabular}
\end{table}

We further examine the relationship between latency and score using correlation tests.  
As summarized in Table~\ref{tab:correlation_latency_score}, Pearson, Spearman, and Kendall coefficients are all small ($|r|<0.30$) and not statistically significant ($p>0.05$).  
In other words, \textbf{we do not find evidence of a significant correlation between problem-solving speed and overall model performance}.

\paragraph{(2) Time-limited performance}

According to Figure~\ref{fig:RQ0_timechart}, four models—Qwen3 235B A22B Thinking 2507, GLM 4.6, DeepSeek R1 0528, and Kimi K2 Thinking—failed to complete all 46 problems within the 152-minute limit.  
These models are evaluated under both unlimited-time and time-limited settings; Figure~\ref{fig:RQ0_timechart_over} compares their scores under the two conditions.

When we only count answers produced within the 152-minute budget, all four models show reduced scores.  
Kimi K2 Thinking is most affected: its total solving time is 565 minutes (nearly 10 hours), and its score drops from 82.6 to 12.5 when late answers are excluded.  
Qwen3 235B A22B Thinking 2507 loses about 24 points when the last 9 minutes are cut, and GLM 4.6 loses about 3 points when we exclude its last 13 minutes of responses.

\begin{figure}[t]
    \centering
    \includegraphics[width=\linewidth]{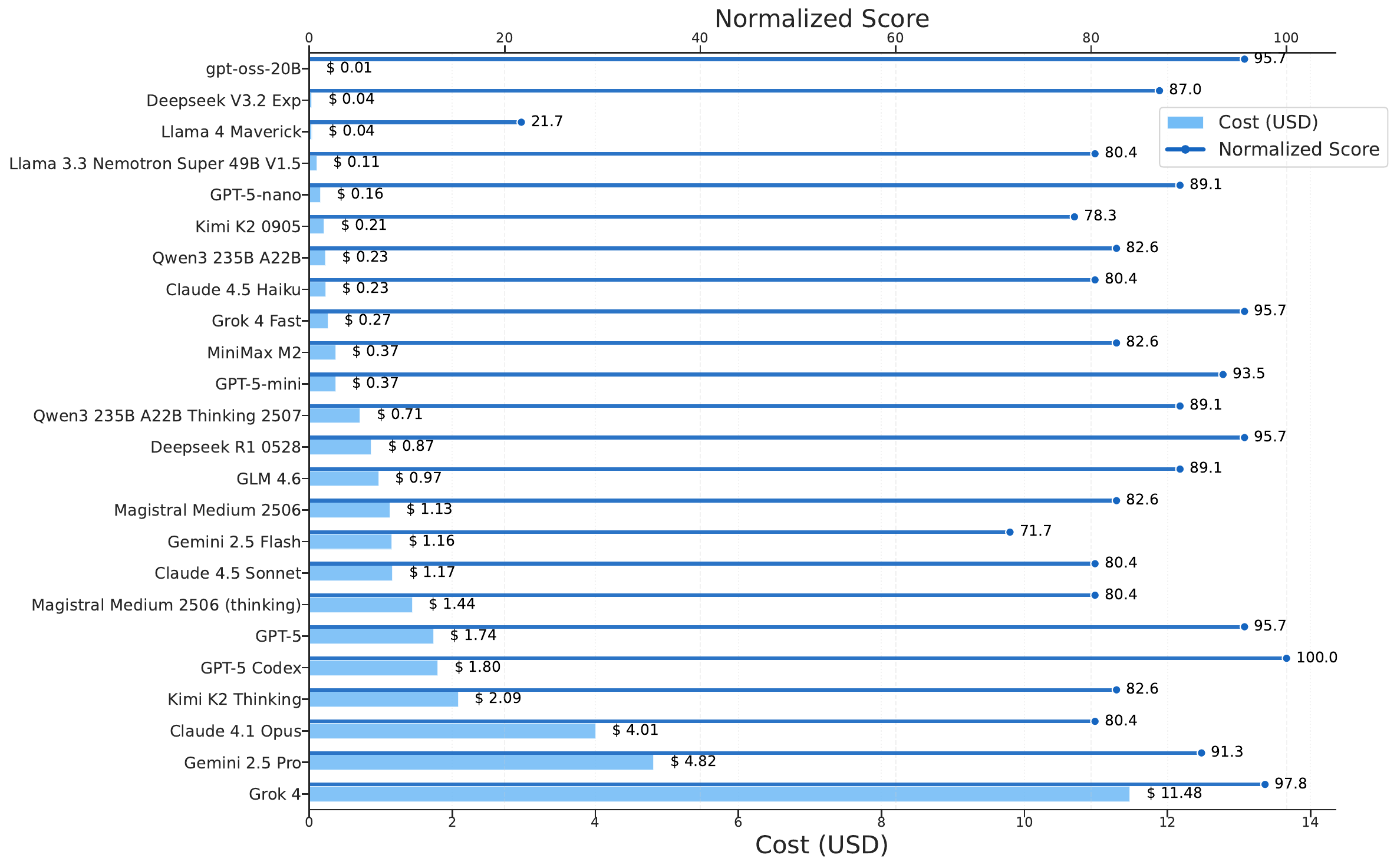}
    \caption{Cost vs. score by model (Text-only, Korean)}
    \label{fig:RQ0_costchart}
\end{figure}

\paragraph{Model-wise cost analysis}

Figure~\ref{fig:RQ0_costchart} shows the total API cost for each model to solve all 46 problems.  
Proprietary models incur relatively high costs: Grok 4 costs \$11.48, Gemini 2.5 Pro costs \$4.82, and Claude 4.1 Opus costs \$4.01.  
By contrast, open-weight models are much cheaper when accessed via OpenRouter: gpt-oss-20B costs \$0.01, DeepSeek V3.2 Exp and Llama 4 Maverick each cost \$0.04.  
Remarkably, gpt-oss-20B achieves a normalized score of 95.7 (3rd place) at the lowest cost (\$0.01), yielding an extremely favorable cost–performance ratio.

These results show that for some models, cost grows disproportionately relative to performance gains, while others achieve near–state-of-the-art scores at a fraction of the cost.

\begin{figure}[!t]
    \centering
    \begin{subfigure}[t]{\textwidth}
        \centering
        \includegraphics[width=\linewidth]{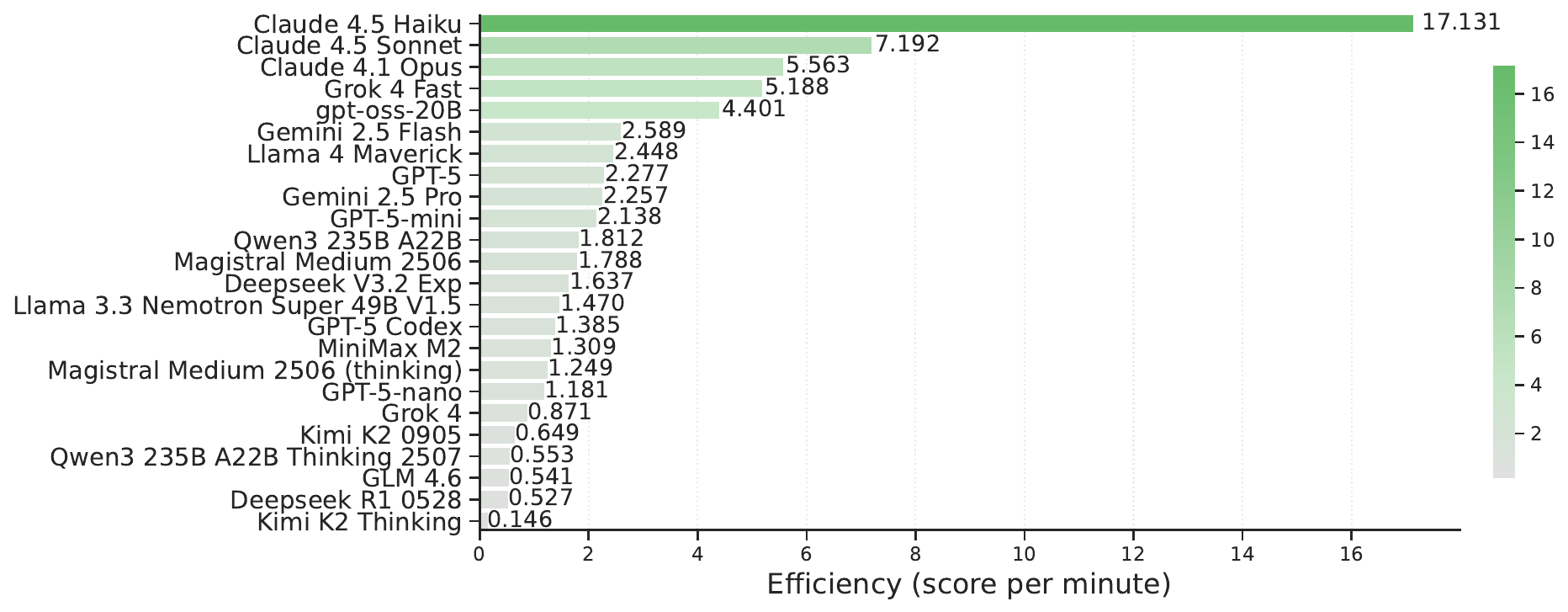}
        \caption{Time-based Efficiency Score ($\text{Eff}_t$) across models (Text-only, Korean)}
        \label{fig:efficiency_time}
    \end{subfigure}
    
    \begin{subfigure}[t]{\textwidth}
        \centering
        \includegraphics[width=\linewidth]{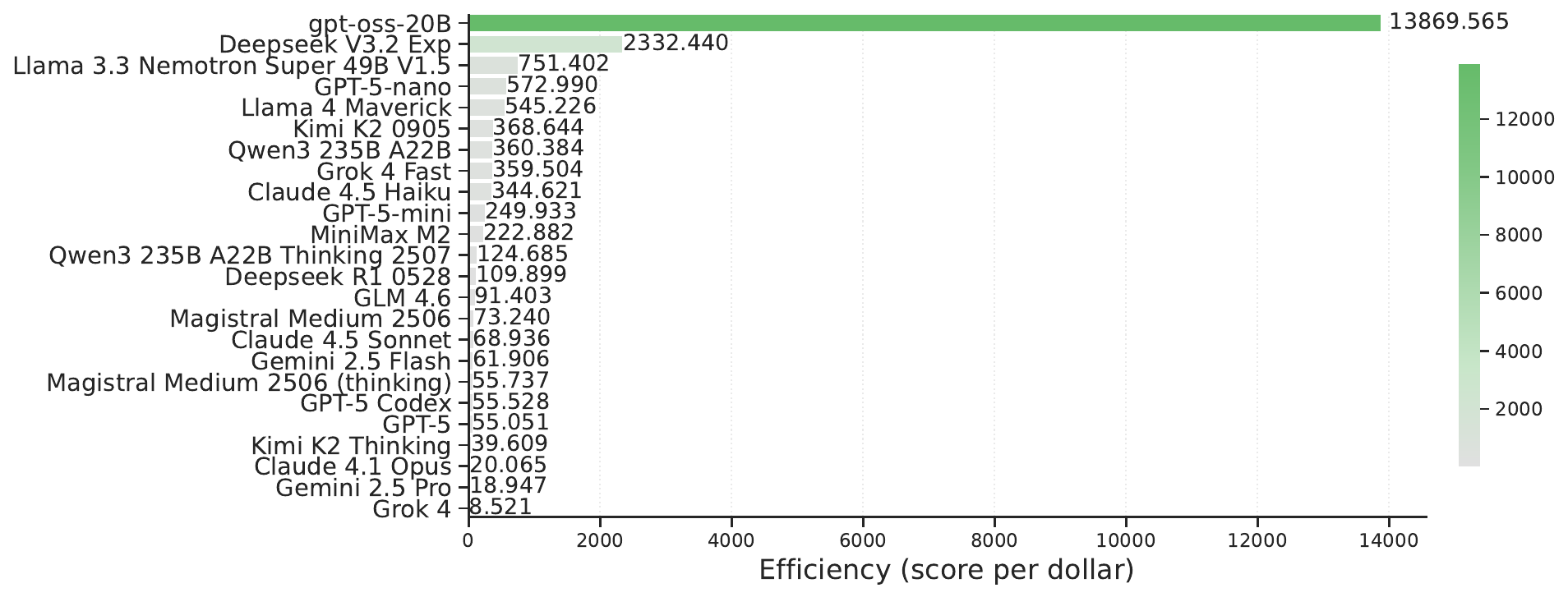}
        \caption{Cost-based Efficiency Score ($\text{Eff}_c$) across models (Text-only, Korean)}
        \label{fig:efficiency_cost}
    \end{subfigure}
    
    \begin{subfigure}[t]{\textwidth}
        \centering
        \includegraphics[width=\linewidth]{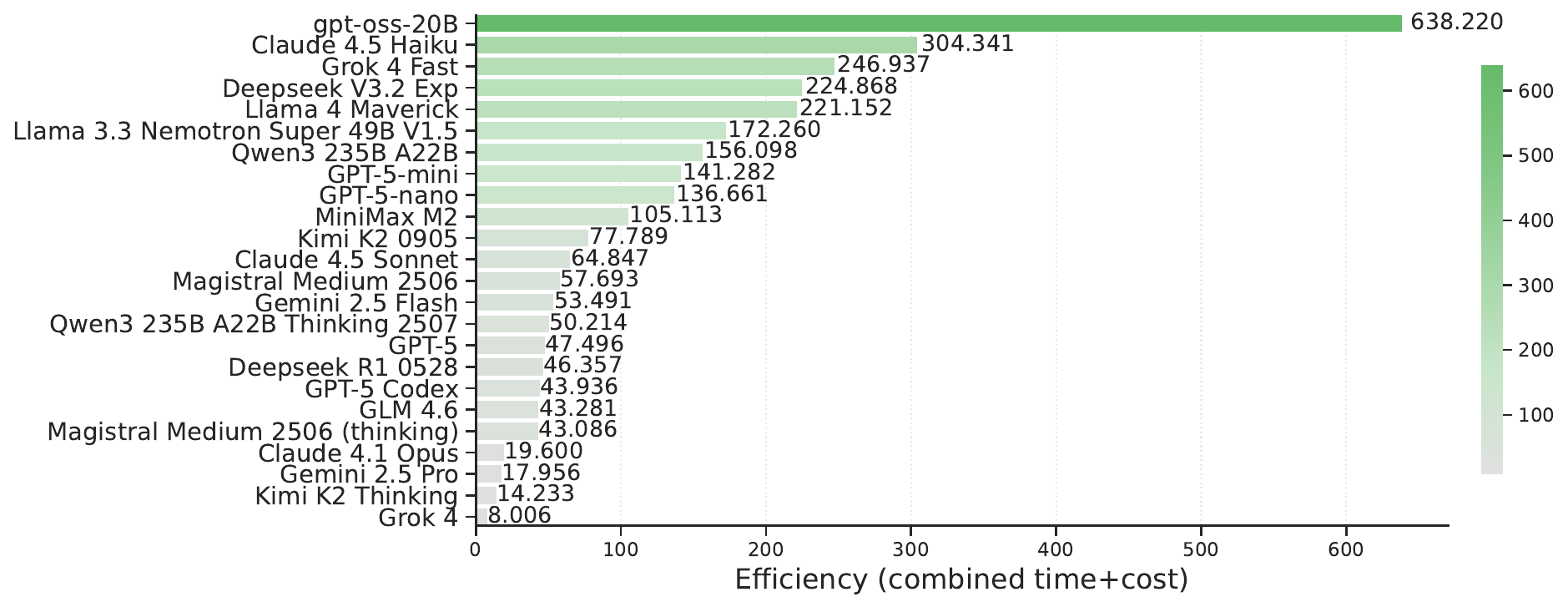}
        \caption{Joint time–cost Efficiency Score ($\text{Eff}_{t,c}$) across models (Text-only, Korean)}
        \label{fig:efficiency_time_cost}
    \end{subfigure}
    \caption{Efficiency comparison across performance, time, and cost (Text-only, Korean)}
    \label{fig:efficiency}
\end{figure}

\paragraph{Model-wise efficiency analysis}

We define three efficiency scores to jointly analyze performance against time and cost:

\[
\begin{aligned}
(1)\;& \text{Eff}_{t} = \frac{\text{score}}{\text{latency (min)}}
\qquad
(2)\;& \text{Eff}_{c} = \frac{\text{score}}{\text{cost (\$)}}
\qquad
(3)\;& \text{Eff}_{t,c} = \frac{\text{score}}{\dfrac{\text{latency}}{152\ \text{min}}+\dfrac{\text{cost}}{1\ \text{\$}}}
\end{aligned}
\]

Here, $\text{Eff}_{t}$ in (1) captures \emph{time efficiency}: higher values indicate that a model obtains a higher score in less time. $\text{Eff}_{c}$ in (2) captures \emph{cost efficiency}: given the same performance, models with lower cost have higher $\text{Eff}_{c}$. Finally, $\text{Eff}_{t,c}$ in (3) is a \emph{joint time–cost efficiency} measure.  
A high $\text{Eff}_{t,c}$ value indicates that a model achieves high performance while consuming relatively little time and monetary cost.

Figure~\ref{fig:efficiency_time} visualizes \textbf{time-normalized performance} using $\text{Eff}_t$.  
Claude 4.5 Haiku achieves the highest $\text{Eff}_t$, solving all 46 problems in 5 minutes with a score of 80.  
Other Claude variants rank second and third, followed by Grok 4 Fast and gpt-oss-20B, indicating strong time efficiency.

Figure~\ref{fig:efficiency_cost} shows \textbf{cost-normalized performance} via $\text{Eff}_c$.  
gpt-oss-20B achieves a normalized score of 95.7 (3rd) at a cost of \$0.01 (1st), resulting in by far the highest $\text{Eff}_c$.  
DeepSeek V3.2 Exp ranks second, while Grok 4, the most expensive model (\$11.48), ranks last on this metric.

Figure~\ref{fig:efficiency_time_cost} visualizes $\text{Eff}_{t,c}$, capturing \textbf{overall performance per combined time and monetary resource}.  
Under this joint metric, gpt-oss-20B is the best model.  
Claude 4.5 Haiku ranks second: despite its relatively lower cost efficiency, its exceptional time efficiency boosts its overall $\text{Eff}_{t,c}$.

Full results for all input combinations are reported in Appendix~\ref{sec:appendix_efficiency_score}.

In summary, models that are optimal in terms of time efficiency differ from those that are optimal in terms of cost efficiency.  
\textbf{If users prioritize fast responses, or if they need multiple repeated runs, Claude 4.5 Haiku and Claude 4.5 Sonnet are attractive choices; if they prioritize low cost, gpt-oss-20B and DeepSeek V3.2 Exp offer highly favorable performance–resource trade-offs.}

\subsection{LLM Performance Under Varying Input and Prompt Conditions}

\subsubsection{Performance Differences by Model Parameter Size}
\label{sec:parameter_size}

\paragraph{Performance by Model Size}

\begin{figure}
\vspace{-1em}
    \centering
    \begin{subfigure}{0.45\linewidth}
        \centering
        \includegraphics[width=\textwidth]{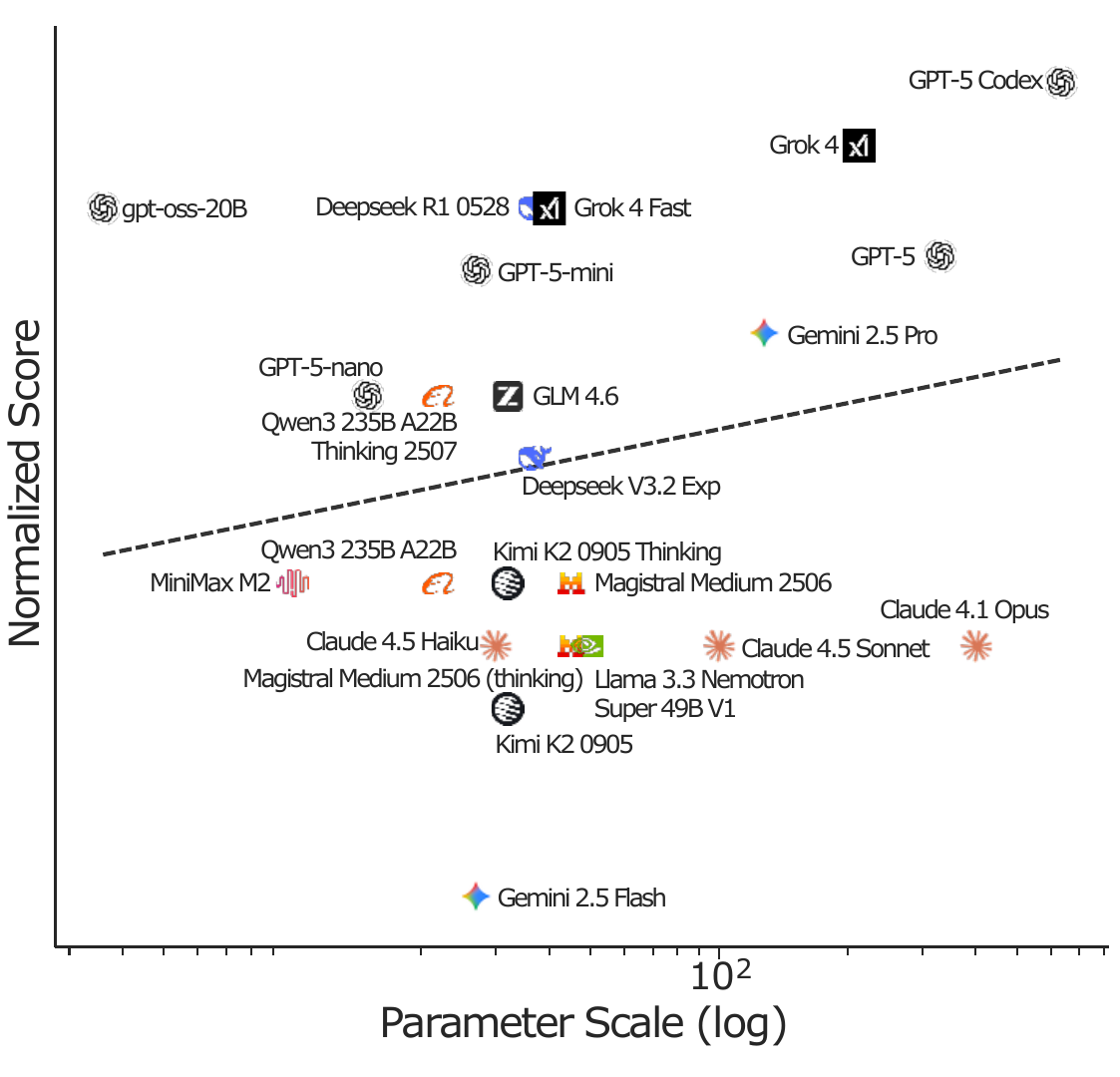}
        \subcaption{Performance by model size. Llama 4 Maverick (Normalized Score: 21.7) is omitted for legibility.}
        \label{fig:parameter_scatter1}
    \end{subfigure}
    \begin{subfigure}{0.45\linewidth}
        \centering
        \includegraphics[width=\textwidth]{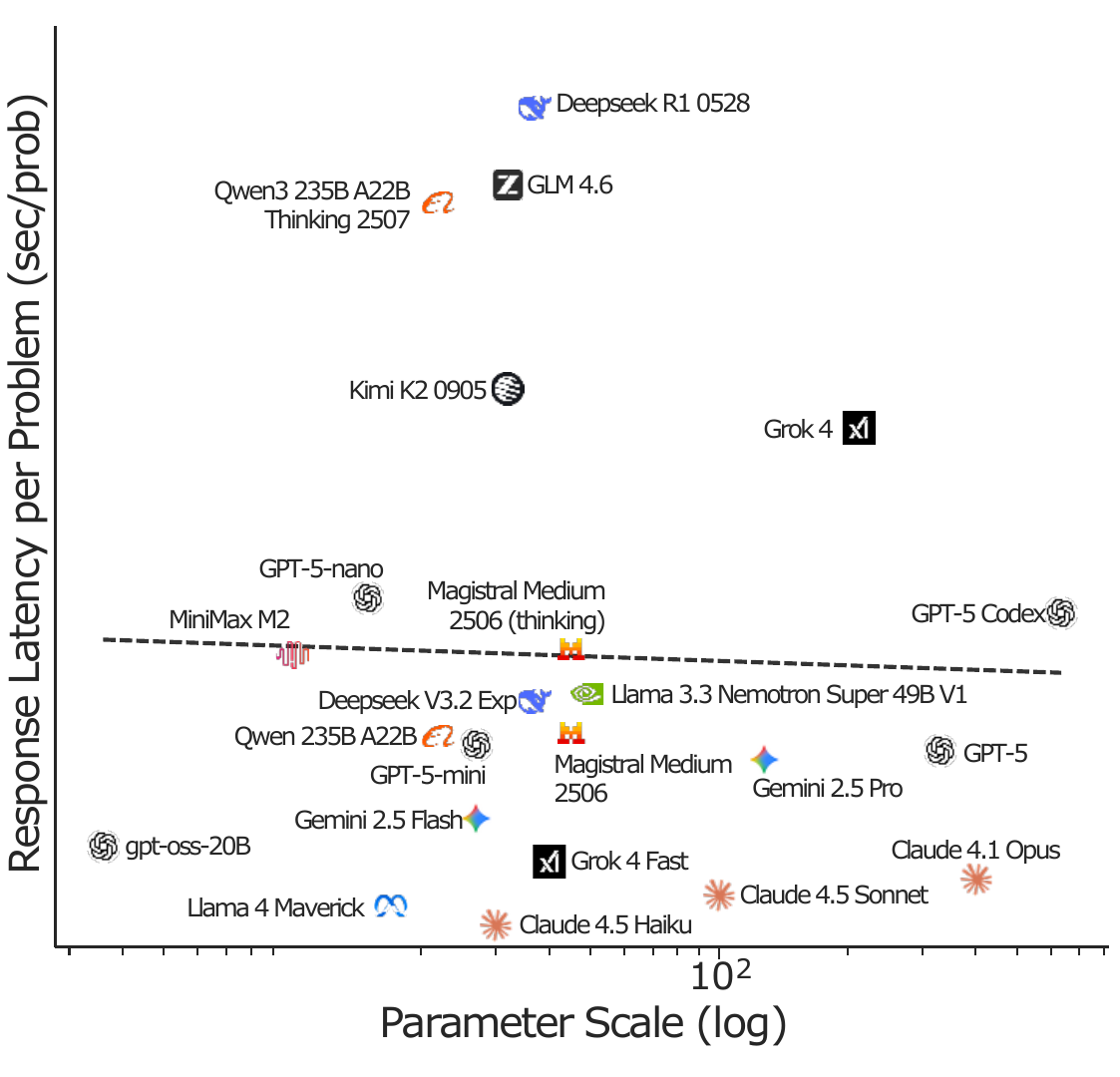}
        \subcaption{Latency by model size. Kimi K2 Thinking (737 sec/problem) is omitted for legibility.}
        \label{fig:parameter_scatter2}
    \end{subfigure}
    \caption{Model performance and per-problem latency by parameter size (Text-only, Korean)}
    \label{fig:parameter_scatter}
\end{figure}

Figure~\ref{fig:parameter_scatter} illustrates how model parameter size relates to performance and per-problem latency.  
As shown in Figure~\ref{fig:parameter_scatter1}, larger models generally achieve higher scores, but this trend is far from strictly linear.  
For example, gpt-oss-20B contains roughly seven times fewer active parameters than Qwen3 235B A22B, yet it outperforms the larger model.  
This demonstrates that parameter count alone is not a definitive determinant of math problem-solving ability.

\paragraph{Relationship Between Model Size and Latency}

Figure~\ref{fig:parameter_scatter2} shows that some large models maintain relatively short latencies despite their size, whereas several mid-sized models show extremely long per-problem processing times.  
This indicates that parameter size produces a complex trade-off structure involving accuracy, latency, and efficiency; no single dimension is sufficient for fully characterizing model performance.

Full results for all modality–language combinations are presented in Appendix~\ref{sec:appendix_size_perf}.

In summary, although larger models generally show stronger performance, \textbf{model size is not an absolute predictor of accuracy, and certain smaller models (e.g., gpt-oss-20B) outperform larger models in both accuracy and efficiency.}

\begin{figure}[!t]
\centering
    \begin{subfigure}[t]{\linewidth}
        \centering
        \includegraphics[width=\textwidth]{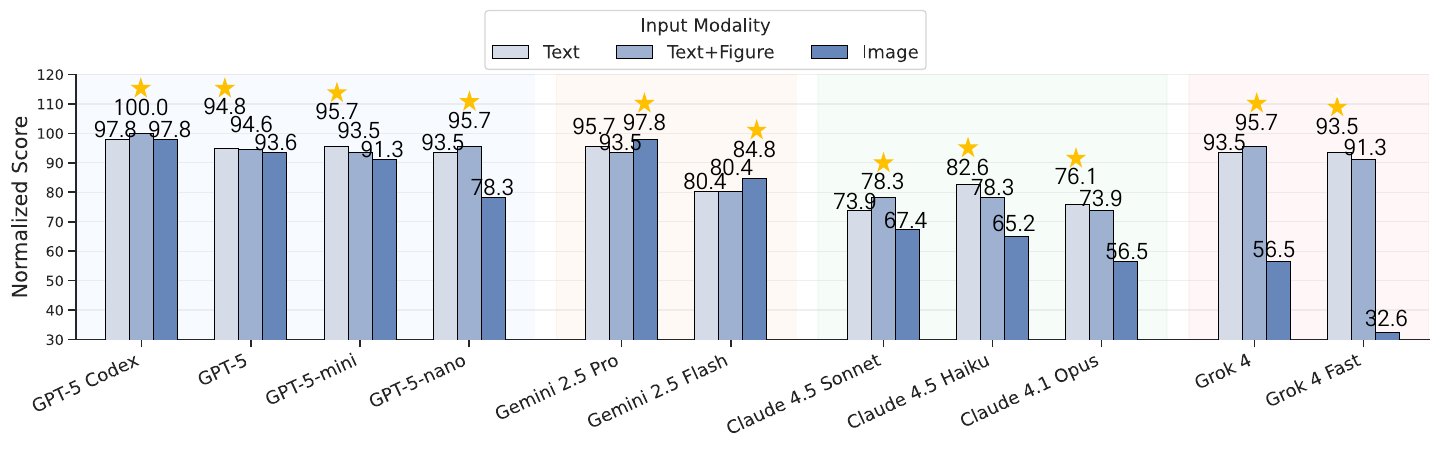}
        \caption{Model-family performance by input modality. Modality with highest score per model is marked with a star.}
        \label{fig:input_modality_perf}
    \end{subfigure}
    \vspace{0.7em}
    \begin{subfigure}[t]{\linewidth}
        \centering
        \includegraphics[width=0.63\textwidth]{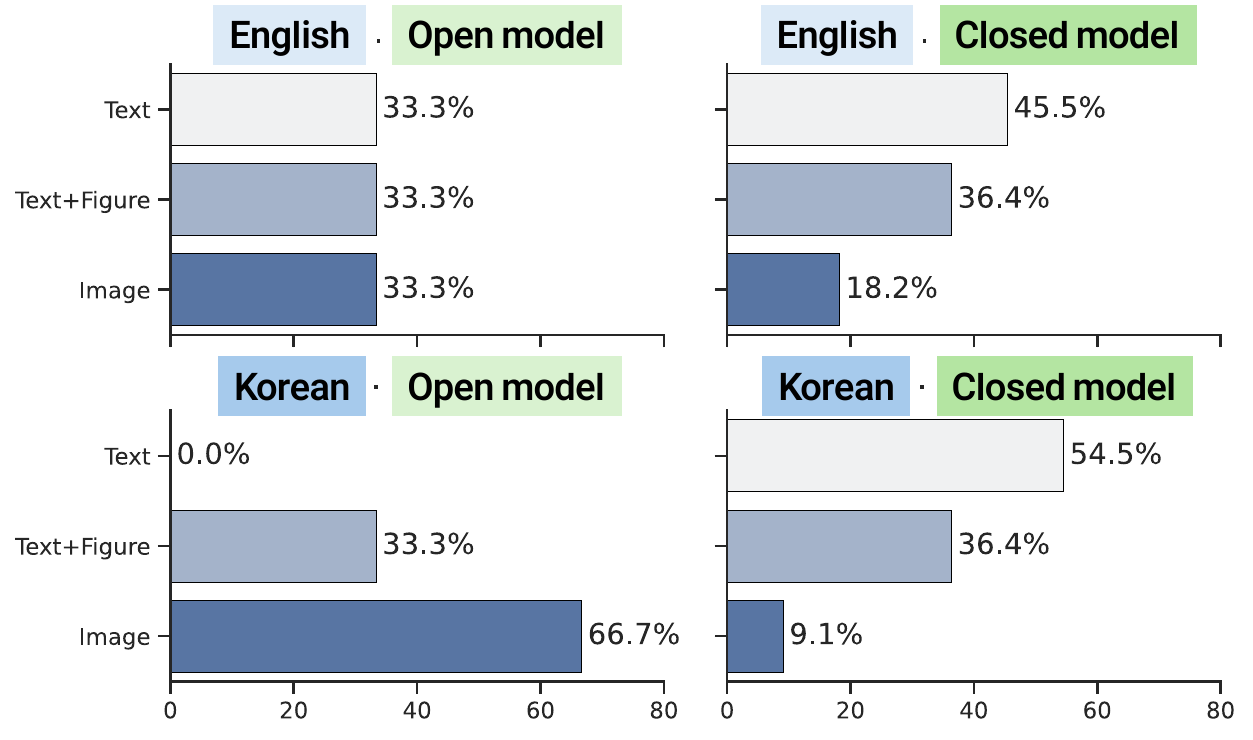}
        \caption{Win rate by input modality, prompt language, and weight accessibility}
        \label{fig:win_rate_modality}
    \end{subfigure}
    \caption{Performance and win rate comparison across input modalities}
\end{figure}

\subsubsection{Performance Differences by Input Modality}

Figure~\ref{fig:input_modality_perf} visualizes how input modality affects proprietary models.  
Key observations are as follows:

\begin{itemize}
\item \textbf{GPT-5 family}: Achieves the highest scores under Text-only or Text+Figure. GPT-5 Codex reaches \textbf{100 points} with Text+Figure. Image-only input lowers performance.
\item \textbf{Claude family}: Image-only input sharply degrades performance across all variants; Text and Text+Figure perform similarly.
\item \textbf{Grok family}: Similar to Claude—Image-only performs worst; Text and Text+Figure perform better.
\item \textbf{Gemini family}: Displays the opposite trend, with \textbf{Image-only outperforming other modalities}, indicating strong visual-processing capability.
\end{itemize}

Figure~\ref{fig:win_rate_modality} shows win rates by input modality, separated by prompt language and weight accessibility.  
As in Figure~\ref{fig:input_modality_perf}, proprietary models prefer Text and Text+Figure, with lower win rates for Image-only.  
Open-weight models show slightly different behavior: with Korean prompts, Image input yields a near-zero win rate, but this is based on only three image-capable open-weight models (Qwen3 235B A22B, Qwen3 235B A22B Thinking 2507, Llama 4 Maverick), limiting interpretability.

\subsubsection{Performance Differences by Prompting Language}

\begin{figure}[!t]
\centering
    \begin{subfigure}[t]{\linewidth}
        \centering
        \includegraphics[width=\textwidth]{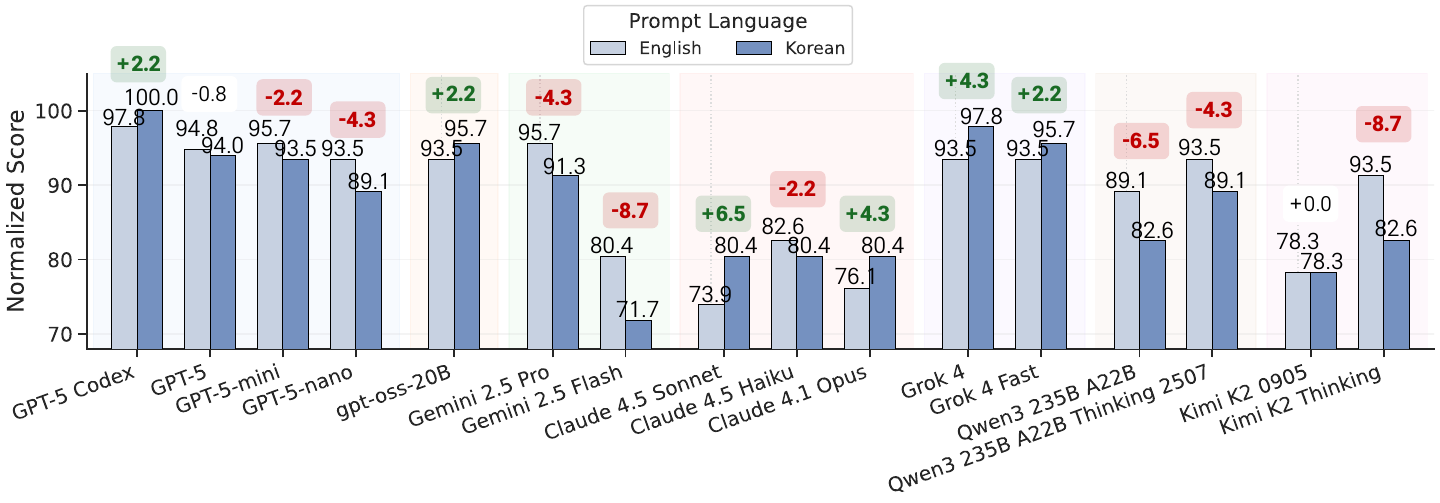}
        \caption{Model-family performance by prompting language. Green = Korean better, Red = English better, White = difference < 2 points.}
        \label{fig:language_perf}
    \end{subfigure}
    \vspace{2em}
    \begin{subfigure}[t]{\linewidth}
        \centering
        \includegraphics[width=0.65\textwidth]{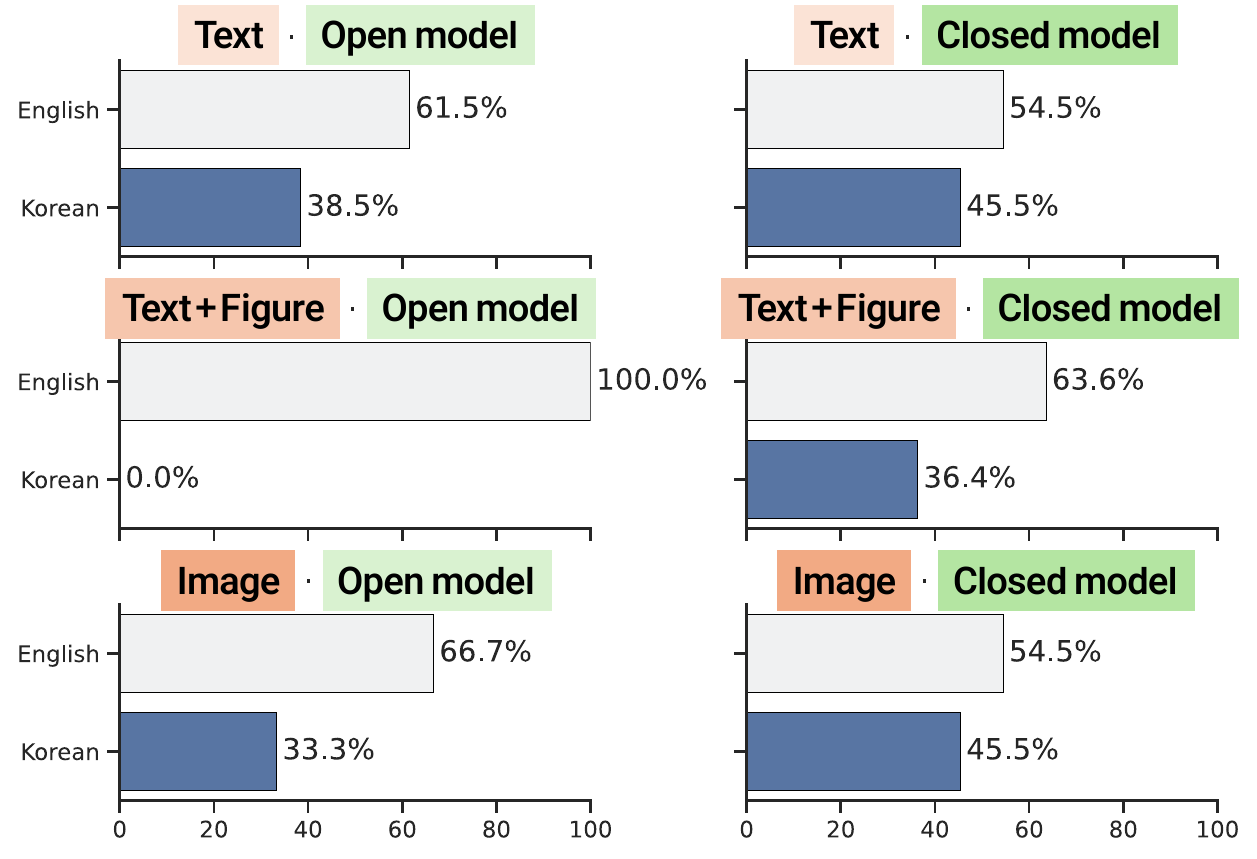}
        \caption{Win rate by prompting language across modalities and accessibility}
        \label{fig:win_rate_lang}
    \end{subfigure}
    \caption{Performance and win rate comparison across prompting languages}
\end{figure}

Figure~\ref{fig:language_perf} shows performance differences between Korean and English prompts across model families.  
GPT-5 Codex and gpt-oss-20B perform better with Korean prompts, while Claude and Grok models also show an advantage with Korean input.  
In contrast, Gemini models and major open-weight models (Qwen, Kimi, etc.) exhibit substantially higher performance with English prompts; Korean prompts lead to significant drops in accuracy.

Figure~\ref{fig:win_rate_lang} presents win rates by prompting language grouped by input modality and weight accessibility.  
Across all conditions, \textbf{English prompts show a clear advantage}.  
Even among Text-only inputs, open-weight models benefit more from English prompts than proprietary models.  
Although results for image-based prompts are influenced by the small number of image-capable open-weight models, English prompts consistently help across modalities.

\subsection{Analysis of LLM Performance by Mathematical Problem Characteristics}

\subsubsection{Performance Differences by Problem Category}
\label{sec:paper_area}

\begin{figure}[!t]
    \centering
    \includegraphics[width=\linewidth]{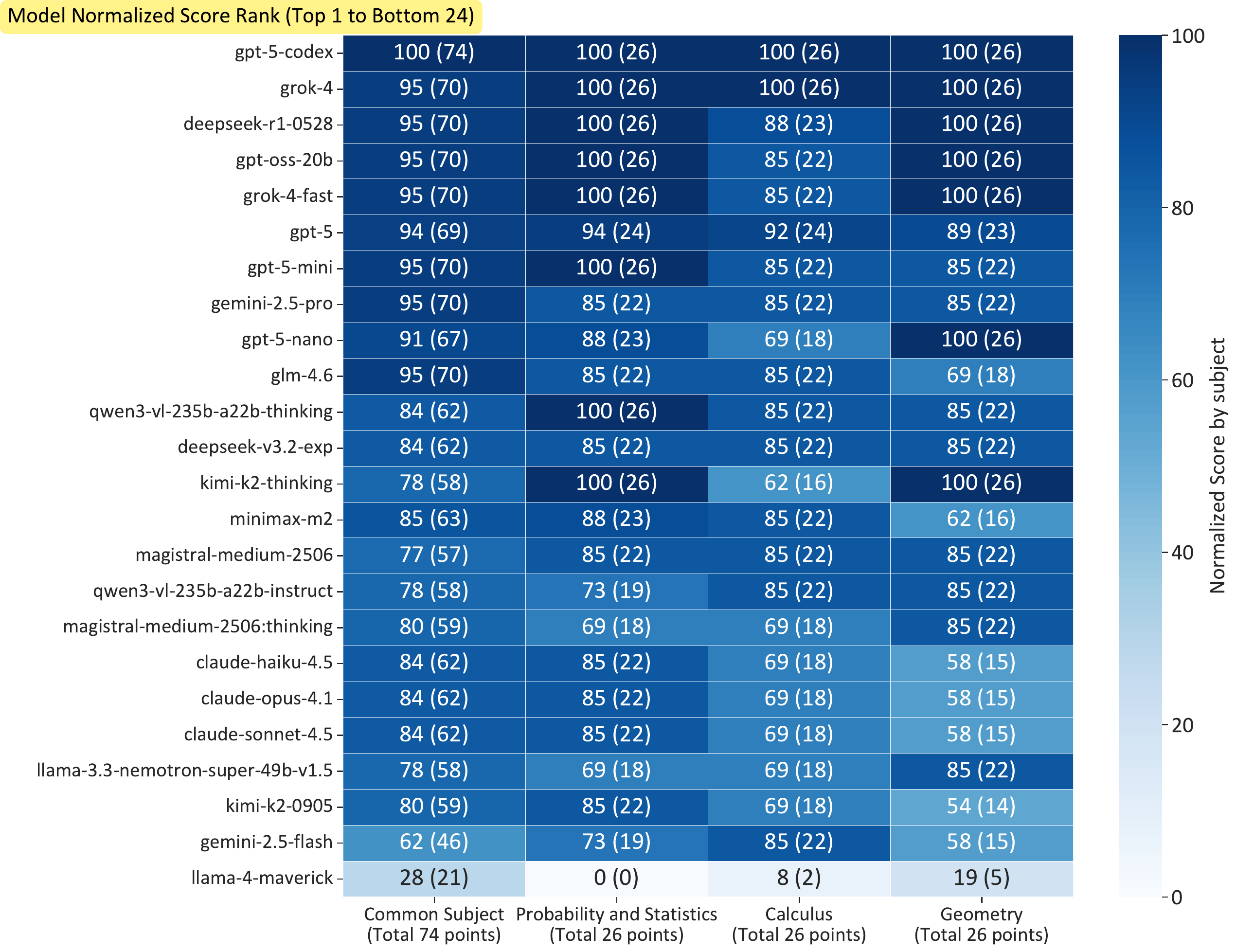}
    \caption{
    LLM performance by problem category.  
    Normalized scores represent each model’s score divided by the maximum for that category.  
    Models are sorted by overall Normalized Score.
    }
    \label{fig:RQ2_problem_area_scre}
\end{figure}

Figure~\ref{fig:RQ2_problem_area_scre} presents LLM performance by subject category.  
The labeled values represent both normalized scores and raw scores (score divided by category maximum).

Models demonstrate their weakest performance in Calculus, with Geometry ranking as the second most challenging category.
The difficulty of Calculus stems from its demand for complex symbolic manipulation and multi-step logical reasoning, whereas Geometry introduces additional variability across models due to its reliance on spatial interpretation and visual understanding.

When averaging across models, performance was lowest in Calculus (77.7pts / 20.2score), followed by Geometry (79.2pts / 20.6score), Common Math (83.9pts / 22.1score), and Probability \& Statistics (84.7pts / 22.0score).

Full results across all modality–language combinations appear in Appendix~\ref{sec:appendix_area_ttributes}.

% Most models perform worst in Geometry, while top-tier models sometimes show lower accuracy in Calculus instead.  
% Geometry also exhibits the largest variance across models, likely because geometric reasoning requires spatial interpretation and visual understanding.  
% In contrast, top-performing models struggle more with Calculus due to the heavier logical reasoning burden.  

% Full results across all modality–language combinations appear in Appendix~\ref{sec:appendix_area_ttributes}.

\subsubsection{Performance Differences by Problem Attributes}
\label{sec:paper_attribute}

\paragraph{Differences by Problem Type (Multiple-choice vs. Short-answer)}

\begin{figure}[!t]
    \centering
    \begin{subfigure}[t]{0.48\linewidth}
        \centering
        \includegraphics[width=\linewidth]{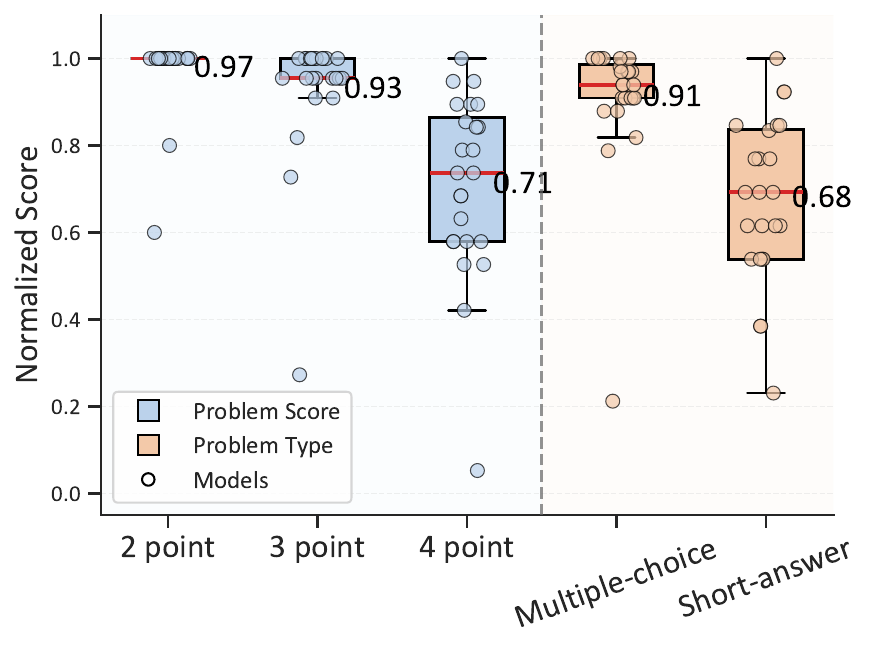}
        \caption{Performance distribution by score (2/3/4 points) and problem type (multiple-choice/short-answer).}
        \label{fig:RQ2_dist}
    \end{subfigure}
    \hfill
    \begin{subfigure}[t]{0.48\linewidth}
        \centering
        \includegraphics[width=\linewidth]{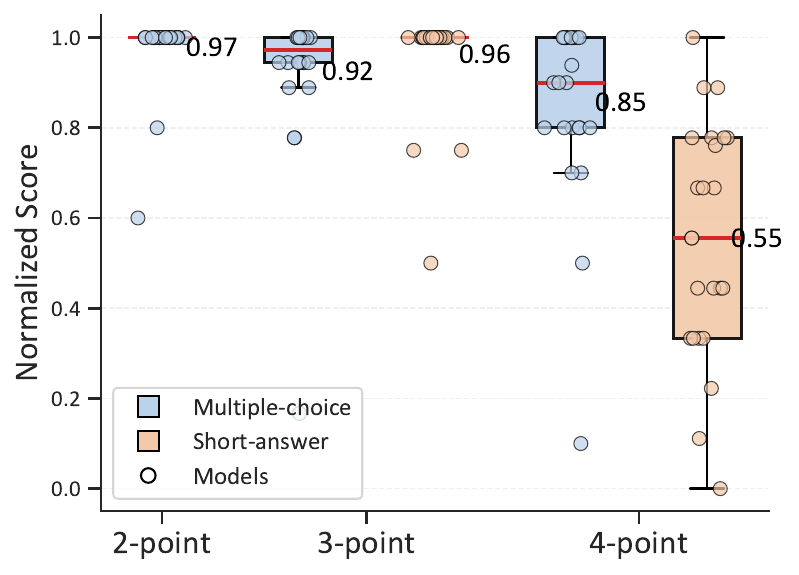}
        \caption{Performance by score–type combinations.  
        (No 2-point multiple-choice problems in 2026 CSAT.)}
        \label{fig:RQ2_dist2}
    \end{subfigure}

    \caption{
    LLM performance by problem characteristics.  
    (a) examines score and type independently;  
    (b) analyzes joint score–type combinations.
    }
    \label{fig:RQ2_combined}
\end{figure}

Figure~\ref{fig:RQ2_dist} compares performance by problem type.  
LLMs perform substantially better on multiple-choice questions: the average Normalized Score is 91 for five-option problems, compared to 68 for short-answer problems.

However, this difference cannot be attributed solely to format.  
In CSAT mathematics, the most difficult problems (4-point problems) are overwhelmingly short-answer, meaning problem type and difficulty are deeply intertwined.

\paragraph{Differences by Problem Score (Difficulty Level)}

A consistent decline in accuracy is observed as problem difficulty increases.  
Normalized scores for 2- and 3-point questions range from 93 to 97, while 4-point questions drop to 71.  
This suggests that the human-designed difficulty scale of CSAT mathematics \textbf{transfers naturally to LLMs}, making the hardest questions difficult for both humans and models.

Figure~\ref{fig:RQ2_dist2} further breaks down performance by the intersection of type (multiple-choice/short-answer) and score (2/3/4).  
Performance is high for all 2–3 point combinations, but all 4-point combinations—multiple-choice or short-answer—show large accuracy drops and increased variance across models.

Overall, LLMs perform well on easier combinations (2–3 points, multiple-choice), but \textbf{struggle significantly on harder combinations, especially 4-point short-answer questions}.  
Complete results for all modality–language combinations are detailed in Appendix~\ref{sec:appendix_item_attributes}.

\subsection{Effectiveness of Reasoning-Oriented Inference Compared to Baseline LLM Behavior}

\subsubsection{Impact of Reasoning Configuration on Model Performance}

\paragraph{Contributions of \texttt{Reasoning\_Effort} and \texttt{Text\_Verbosity}}

\begin{figure}[!t]
    \centering
    \begin{subfigure}[t]{0.49\linewidth}
        \centering
        \includegraphics[width=\linewidth]{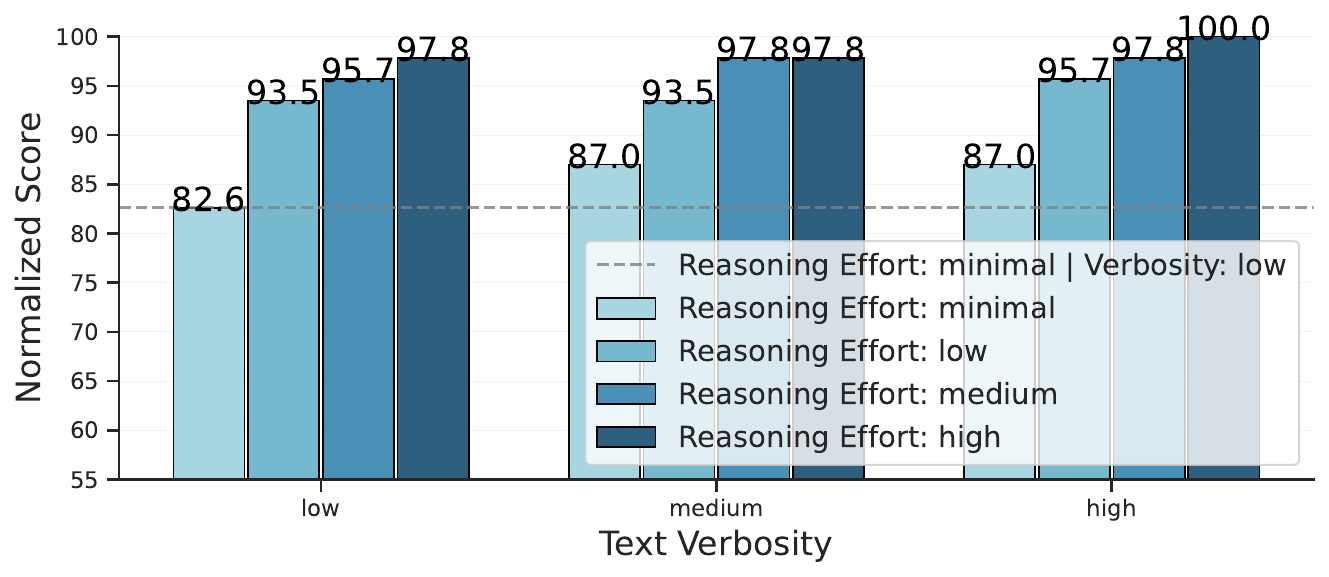}
        \caption{Performance variation across \texttt{Reasoning\_Effort} and \texttt{Text\_Verbosity} combinations (GPT-5 family)}
        \label{fig:RQ3_score}
    \end{subfigure}
    \hfill
    \begin{subfigure}[t]{0.49\linewidth}
        \centering
        \includegraphics[width=\linewidth]{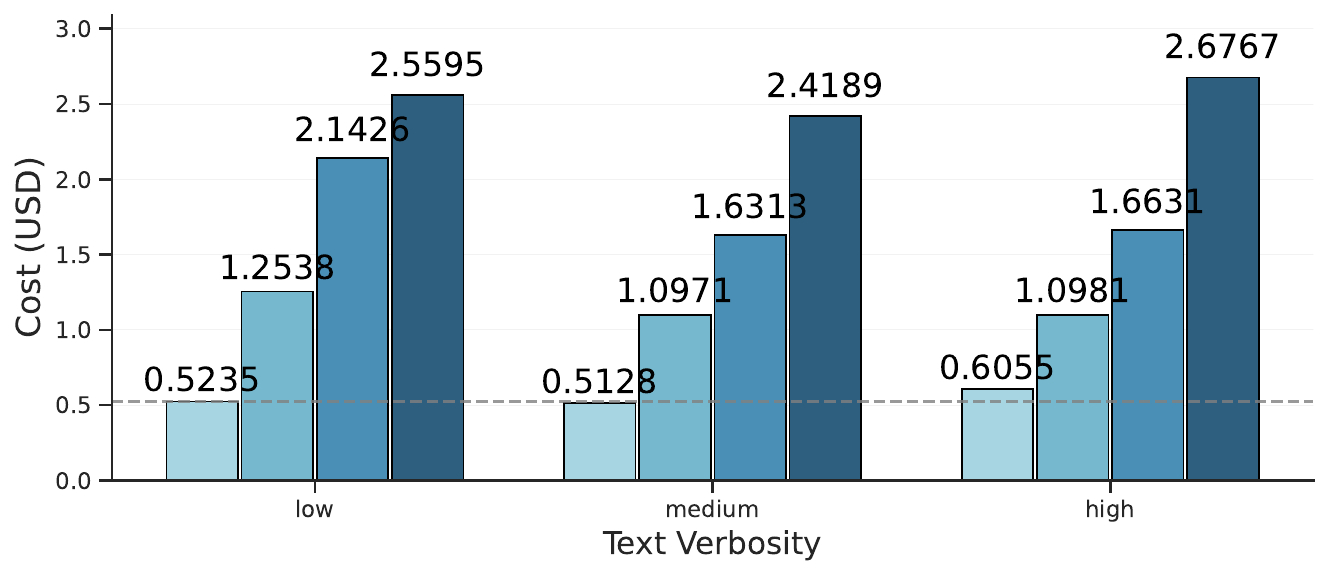}
        \caption{Cost variation across \texttt{Reasoning\_Effort} and \texttt{Text\_Verbosity} combinations}
        \label{fig:RQ3_score2}
    \end{subfigure}
    \caption{Comparison of \texttt{Reasoning\_Effort} and \texttt{Text\_Verbosity} in GPT-5 configuration}
    \label{fig:RQ3_combined}
\end{figure}

Figure~\ref{fig:RQ3_score} shows how different combinations of \texttt{Reasoning\_Effort} and \texttt{Text\_Verbosity} affect GPT-5 performance.  
Increasing \texttt{Reasoning\_Effort} consistently improves accuracy, whereas increasing \texttt{Text\_Verbosity} produces almost no performance change.

This aligns with the definitions provided in the GPT-5 documentation.\footnote{\url{https://platform.openai.com/docs/guides/latest-model}}  
\texttt{Text\_Verbosity} controls the number of output tokens—its effect is limited to explanation length and does not alter internal reasoning.  
In contrast, \texttt{Reasoning\_Effort} explicitly controls the number of internal reasoning tokens generated prior to producing an answer, thus directly affecting the model’s internal inference process.

Therefore, the observed performance gains primarily reflect increased reasoning effort, not verbosity.

\begin{figure}[!t]
    \centering
    \begin{subfigure}[t]{\textwidth}
        \centering
        \includegraphics[width=\linewidth]{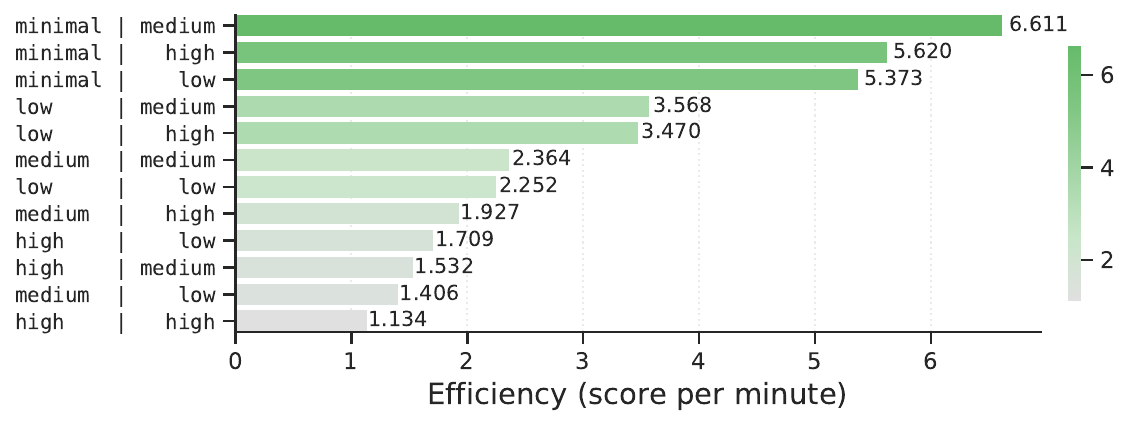}
        \caption{Performance–time efficiency under GPT-5 reasoning configurations}
        \label{fig:gpt5_reasoning_efficiency_time}
    \end{subfigure}
    
    \begin{subfigure}[t]{\textwidth}
        \centering
        \includegraphics[width=\linewidth]{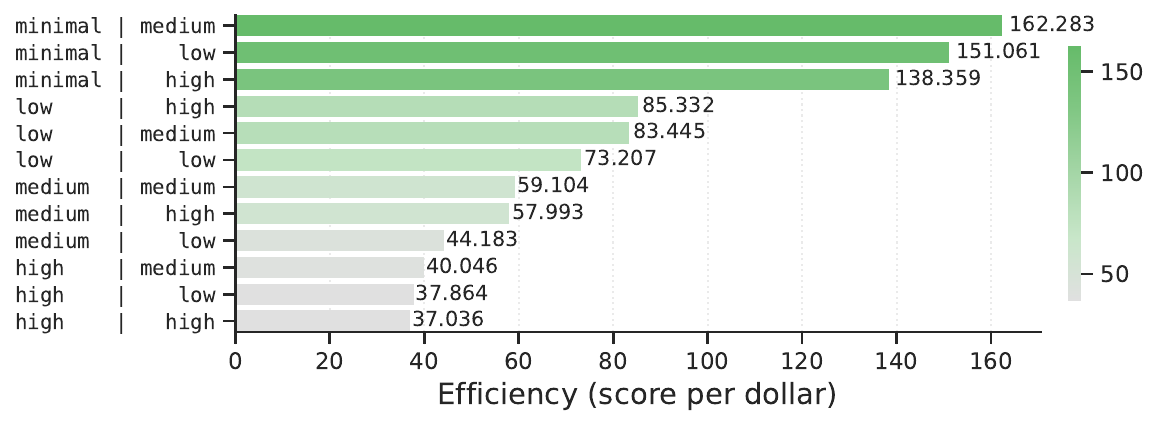}
        \caption{Performance–cost efficiency under GPT-5 reasoning configurations}
        \label{fig:gpt5_reasoning_efficiency_cost}
    \end{subfigure}
    
    \begin{subfigure}[t]{\textwidth}
        \centering
        \includegraphics[width=\linewidth]{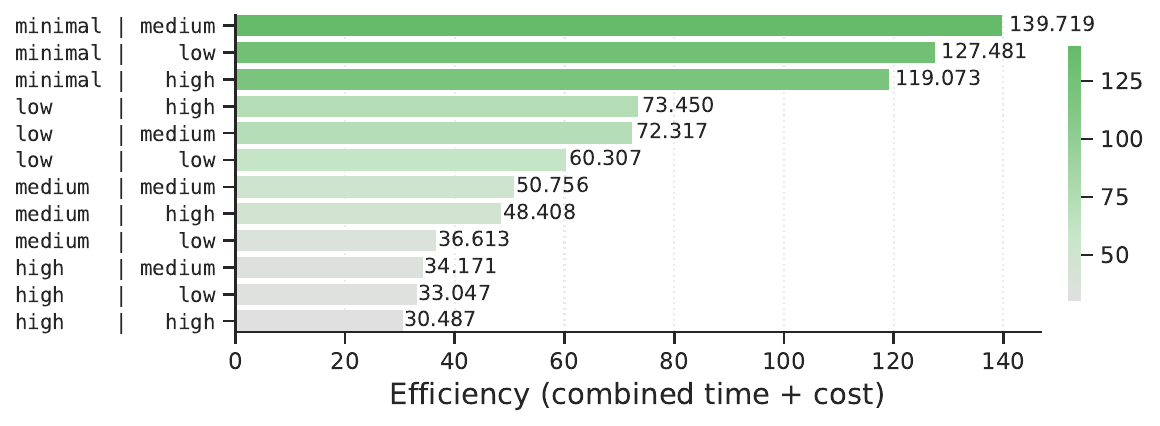}
        \caption{Combined performance–time–cost efficiency under GPT-5 reasoning configurations}
        \label{fig:gpt5_reasoning_efficiency_time_cost}
    \end{subfigure}

    \caption{
    Efficiency comparison under GPT-5 reasoning configurations  
    (notation: \texttt{Reasoning\_Effort} | \texttt{Text\_Verbosity})
    }
\end{figure}

\subsubsection{Reasoning Cost and the Resulting Efficiency Trade-off}

\paragraph{Is extra reasoning cost-effective for GPT-5?}

As shown in Figure~\ref{fig:RQ3_combined}, reasoning configurations induce large changes in cost.  
Thus, evaluating reasoning requires considering not only accuracy but also efficiency.  
Figures~\ref{fig:gpt5_reasoning_efficiency_time}–\ref{fig:gpt5_reasoning_efficiency_time_cost} visualize how performance, time, and cost interact across settings.

Overall, increasing \texttt{Reasoning\_Effort} improves performance but substantially increases both latency and token usage, reducing total efficiency.

The \textbf{medium–minimal} combination (\texttt{Reasoning\_Effort}=medium, \texttt{Text\_Verbosity}=minimal) achieves the highest values across all Efficiency Scores (performance–time, performance–cost, and performance–time–cost).  
Minimal verbosity minimizes token consumption while preserving accuracy, giving the best return per unit time and cost.

Conversely, the \textbf{high–high} configuration performs worst across all efficiency metrics.  
while achieving the highest accuracy (100.0) among all effort–verbosity pairs—about 2–3 points higher than typical medium-effort configurations.
Although accuracy increases slightly at maximum effort, token usage and cost rise by roughly \textbf{4–5 times}, and latency increases significantly.  
As a result, efficiency scores drop to less than half of the baseline.

\paragraph{Summary: Performance Gains vs. Cost Explosion}

Increasing reasoning effort in GPT-5 yields only modest performance gains but incurs substantial increases in both cost and latency.  
In many cases, \textbf{performance improvements of only 2–5 points require 4–5 times more tokens}.  
Thus:

\begin{itemize}
\item \textbf{High reasoning effort, as expected, increases performance but is not cost-efficient.}
\item \textbf{Minimal or medium reasoning is more efficient than high reasoning across all metrics.}
\item \textbf{Baseline GPT-5 or minimal reasoning often provides the best practical trade-off.}
\item \textbf{High reasoning effort is inefficient for large-scale or budget-limited evaluation settings.}
\end{itemize}

Overall, \textbf{additional reasoning effort benefits performance far less than it increases cost}, but if one aims only to raise the score without regard to cost, it can still be considered.
For large-scale or time-sensitive math evaluation tasks, baseline or minimal reasoning remains the most rational choice.
\section{Key Findings}
\label{sec:6_discussion}

This study conducted a comprehensive evaluation of 24 large language models (LLMs) on the 2026 Korean College Scholastic Ability Test (CSAT) Mathematics exam under a strictly zero-leakage setting. Our analysis extends beyond raw accuracy and examines how models differ across input conditions, reasoning configurations, cost profiles, and mathematical content types. The results demonstrate that LLM mathematical reasoning ability must be evaluated from a \textit{multidimensional} perspective—encompassing reasoning strategy, efficiency, modality sensitivity, and domain-specific weaknesses—rather than solely by the number of correct answers.

\paragraph{RQ1. Overall Mathematical Problem-Solving Ability of Current LLMs}

GPT-5 Codex achieved perfect normalized score (100.0) under multiple input–language configurations, including Text+Figure/English, Text-only/Korean, Image-only/Korean, and Text+Figure/Korean, demonstrating robust performance across both textual and visual modalities. In addition, the GPT-5 base model reached a perfect score under the Text-only/English setting. No other model–condition combination achieved a full score, indicating that perfect CSAT-level mathematical performance is currently limited to the GPT-5 family.

% GPT-5 Codex achieved a perfect score (100.0) under the Text-only and Korean-prompt condition, outperforming all other evaluated models. Grok 4, GPT-OSS-20B, GPT-5, and DeepSeek R1 also scored above 95, indicating strong mathematical reasoning capabilities beyond rote pattern recall.  

A notable finding is that \textbf{gpt-oss-20B}, despite its relatively small parameter count, achieved 95.7 points—surpassing some models more than seven times its size. This supports the observation that model scale is not an absolute determinant of mathematical reasoning ability. When cost and latency are considered jointly, gpt-oss-20B exhibited exceptional efficiency, with Claude 4.5 Haiku, Grok 4 Fast, and DeepSeek V3.2 Exp also offering strong performance-per-cost trade-offs.  

Most high-performing models completed all 46 questions within the human exam time equivalent (100 minutes), although a small number of models generated excessively long reasoning traces that led to substantial latency inflation.

\paragraph{RQ2. Effects of Input Conditions on Performance}

The effect of prompt language varied by model scale. Large, modern models such as Claude 4.5 Sonnet and GPT-5 Codex performed better with Korean prompts, whereas smaller models (e.g., GPT-5-mini, Grok 4 Fast) improved by up to 10 points when given English prompts. This suggests that \textbf{smaller models are more sensitive to language imbalance in their training data and exhibit weaker multilingual generalization}.  

Clear trends also emerged for input modality. \textbf{Text-only inputs achieved the highest overall accuracy and outperformed Image-only inputs for nearly all models.} Text+Figure inputs yielded accuracy within 0.3\% of Text-only, effectively matching it. Even in geometry questions, text inputs were sufficient, reflecting the exam’s design: for the 2026 CSAT, relevant geometric information could be inferred fully from textual descriptions.  
However, this outcome is specific to this exam format. In settings where visual reasoning is indispensable, modality effects may differ substantially.

\paragraph{RQ3. Effects of Mathematical Problem Characteristics}

When averaging across all models and condition combinations, Calculus yielded the lowest performance, with an average normalized score of 75.3 (19.6 pts).
Geometry followed with 78.4, while Common Mathematics and Probability \& Statistics remained in the low-to-mid 80 range.
These results indicate that \textbf{tasks requiring multi-step logical reasoning (Calculus) and spatial reasoning (Geometry) continue to present the greatest challenges for current LLMs.}

% Across all models, geometry showed the lowest accuracy, with an average normalized score of 77.7\%, substantially lower than common mathematics and calculus (90–100\%). For mid- and large-scale models, geometry accuracy decreased by up to 15 percentage points relative to other domains; for smaller models, the gap widened to as much as 46 percentage points. These results show that \textbf{geometric reasoning—requiring spatial understanding, figure interpretation, and multi-step relational reasoning—remains a difficult area for current LLMs}.  

The same pattern was observed across problem attribute difficulty. While 2–3 point problems achieved normalized scores of 0.91–0.97, 4-point problems dropped sharply to 0.68–0.71. This reveals that the \textbf{CSAT’s human-designed difficulty structure systematically induces performance stratification in LLMs as well}.  
Problem type showed similar trends: multiple-choice problems (0.93–0.97) outperformed constructed-response problems (0.68–0.71), though this is confounded by difficulty since constructed-response problems disproportionately correspond to high-difficulty questions.

\paragraph{RQ4. Benefits and Limitations of Enhanced Reasoning}

Within the GPT-5 family, increasing \texttt{Reasoning\_Effort} from minimal to high raised performance from 82.6 to 100.0 (+17.4), but the token usage rose from approximately 60 tokens/problem to 240–268 tokens/problem, a \textbf{four- to five-fold increase}. Efficiency Scores were highest under the minimal-verbosity configurations (1.58–1.70) and dropped sharply under high reasoning effort, reaching as low as 0.37.  

The effect of \texttt{Text\_Verbosity} was negligible, and \textbf{performance gains were driven almost entirely by \texttt{Reasoning\_Effort}}.  
Notably, minimal or low reasoning sometimes \emph{reduced} performance relative to the baseline, indicating that \textbf{forcing additional reasoning onto a model that already performs strong internal inference can be counterproductive}.  

These findings confirm that although additional reasoning effort can improve accuracy, it substantially increases computational cost and latency. Achieving an optimal balance of reasoning depth therefore requires \textbf{adaptive configuration based on model capability and problem difficulty}. Excessive reasoning is generally inefficient, especially in large-scale or resource-limited evaluation environments.
\section{Limitations}
\label{sec:7_limitation}

This study has several limitations stemming from the experimental design and evaluation environment.

\paragraph{Limited Scope of Model Deployment Settings}

All model evaluations were conducted through the OpenRouter API to ensure a uniform inference environment, reproducibility, and fair comparison across heterogeneous LLMs. While this approach offers consistency, it also introduces several constraints.  

First, LLMs served through OpenRouter may exhibit performance differences compared to their native deployments. Variations in quantization levels, optimization settings, or backend configurations can lead to slight deviations from the models’ originally intended performance profiles. Although such discrepancies are not fully avoidable, we mitigated their impact by applying OpenRouter’s default configurations uniformly across all models to preserve the fairness of relative comparisons. Future work will complement this setup by running open-weight models locally under tightly controlled closed-examination conditions.

Second, due to OpenRouter access constraints and the selection criteria based on leading models listed in AIME 2025, Korean-specialized LLMs were necessarily excluded from this benchmark. Korean-optimized models may exhibit different sensitivity to prompt language, and their absence limits the generalizability of the multilingual findings. Incorporating such models remains an important direction for further investigation.

\paragraph{Restricted Scope of Reasoning-Enhanced Experiments}

Experiments involving \texttt{Reasoning\_Effort} and verbosity adjustments were conducted exclusively on the GPT-5 family. At present, “reasoning,” “thinking,” and related modes differ significantly in definition and implementation across model providers, making consistent cross-model comparison difficult. For methodological coherence, we restricted reasoning analysis to the GPT-5 series, which offers formally documented and controllable reasoning parameters. Consequently, reasoning-related conclusions in this study are specific to GPT-5 and should be generalized to other architectures only with caution. Broader cross-model reasoning benchmarks are an important direction for future work.

\paragraph{Absence of Human Benchmark Comparison}

Although this study evaluates the 2026 CSAT Mathematics exam, official statistics on human test-taker performance were unavailable at the time of analysis and at the time of submission\footnote{2025.11.27}. As a result, we were unable to directly compare LLM performance patterns with those of human examinees. Once the official score distributions and per-problem correctness statistics are released, future work will analyze the gap between “problems difficult for humans” and “problems difficult for LLMs,” enabling deeper insight into the cognitive characteristics and limitations of LLMs relative to human reasoning.
\section{Conclusion}
\label{sec:8_conclusion}

This study introduces a benchmark and evaluation framework designed to assess the mathematical reasoning capabilities of state-of-the-art LLMs under a fully contamination-free setting using the 2026 Korean CSAT Mathematics exam. All 46 problems were digitized within two hours of the exam’s public release, and 24 models were evaluated across systematically varied input modalities, prompt languages, and reasoning intensities. \\

Experimental results show that while the strongest commercial models achieve scores above 90 on CSAT-level mathematics, their performance varies substantially depending on problem characteristics (e.g., Calculus, high-point problems), input conditions (language and modality), and the presence of time constraints. Although reasoning-enhanced modes improve absolute accuracy, they incur significant computational and monetary costs, making baseline configurations more practical and cost-efficient for many use cases. \\

This work provides three main contributions. First, it establishes a fully contamination-free evaluation environment that fundamentally eliminates the data leakage concerns present in prior benchmarks. Second, it formalizes a standardized digitization pipeline that converts human-targeted exam materials into LLM-ready evaluation data. Third, it proposes a practical assessment perspective that extends beyond accuracy to incorporate time, cost, and overall efficiency. \\

The study nevertheless has several limitations, including restrictions arising from the use of OpenRouter APIs, the absence of Korean-specialized models, the reasoning experiments being limited to the GPT-5 family, and the lack of direct comparison with human test-taker statistics. Future work will include longitudinal analyses using CSAT data, alignment with official scoring distributions, extensions to additional subject areas, and evaluation of Korean-optimized models. \\

Ultimately, this study highlights the need to shift LLM assessment from “how many problems the model gets correct” to “how efficiently and reliably the model problems under real-world constraints.” We expect that the proposed evaluation framework will serve as a new reference point for assessing and deploying LLMs in high-stakes exam settings such as the CSAT.

% \section*{Acknowledgments}
% This was was supported in part by......

%Bibliography
% \bibliographystyle{unsrt}  
\bibliographystyle{plainnat}
\bibliography{references}  

\newpage
\appendix

\section{Full Model Results: Performance, Latency, and Cost}
\label{sec:appendix_whole_performance}

This appendix supplements the summary provided in the main text by presenting the complete results for all models across every combination of input modality and prompt language.  
Table~\ref{tab:score} reports the normalized performance scores for each model–condition pair.  
Table~\ref{tab:latency} summarizes the average response latency under the same conditions, and  
Table~\ref{tab:cost} provides the total cost incurred to evaluate each model.

\begin{table}[!ht]
\centering
\caption{Normalized performance scores across all models and all input–language combinations.}
\begin{adjustbox}{width=\linewidth}
\begin{tabular}{c l ccc ccc}
\toprule
\multirow{2}{*}{Model Type} & \multirow{2}{*}{Model}
  & \multicolumn{3}{c}{English}
  & \multicolumn{3}{c}{Korean} \\
\cmidrule{3-8}
 &  & Text & Image & Text+Figure & Text & Image & Text+Figure \\
\midrule
\multirow{11}{*}{Closed-weight} 
& \includegraphics[height=0.8em]{icon/openai.pdf} GPT-5 Codex
& 97.8 & 97.8 & 100.0 & 100.0 & 100.0 & 100.0 \\
& \includegraphics[height=0.8em]{icon/openai.pdf} GPT-5
& 95.7 & 100.0 & 95.7 & 95.7 & 97.8 & 97.8 \\
& \includegraphics[height=0.8em]{icon/openai.pdf} GPT-5-mini
& 95.7 & 91.3 & 93.5 & 93.5 & 91.3 & 93.5 \\
& \includegraphics[height=0.8em]{icon/openai.pdf} GPT-5-nano
& 93.5 & 78.3 & 95.7 & 89.1 & 67.4 & 89.1 \\
\cmidrule{2-8}
& \includegraphics[height=0.8em]{icon/xai.pdf} Grok 4
& 93.5 & 56.5 & 95.7 & 97.8 & 52.2 & 95.7 \\
& \includegraphics[height=0.8em]{icon/xai.pdf} Grok 4 Fast
& 93.5 & 32.6 & 91.3 & 95.7 & 37.0 & 97.8 \\
\cmidrule{2-8}
& \includegraphics[height=0.8em]{icon/claude.png} Claude 4.5 Sonnet
& 73.9 & 67.4 & 78.3 & 80.4 & 71.7 & 82.6 \\
& \includegraphics[height=0.8em]{icon/claude.png} Claude 4.5 Haiku
& 82.6 & 65.2 & 78.3 & 80.4 & 67.4 & 82.6 \\
& \includegraphics[height=0.8em]{icon/claude.png} Claude 4.1 Opus
& 76.1 & 56.5 & 73.9 & 80.4 & 69.6 & 76.1 \\
\cmidrule{2-8}
& \includegraphics[height=0.8em]{icon/gemini.png} Gemini 2.5 Pro
& 95.7 & 97.8 & 93.5 & 91.3 & 93.5 & 93.5 \\
& \includegraphics[height=0.8em]{icon/gemini.png} Gemini 2.5 Flash
& 80.4 & 84.8 & 80.4 & 71.7 & 73.9 & 71.7 \\
\midrule
\multirow{13}{*}{Open-weight}
& \includegraphics[height=0.8em]{icon/openai.pdf} gpt-oss-20B
& 93.5 & -- & -- & 95.7 & -- & -- \\
\cmidrule{2-8}
& \includegraphics[height=0.5em]{icon/qwen.pdf} Qwen3 235B A22B
& 89.1 & 73.9 & 91.3 & 82.6 & 69.6 & 84.8 \\
& \includegraphics[height=0.5em]{icon/qwen.pdf} Qwen3 235B A22B Thinking 2507
& 93.5 & 84.8 & 91.3 & 89.1 & 97.8 & 84.8 \\
\cmidrule{2-8}
& \includegraphics[height=0.8em]{icon/deepseek.pdf} Deepseek R1 0528
& 93.5 & -- & -- & 95.7 & -- & -- \\
& \includegraphics[height=0.8em]{icon/deepseek.pdf} Deepseek V3.2 Exp
& 91.3 & -- & -- & 87.0 & -- & -- \\
\cmidrule{2-8}
& \includegraphics[height=0.8em]{icon/zai.pdf} GLM 4.6
& 87.0 & -- & -- & 89.1 & -- & -- \\
\cmidrule{2-8}
& \includegraphics[height=0.8em]{icon/mistral.png} Magistral Medium 2506
& 71.7 & -- & -- & 82.6 & -- & -- \\
& \includegraphics[height=0.8em]{icon/mistral.png} Magistral Medium 2506 Thinking
& 84.8 & -- & -- & 80.4 & -- & -- \\
\cmidrule{2-8}
& \includegraphics[height=0.8em]{icon/minimax.pdf} MiniMax M2
& 97.8 & -- & -- & 82.6 & -- & -- \\
\cmidrule{2-8}
& \includegraphics[height=0.8em]{icon/meta.pdf} Llama 4 Maverick
& 23.9 & 65.2 & 32.6 & 21.7 & 43.5 & 28.3 \\
& \includegraphics[height=0.8em]{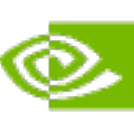} Llama 3.3 Nemotron Super 49B V1.5
& 80.4 & -- & -- & 80.4 & -- & -- \\
\cmidrule{2-8}
& \includegraphics[height=0.8em]{icon/moonshot.pdf} Kimi K2 0905
& 78.3 & -- & -- & 78.3 & -- & -- \\
& \includegraphics[height=0.8em]{icon/moonshot.pdf} Kimi K2 Thinking
& 91.3 & -- & -- & 82.6 & -- & -- \\
\bottomrule
\end{tabular}
\end{adjustbox}
\label{tab:score}
\end{table}

\begin{table}[!t]
\centering
\caption{Average latency (ms) across all models and input–language combinations.}
\begin{adjustbox}{width=\linewidth}
\begin{tabular}{c l rrr rrr}
\toprule
\multirow{2}{*}{Model Type} & \multirow{2}{*}{Model}
  & \multicolumn{3}{c}{English}
  & \multicolumn{3}{c}{Korean} \\
\cmidrule{3-8}
 &  & Text & Image & Text+Figure & Text & Image & Text+Figure \\
\midrule
\multirow{11}{*}{Closed-weight} 
& \includegraphics[height=0.8em]{icon/openai.pdf} GPT-5 Codex 
& 5457.4 & 5808.1 & 6130.2 & 4333.3 & 7426.4 & 5393.7 \\
 & \includegraphics[height=0.8em]{icon/openai.pdf} GPT-5 
 & 2134.3 & 4839.8 & 3561.1 & 2521.6 & 4832.4 & 3882.0 \\
 & \includegraphics[height=0.8em]{icon/openai.pdf} GPT-5-mini 
 & 2297.2 & 2221.5 & 2012.6 & 2623.8 & 2875.0 & 2402.5 \\
 & \includegraphics[height=0.8em]{icon/openai.pdf} GPT-5-nano 
 & 3081.7 & 2816.4 & 3354.8 & 4527.9 & 3454.1 & 2860.4 \\
\cmidrule{2-8}
 & \includegraphics[height=0.8em]{icon/xai.pdf} Grok 4 
 & 6234.1 & 16463.0 & 4371.7 & 6734.4 & 16875.1 & 6302.7 \\
 & \includegraphics[height=0.8em]{icon/xai.pdf} Grok 4 Fast 
 & 840.3 & 2866.6 & 861.6 & 1106.7 & 2812.5 & 868.4 \\
\cmidrule{2-8}
 & \includegraphics[height=0.8em]{icon/claude.png} Claude 4.5 Sonnet 
 & 641.3 & 719.3 & 655.3 & 670.7 & 751.7 & 676.8 \\
 & \includegraphics[height=0.8em]{icon/claude.png} Claude 4.5 Haiku 
 & 275.9 & 308.9 & 284.4 & 281.6 & 343.0 & 294.1 \\
 & \includegraphics[height=0.8em]{icon/claude.png} Claude 4.1 Opus 
 & 780.1 & 927.9 & 829.2 & 867.1 & 958.6 & 858.0 \\
\cmidrule{2-8}
 & \includegraphics[height=0.8em]{icon/gemini.png} Gemini 2.5 Pro 
 & 2742.4 & 2923.7 & 2581.8 & 2427.3 & 2770.0 & 2421.2 \\
 & \includegraphics[height=0.8em]{icon/gemini.png} Gemini 2.5 Flash 
 & 1537.0 & 1461.4 & 1408.9 & 1661.7 & 1736.1 & 1717.4 \\
\midrule
\multirow{13}{*}{Open-weight} 
 & \includegraphics[height=0.8em]{icon/openai.pdf} gpt-oss-20B 
 & 1729.8 & - & - & 1304.6 & - & - \\
\cmidrule{2-8}
 & \includegraphics[height=0.5em]{icon/qwen.pdf} Qwen3 235B A22B 
 & 2890.0 & 3888.1 & 2309.5 & 2735.6 & 6997.7 & 5865.2 \\
 & \includegraphics[height=0.5em]{icon/qwen.pdf} Qwen3 235B A22B Thinking 2507 
 & 12616.1 & 9157.2 & 9288.6 & 9665.3 & 10350.5 & 6639.4 \\
\cmidrule{2-8}
 & \includegraphics[height=0.8em]{icon/deepseek.pdf} Deepseek R1 0528 
 & 11540.0 & - & - & 10885.7 & - & - \\
 & \includegraphics[height=0.8em]{icon/deepseek.pdf} Deepseek V3.2 Exp 
 & 3161.0 & - & - & 3057.1 & - & - \\
\cmidrule{2-8}
 & \includegraphics[height=0.8em]{icon/zai.pdf} GLM 4.6 
 & 13152.2 & - & - & 9884.7 & - & - \\
\cmidrule{2-8}
 & \includegraphics[height=0.8em]{icon/mistral.png} Magistral Medium 2506 
 & 3986.7 & - & - & 2771.8 & - & - \\
 & \includegraphics[height=0.8em]{icon/mistral.png} Magistral Medium 2506 (thinking) 
 & 4998.6 & - & - & 3862.6 & - & - \\
\cmidrule{2-8}
 & \includegraphics[height=0.8em]{icon/minimax.pdf} MiniMax M2 
 & 4037.6 & - & - & 3786.8 & - & - \\
\cmidrule{2-8}
 & \includegraphics[height=0.8em]{icon/meta.pdf} Llama 4 Maverick 
 & 589.7 & 1151.0 & 723.6 & 531.9 & 915.0 & 1186.2 \\
 & \includegraphics[height=0.8em]{icon/nvidia.pdf} Llama 3.3 Nemotron Super 49B V1.5 
 & 3235.2 & - & - & 3262.7 & - & - \\
\cmidrule{2-8}
 & \includegraphics[height=0.8em]{icon/moonshot.pdf} Kimi K2 0905 
 & 10250.8 & - & - & 7242.8 & - & - \\
 & \includegraphics[height=0.8em]{icon/moonshot.pdf} Kimi K2 Thinking 
 & 26214.9 & - & - & 33906.9 & - & - \\
\bottomrule
\end{tabular}
\end{adjustbox}
\label{tab:latency}
\end{table}

\begin{table}[!t]
\centering
\caption{Total evaluation cost (USD) across all models and input–language combinations.}
\begin{adjustbox}{width=\linewidth}
\begin{tabular}{c l lll lll}
\toprule
\multirow{2}{*}{Model Type} & \multirow{2}{*}{Model}
  & \multicolumn{3}{c}{English}
  & \multicolumn{3}{c}{Korean} \\
\cmidrule{3-8}
 &  & Text & Image & Text+Figure & Text & Image & Text+Figure \\
\midrule
\multirow{11}{*}{Closed-weight} 
& \includegraphics[height=0.8em]{icon/openai.pdf} GPT-5 Codex
& 2.5181 & 1.9222 & 2.0817 & 1.8009 & 2.3985 & 2.0699 \\
& \includegraphics[height=0.8em]{icon/openai.pdf} GPT-5 
& 1.4378 & 1.6743 & 1.7073 & 1.7384 & 1.7896 & 1.7618 \\
& \includegraphics[height=0.8em]{icon/openai.pdf} GPT-5-mini 
& 0.3187 & 0.3022 & 0.2683 & 0.3741 & 0.3699 & 0.3258 \\
& \includegraphics[height=0.8em]{icon/openai.pdf} GPT-5-nano 
& 0.1063 & 0.1126 & 0.1250 & 0.1555 & 0.1287 & 0.1064 \\
\cmidrule{2-8}
& \includegraphics[height=0.8em]{icon/xai.pdf} Grok 4
& 10.5872 & 24.3791 & 7.5040 & 11.4779 & 26.1858 & 9.8555 \\
& \includegraphics[height=0.8em]{icon/xai.pdf} Grok 4 Fast
& 0.2028 & 0.6162 & 0.2071 & 0.2662 & 0.4924 & 0.2024 \\
\cmidrule{2-8}
& \includegraphics[height=0.8em]{icon/claude.png} Claude 4.5 Sonnet
& 1.0253 & 1.2765 & 1.1099 & 1.1663 & 1.4694 & 1.2543 \\
& \includegraphics[height=0.8em]{icon/claude.png} Claude 4.5 Haiku 
& 0.2084 & 0.2415 & 0.2214 & 0.2333 & 0.2842 & 0.2456 \\
& \includegraphics[height=0.8em]{icon/claude.png} Claude 4.1 Opus  
& 3.3942 & 3.8985 & 3.6921 & 4.0069 & 4.3796 & 4.0680 \\
\cmidrule{2-8}
& \includegraphics[height=0.8em]{icon/gemini.png} Gemini 2.5 Pro   
& 5.3461 & 5.8185 & 5.2356 & 4.8186 & 5.4515 & 4.9693 \\
& \includegraphics[height=0.8em]{icon/gemini.png} Gemini 2.5 Flash 
& 1.0569 & 1.0381 & 0.9517 & 1.1582 & 1.2454 & 1.2030 \\
\midrule
\multirow{13}{*}{Open-weight} 
& \includegraphics[height=0.8em]{icon/openai.pdf} gpt-oss-20B
& 0.0075 & -      & -      & 0.0069 & -      & -      \\
\cmidrule{2-8}
& \includegraphics[height=0.5em]{icon/qwen.pdf} Qwen3 235B A22B  
& 0.1886 & 0.3279 & 0.1835 & 0.2292 & 0.5302 & 0.3951 \\
& \includegraphics[height=0.5em]{icon/qwen.pdf} Qwen3 235B A22B Thinking 2507 
& 0.8418 & 0.8858 & 0.7319 & 0.7146 & 0.6988 & 0.6756 \\
\cmidrule{2-8}
& \includegraphics[height=0.8em]{icon/deepseek.pdf} Deepseek R1 0528 
& 1.0082 & -      & -      & 0.8708 & -      & -      \\
& \includegraphics[height=0.8em]{icon/deepseek.pdf} Deepseek V3.2 Exp
& 0.0332 & -      & -      & 0.0373 & -      & -      \\
\cmidrule{2-8}
& \includegraphics[height=0.8em]{icon/zai.pdf} GLM 4.6 
& 1.4065 & -      & -      & 0.9748 & -      & -      \\
\cmidrule{2-8}
& \includegraphics[height=0.8em]{icon/mistral.png} Magistral Medium 2506 
& 1.6690 & -      & -      & 1.1278 & -      & -      \\
& \includegraphics[height=0.8em]{icon/mistral.png} Magistral Medium 2506 (thinking)
& 2.1733 & -      & -      & 1.4425 & -      & -      \\
\cmidrule{2-8}
& \includegraphics[height=0.8em]{icon/minimax.pdf} MiniMax M2 
& 0.4169 & -      & -      & 0.3706 & -      & -      \\
\cmidrule{2-8}
& \includegraphics[height=0.8em]{icon/meta.pdf} Llama 4 Maverick 
& 0.0440 & 0.0378 & 0.0353 & 0.0398 & 0.0402 & 0.0328 \\
& \includegraphics[height=0.8em]{icon/nvidia.pdf} Llama 3.3 Nemotron Super 49B V1.5  
& 0.1092 & -      & -      & 0.1070 & -      & -      \\
\cmidrule{2-8}
& \includegraphics[height=0.8em]{icon/moonshot.pdf} Kimi K2 0905 
& 0.2891 & -      & -      & 0.2124 & -      & -      \\
& \includegraphics[height=0.8em]{icon/moonshot.pdf} Kimi K2 Thinking 
& 1.4857 & -      & -      & 2.0854 & -      & -      \\
\bottomrule
\end{tabular}
\end{adjustbox}
\label{tab:cost}
\end{table}

\FloatBarrier

\section{Full Results on Model-wise Performance–Time–Cost Efficiency}
\label{sec:appendix_efficiency_score}

\begin{figure}[!t]
    \centering
    \begin{subfigure}[t]{\textwidth}
        \centering
        \includegraphics[width=\linewidth]{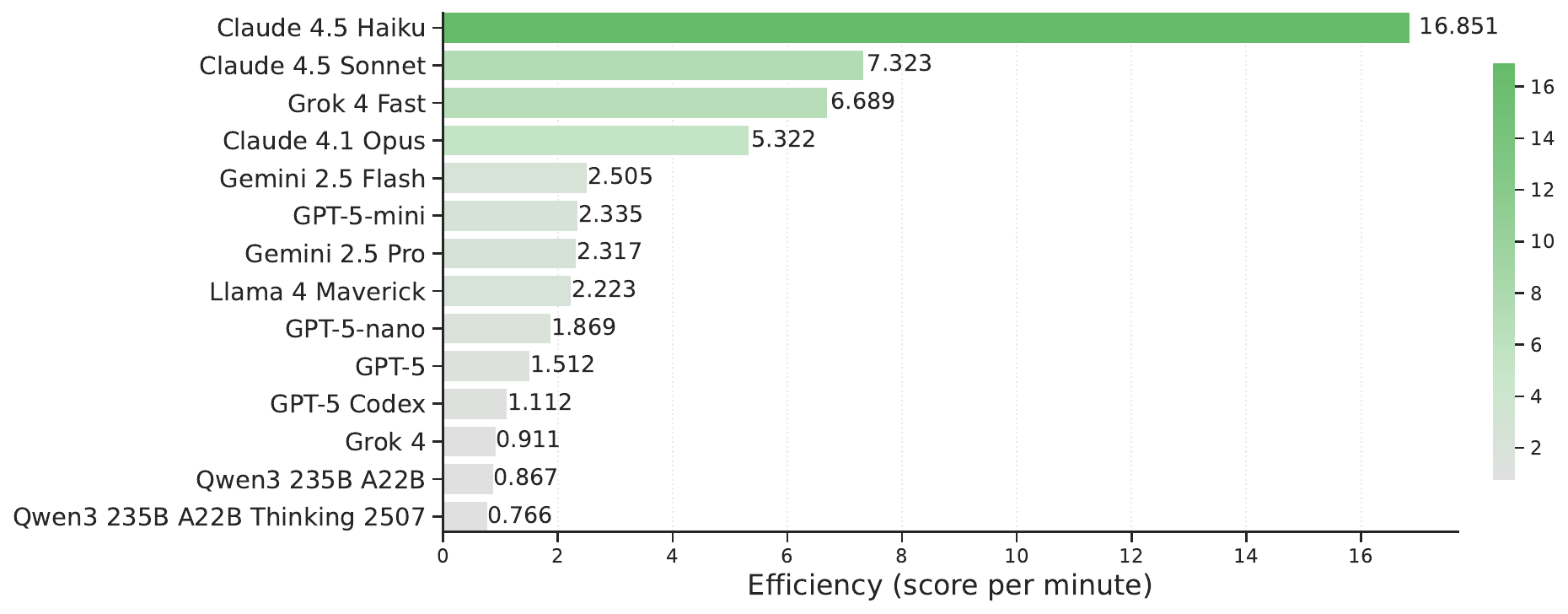}
        \caption{Comparison of performance–time Efficiency Scores \textbf{(input modality: Text+Figure, prompt language: Korean)}}
        \label{fig:efficiency_time_ko_ti}
    \end{subfigure}
    
    \begin{subfigure}[t]{\textwidth}
        \centering
        \includegraphics[width=\linewidth]{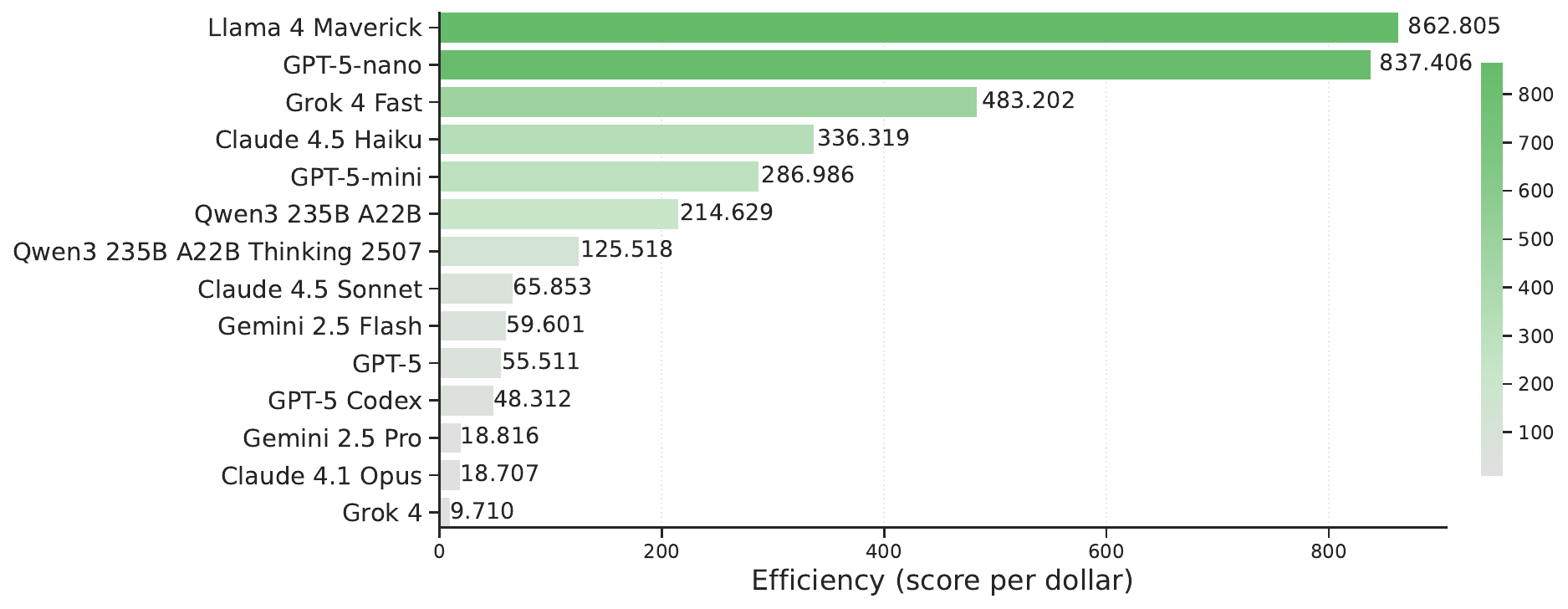}
        \caption{Comparison of performance–cost Efficiency Scores \textbf{(input modality: Text+Figure, prompt language: Korean)}}
        \label{fig:efficiency_cost_ko_ti}
    \end{subfigure}
    
    \begin{subfigure}[t]{\textwidth}
        \centering
        \includegraphics[width=\linewidth]{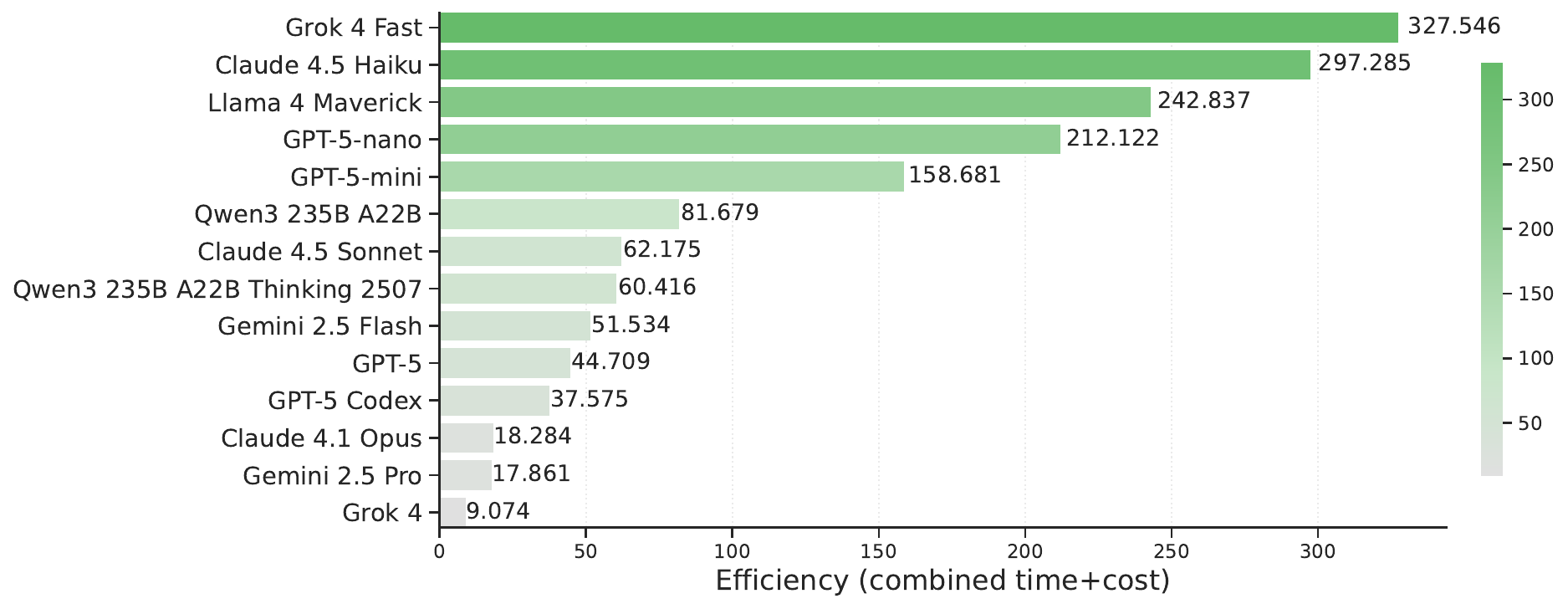}
        \caption{Comparison of performance–time–cost Efficiency Scores \textbf{(input modality: Text+Figure, prompt language: Korean)}}
        \label{fig:efficiency_time_cost_ko_ti}
    \end{subfigure}
    \caption{Comparison of performance–time–cost Efficiency Scores across models \textbf{(input modality: Text+Figure, prompt language: Korean)}}
    \label{fig:efficiency_ko_ti}
\end{figure}

\begin{figure}[!t]
    \centering
    \begin{subfigure}[t]{\textwidth}
        \centering
        \includegraphics[width=\linewidth]{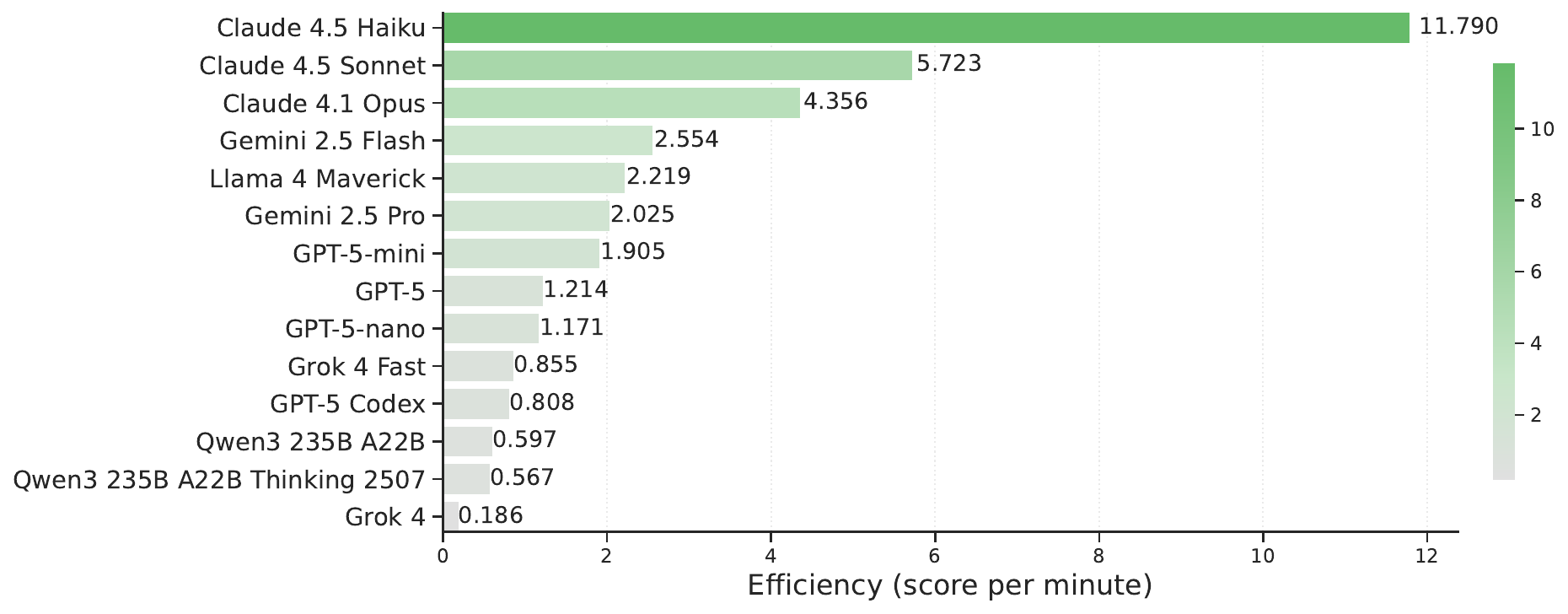}
        \caption{Comparison of performance–time Efficiency Scores \textbf{(input modality: Image-only, prompt language: Korean)}}
        \label{fig:efficiency_time_ko_i}
    \end{subfigure}
    
    \begin{subfigure}[t]{\textwidth}
        \centering
        \includegraphics[width=\linewidth]{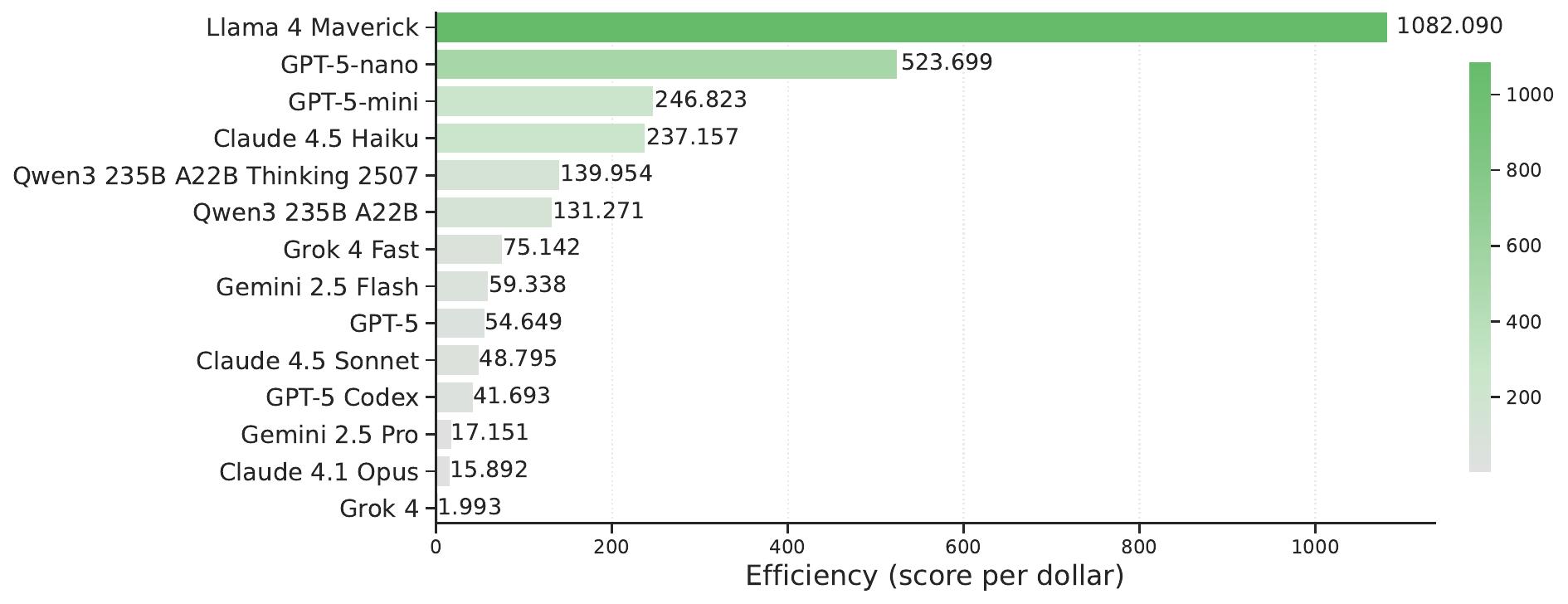}
        \caption{Comparison of performance–cost Efficiency Scores \textbf{(input modality: Image-only, prompt language: Korean)}}
        \label{fig:efficiency_cost_ko_i}
    \end{subfigure}
    
    \begin{subfigure}[t]{\textwidth}
        \centering
        \includegraphics[width=\linewidth]{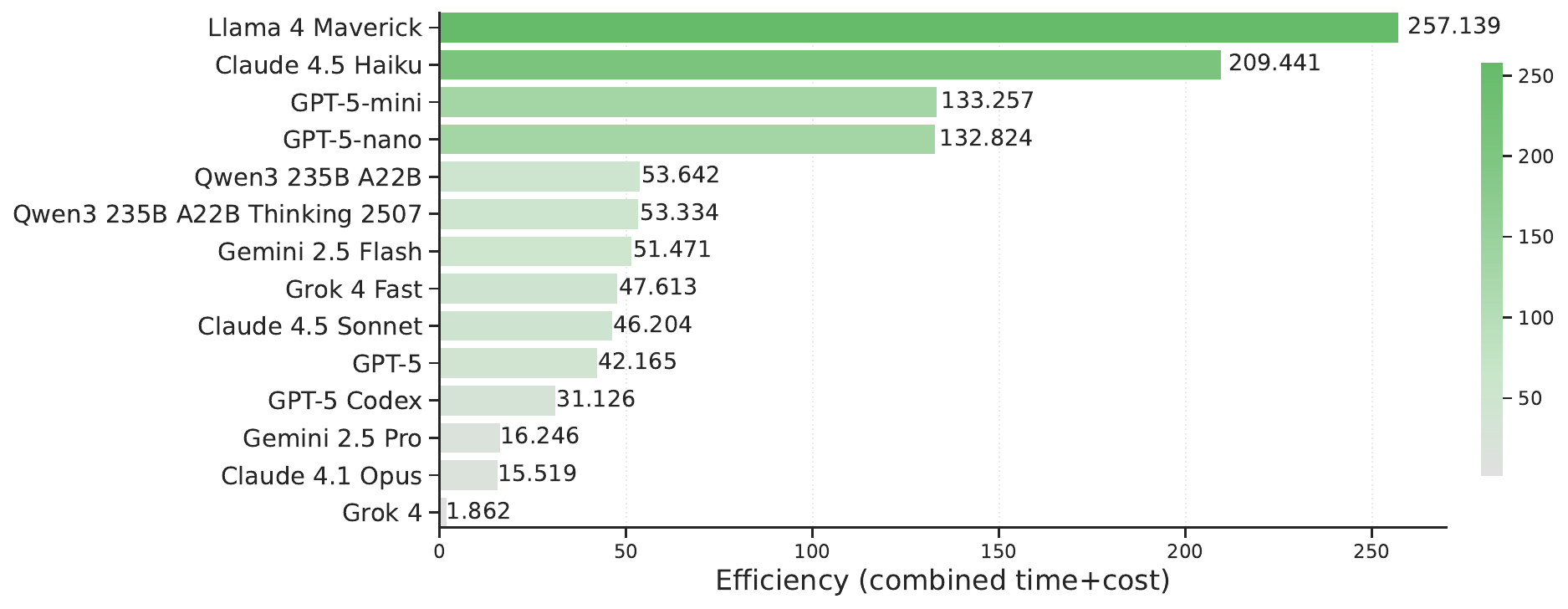}
        \caption{Comparison of performance–time–cost Efficiency Scores \textbf{(input modality: Image-only, prompt language: Korean)}}
        \label{fig:efficiency_time_cost_ko_i}
    \end{subfigure}
    \caption{Comparison of performance–time–cost Efficiency Scores across models \textbf{(input modality: Image-only, prompt language: Korean)}}
    \label{fig:efficiency_ko_i}
\end{figure}

\begin{figure}[!t]
    \centering
    \begin{subfigure}[t]{\textwidth}
        \centering
        \includegraphics[width=\linewidth]{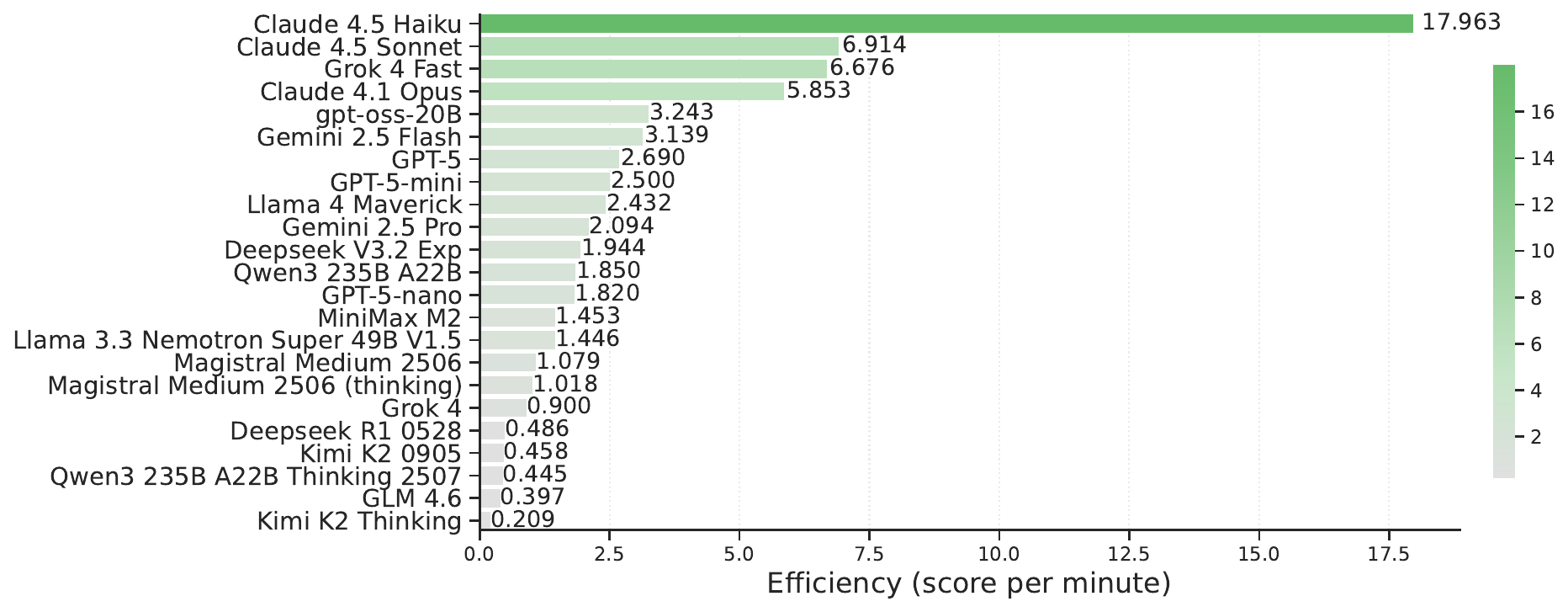}
        \caption{Comparison of performance–time Efficiency Scores \textbf{(input modality: Text-only, prompt language: English)}}
        \label{fig:efficiency_time_en_t}
    \end{subfigure}
    
    \begin{subfigure}[t]{\textwidth}
        \centering
        \includegraphics[width=\linewidth]{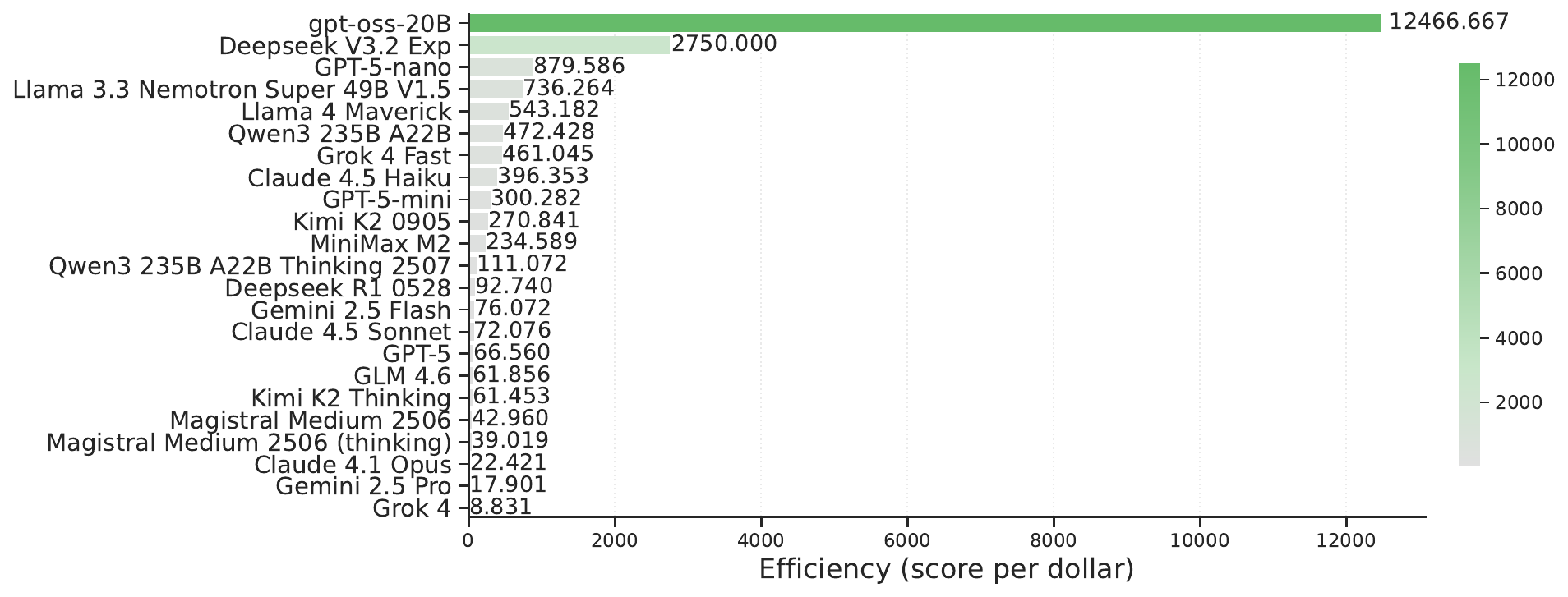}
        \caption{Comparison of performance–cost Efficiency Scores \textbf{(input modality: Text-only, prompt language: English)}}
        \label{fig:efficiency_cost_en_t}
    \end{subfigure}
    
    \begin{subfigure}[t]{\textwidth}
        \centering
        \includegraphics[width=\linewidth]{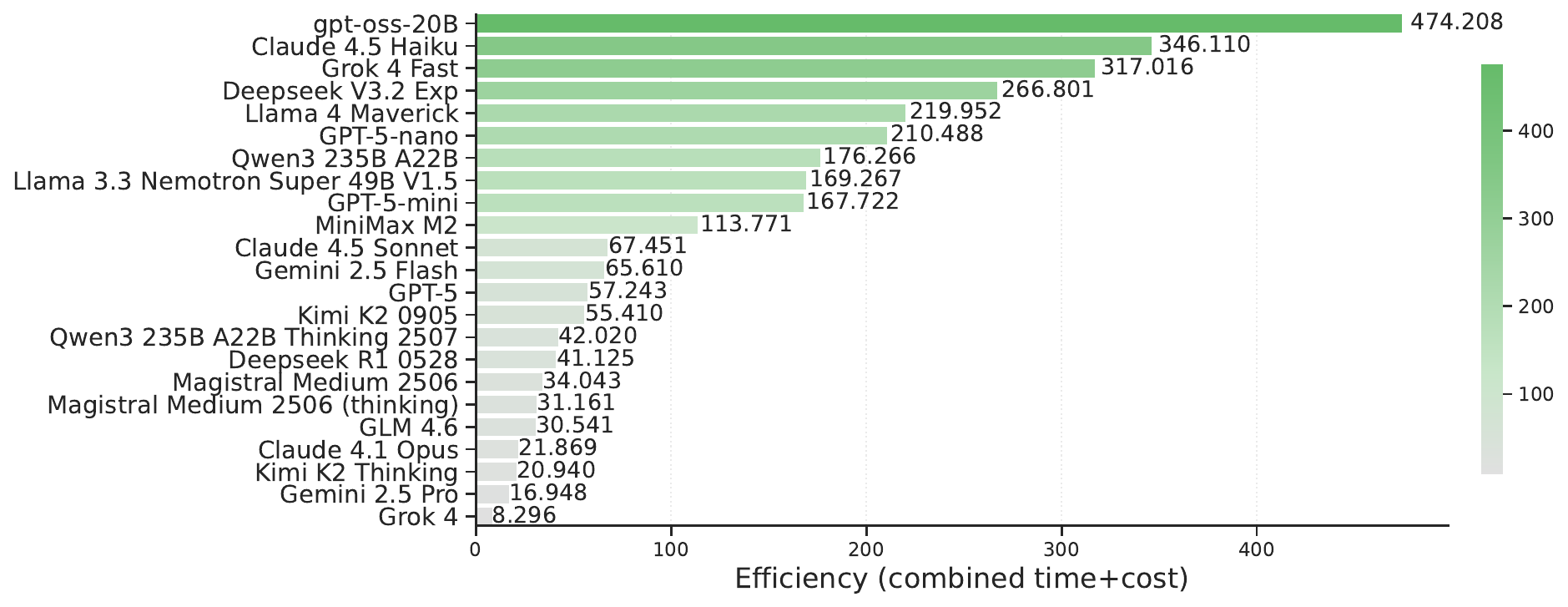}
        \caption{Comparison of performance–time–cost Efficiency Scores \textbf{(input modality: Text-only, prompt language: English)}}
        \label{fig:efficiency_time_cost_en_t}
    \end{subfigure}
    \caption{Comparison of performance–time–cost Efficiency Scores across models \textbf{(input modality: Text-only, prompt language: English)}}
    \label{fig:efficiency_en_t}
\end{figure}

\begin{figure}[!t]
    \centering
    \begin{subfigure}[t]{\textwidth}
        \centering
        \includegraphics[width=\linewidth]{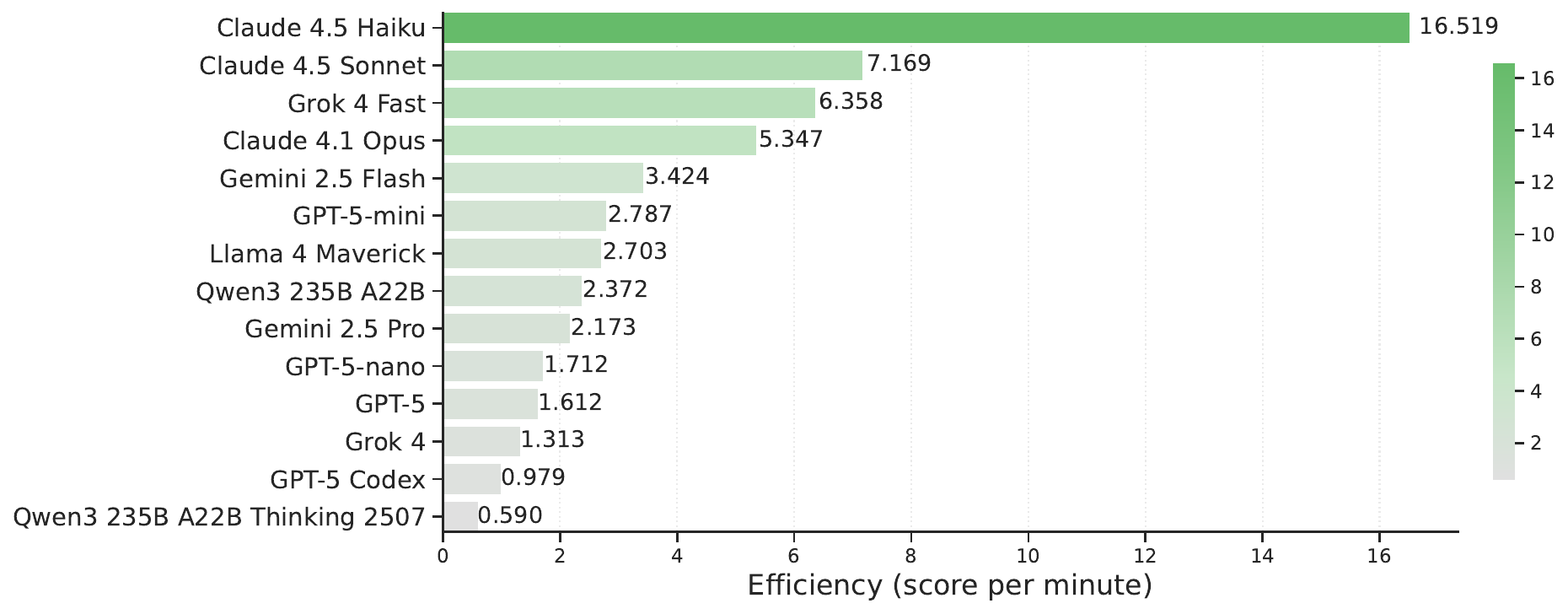}
        \caption{Comparison of performance–time Efficiency Scores \textbf{(input modality: Text+Figure, prompt language: English)}}
        \label{fig:efficiency_time_en_ti}
    \end{subfigure}
    
    \begin{subfigure}[t]{\textwidth}
        \centering
        \includegraphics[width=\linewidth]{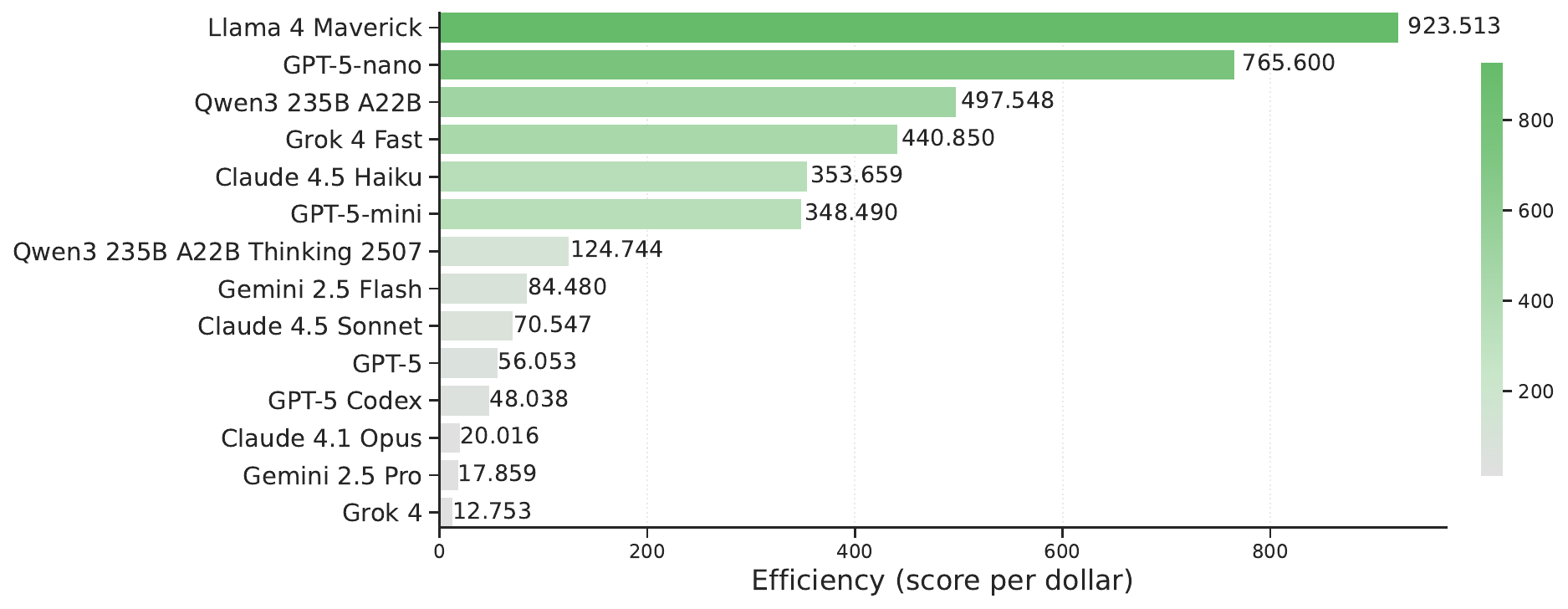}
        \caption{Comparison of performance–cost Efficiency Scores \textbf{(input modality: Text+Figure, prompt language: English)}}
        \label{fig:efficiency_cost_en_ti}
    \end{subfigure}
    
    \begin{subfigure}[t]{\textwidth}
        \centering
        \includegraphics[width=\linewidth]{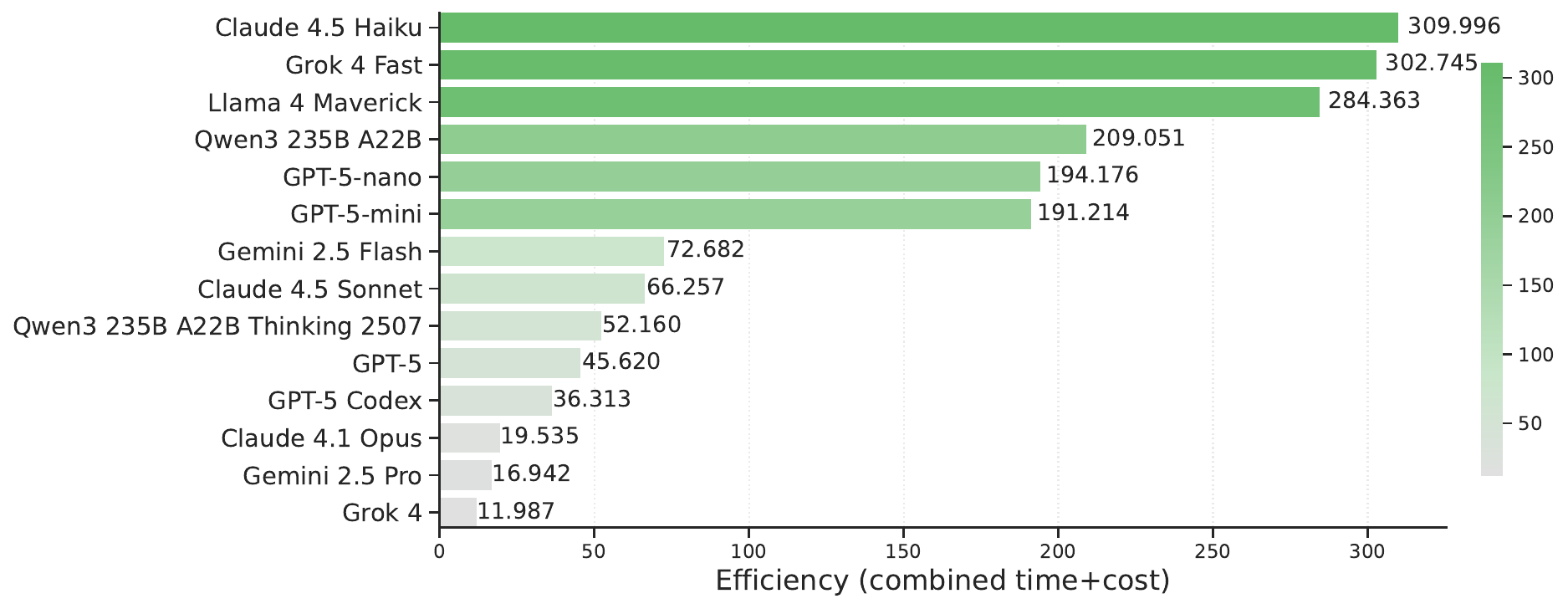}
        \caption{Comparison of performance–time–cost Efficiency Scores \textbf{(input modality: Text+Figure, prompt language: English)}}
        \label{fig:efficiency_time_cost_en_ti}
    \end{subfigure}
    \caption{Comparison of performance–time–cost Efficiency Scores across models \textbf{(input modality: Text+Figure, prompt language: English)}}
    \label{fig:efficiency_en_ti}
\end{figure}

\begin{figure}[!t]
    \centering
    \begin{subfigure}[t]{\textwidth}
        \centering
        \includegraphics[width=\linewidth]{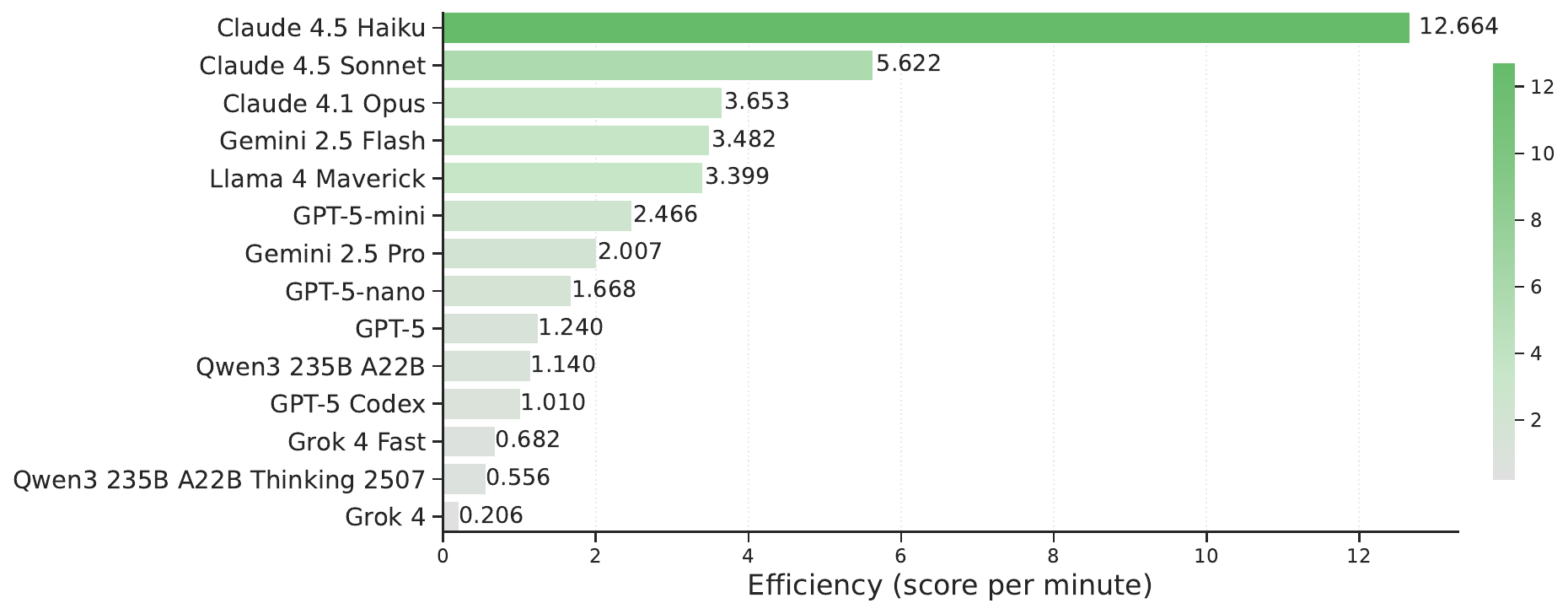}
        \caption{Comparison of performance–time Efficiency Scores \textbf{(input modality: Image-only, prompt language: English)}}
        \label{fig:efficiency_time_en_i}
    \end{subfigure}
    
    \begin{subfigure}[t]{\textwidth}
        \centering
        \includegraphics[width=\linewidth]{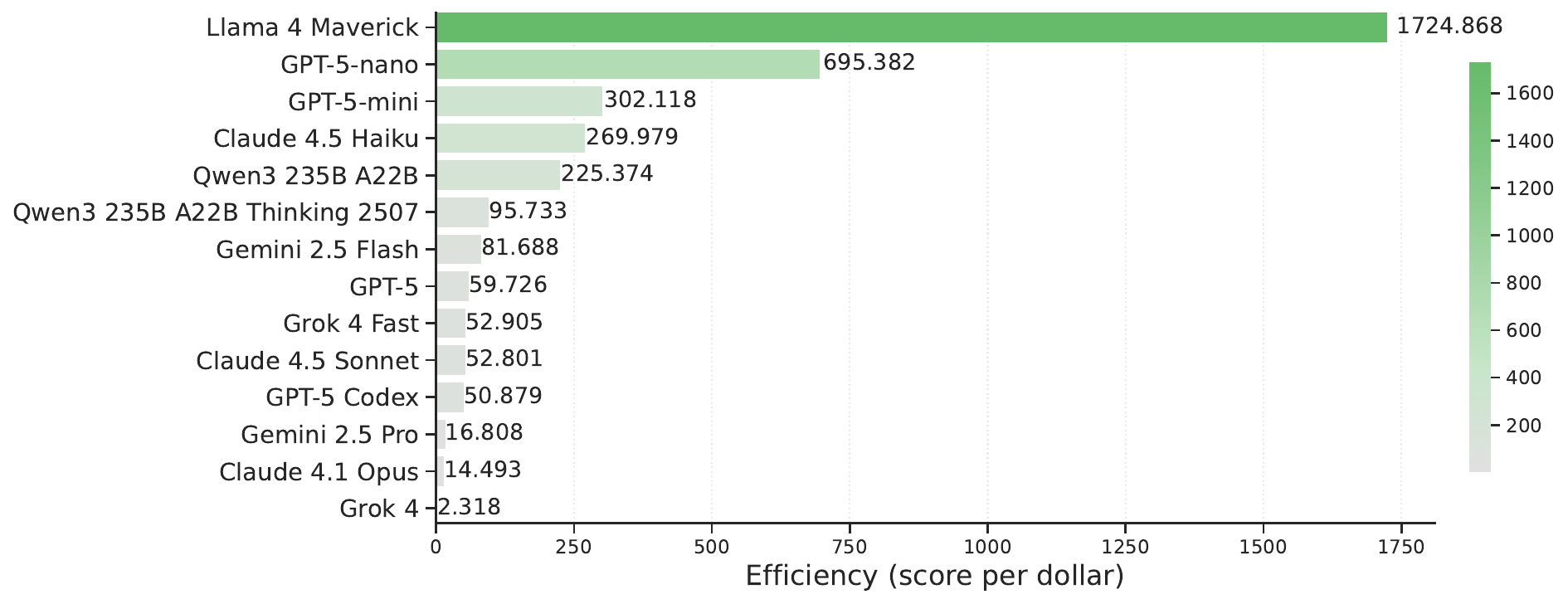}
        \caption{Comparison of performance–cost Efficiency Scores \textbf{(input modality: Image-only, prompt language: English)}}
        \label{fig:efficiency_cost_en_i}
    \end{subfigure}
    
    \begin{subfigure}[t]{\textwidth}
        \centering
        \includegraphics[width=\linewidth]{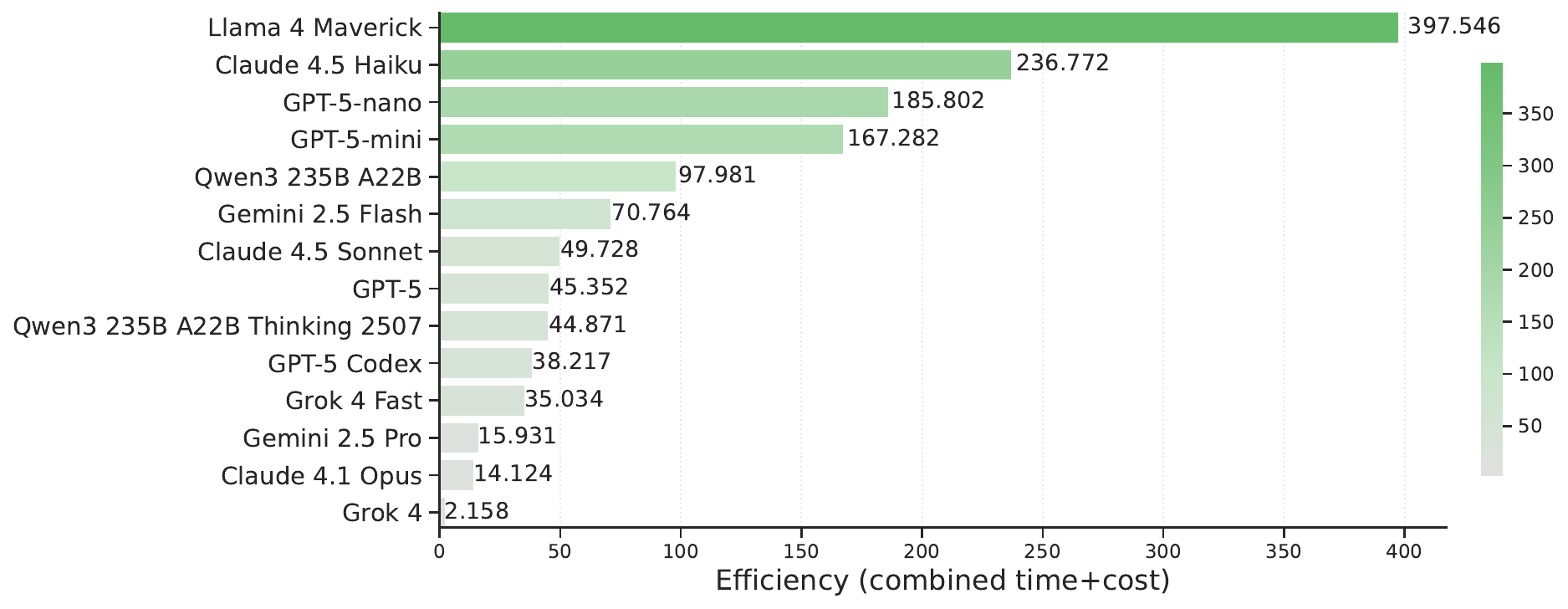}
        \caption{Comparison of performance–time–cost Efficiency Scores \textbf{(input modality: Image-only, prompt language: English)}}
        \label{fig:efficiency_time_cost_en_i}
    \end{subfigure}
    \caption{Comparison of performance–time–cost Efficiency Scores across models \textbf{(input modality: Image-only, prompt language: English)}}
    \label{fig:efficiency_en_i}
\end{figure}

This appendix presents the full set of Efficiency Scores by model, extending the analysis in Section~\ref{sec:leaderboard} to all input–language combinations (Text / Image / Text+Figure $\times$ Korean / English).  
Figure~\ref{fig:efficiency_ko_ti} compares Efficiency Scores for the (Text+Figure, Korean) condition;  
Figure~\ref{fig:efficiency_ko_i} for the (Image-only, Korean) condition;  
Figure~\ref{fig:efficiency_en_t} for the (Text-only, English) condition;  
Figure~\ref{fig:efficiency_en_ti} for the (Text+Figure, English) condition; and  
Figure~\ref{fig:efficiency_en_i} for the (Image-only, English) condition.  
Models that do not support image input are excluded from the Image-only and Text+Figure conditions.  
The corresponding plots for the (Text-only, Korean) condition are provided in Figure~\ref{fig:efficiency} in the main text.

\FloatBarrier

\section{Full Results by Model Size: Performance and Latency}
\label{sec:appendix_size_perf}

\begin{figure}
    \centering
    \includegraphics[width=\textwidth]{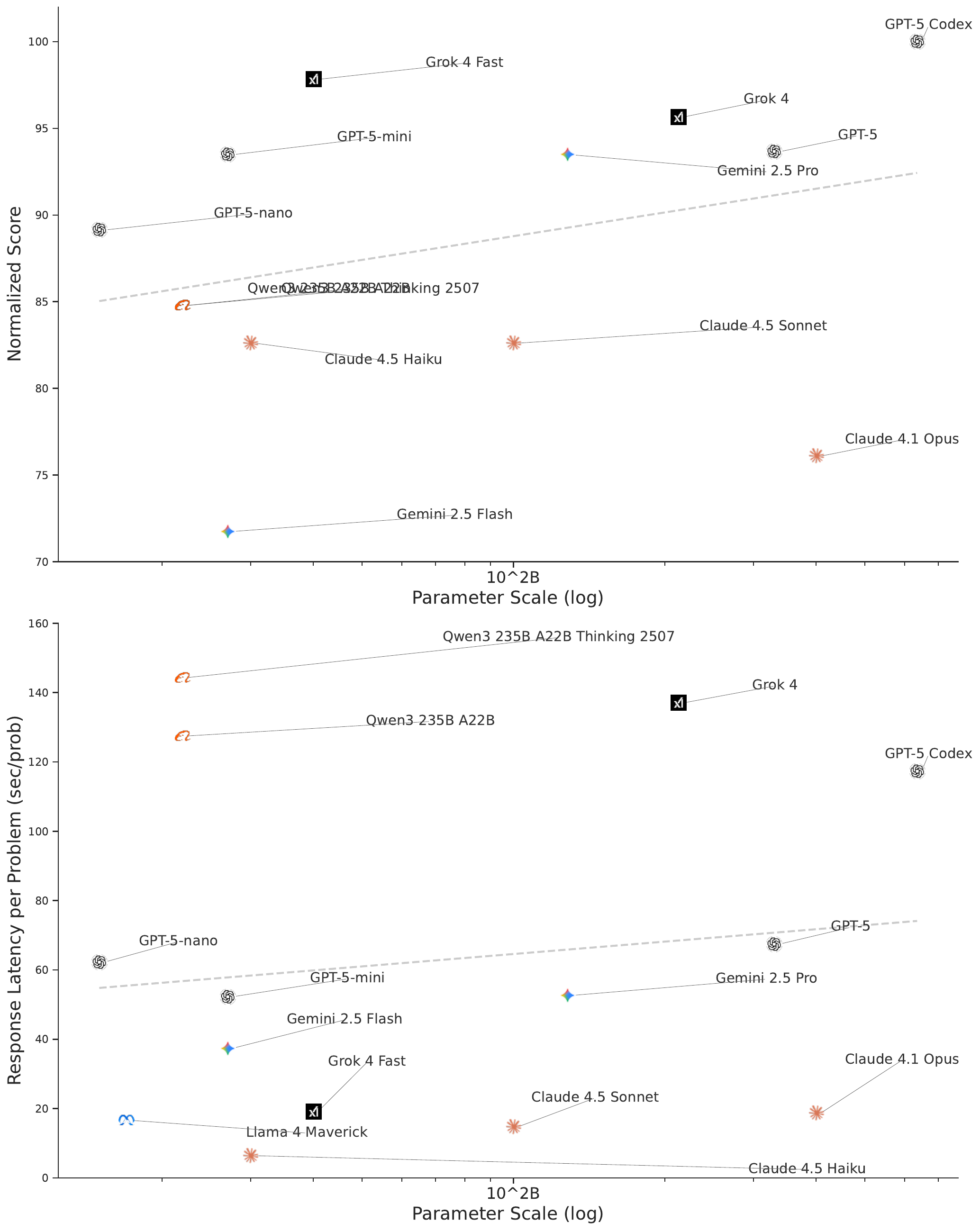}
    \caption{Performance (Normalized Score) and average time per problem by model parameter size \textbf{(input modality: Text+Figure, prompt language: Korean)}. The Llama 4 Maverick model is omitted because its Normalized Score (28.3) is substantially lower than other models.}
    \label{fig:parameter_scatter_ko_ti}
\end{figure}

\begin{figure}
    \centering
    \includegraphics[width=\textwidth]{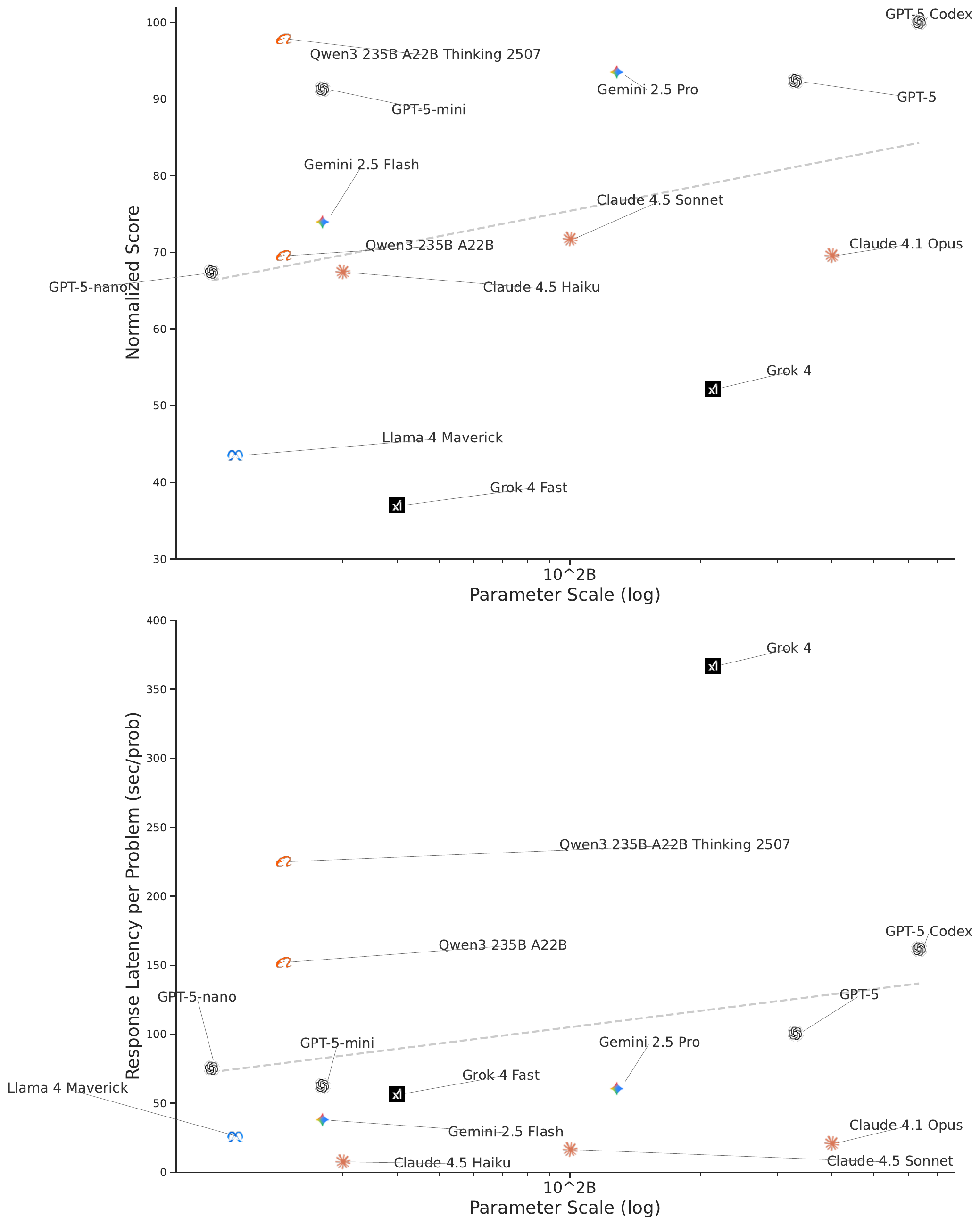}
    \caption{Performance (Normalized Score) and average time per problem by model parameter size \textbf{(input modality: Image-only, prompt language: Korean)}}
    \label{fig:parameter_scatter_ko_i}
\end{figure}

\begin{figure}
    \centering
    \includegraphics[width=\textwidth]{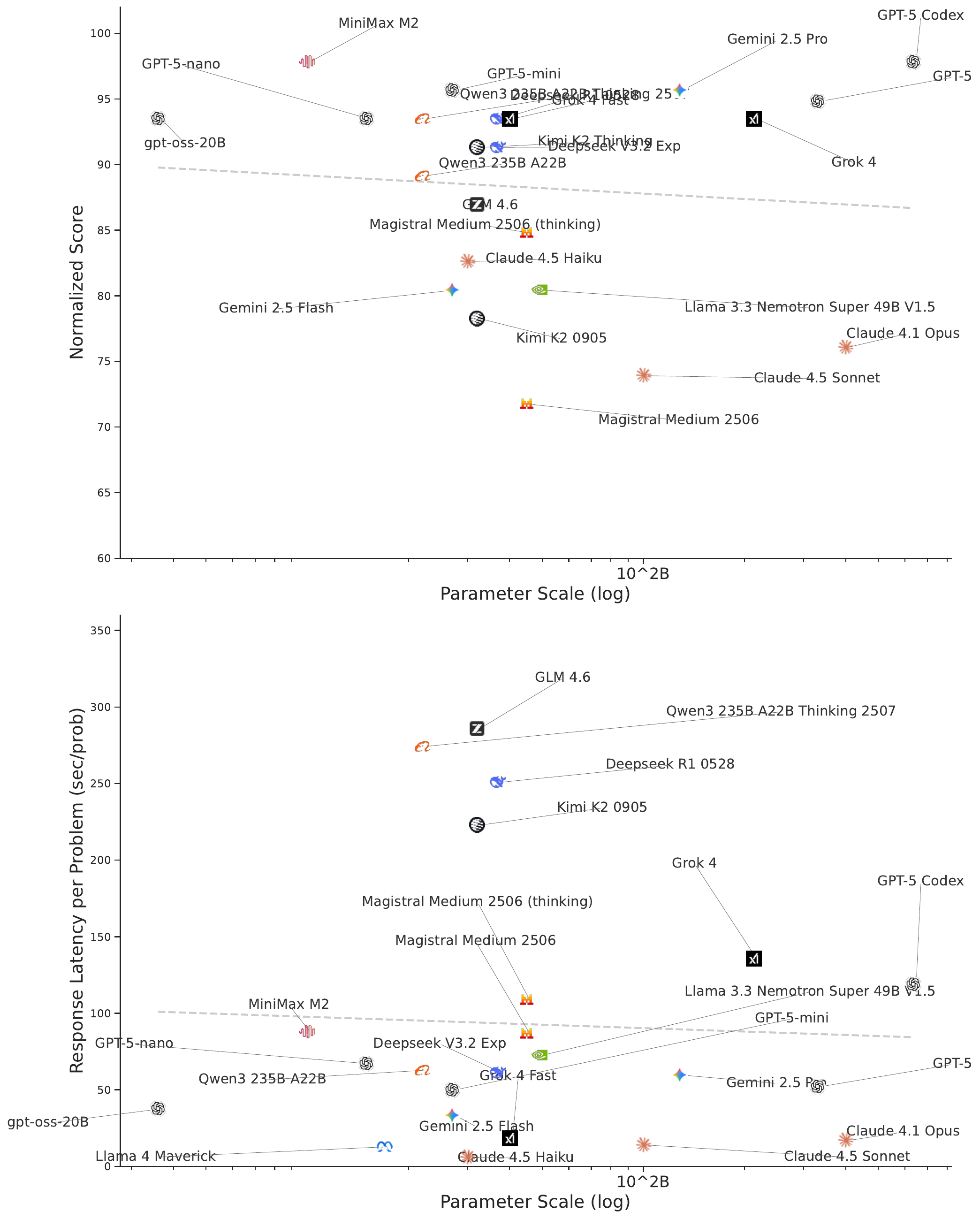}
    \caption{Performance (Normalized Score) and average time per problem by model parameter size \textbf{(input modality: Text-only, prompt language: English)}. The Llama 4 Maverick model is omitted because its Normalized Score (23.9) is substantially lower than other models, and the Kimi K2 Thinking model is omitted because its average time per problem (570 sec/prob) is much higher than the others.}
    \label{fig:parameter_scatter_en_t}
\end{figure}

\begin{figure}
    \centering
    \includegraphics[width=\textwidth]{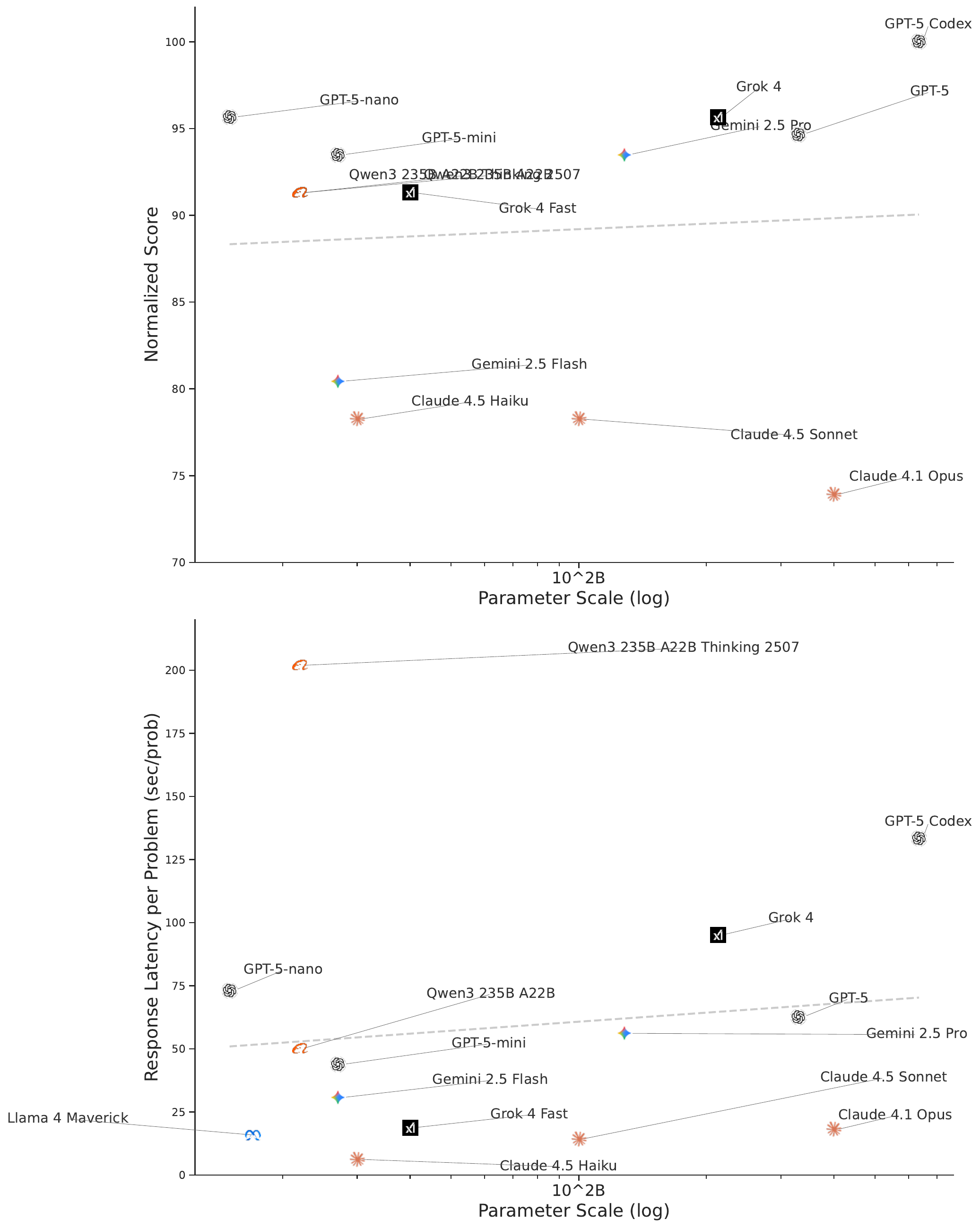}
    \caption{Performance (Normalized Score) and average time per problem by model parameter size \textbf{(input modality: Text+Figure, prompt language: English)}. The Llama 4 Maverick model is omitted because its Normalized Score (32.6) is substantially lower than other models.}
    \label{fig:parameter_scatter_en_ti}
\end{figure}

\begin{figure}
    \centering
    \includegraphics[width=\textwidth]{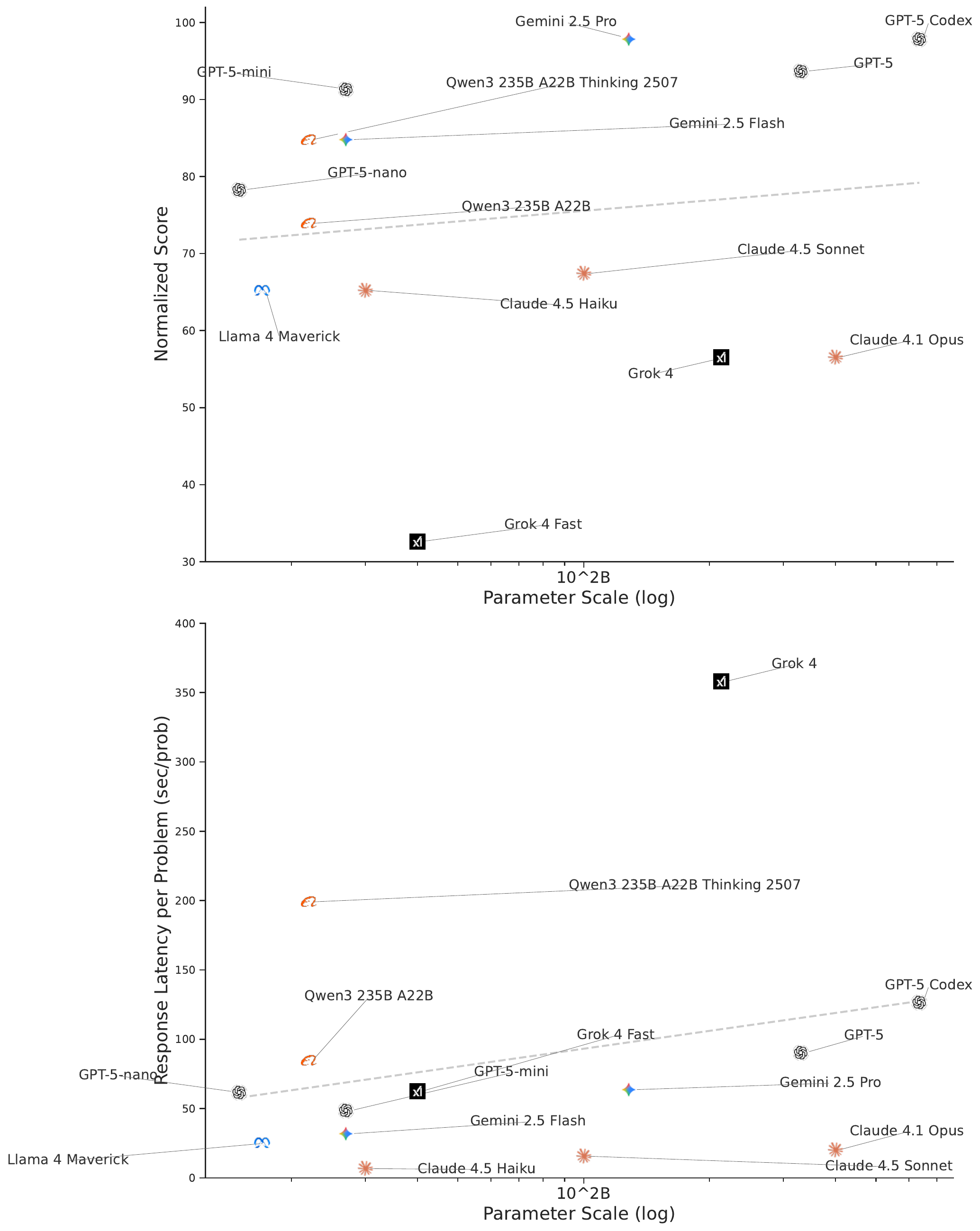}
    \caption{Performance (Normalized Score) and average time per problem by model parameter size \textbf{(input modality: Image-only, prompt language: English)}}
    \label{fig:parameter_scatter_en_i}
\end{figure}

This appendix reports the full results on model size, extending the analysis in Section~\ref{sec:parameter_size} to all input–language combinations (Text / Image / Text+Figure $\times$ Korean / English).  
Figure~\ref{fig:parameter_scatter_ko_ti} compares performance (Normalized Score) and average time per problem for the (Text+Figure, Korean) condition;  
Figure~\ref{fig:parameter_scatter_ko_i} for the (Image-only, Korean) condition;  
Figure~\ref{fig:parameter_scatter_en_t} for the (Text-only, English) condition;  
Figure~\ref{fig:parameter_scatter_en_ti} for the (Text+Figure, English) condition; and  
Figure~\ref{fig:parameter_scatter_en_i} for the (Image-only, English) condition.  
Models that do not support image input are excluded from the Image-only and Text+Figure conditions.  
The corresponding plots for the (Text-only, Korean) condition are provided in Figure~\ref{fig:parameter_scatter} in the main text.

\FloatBarrier

\section{Full Results by Problem Area}
\label{sec:appendix_area_ttributes}

This appendix provides the complete results by problem area (Common Mathematics, Probability and Statistics, Calculus, Geometry), extending the analysis in Section~\ref{sec:paper_area}.  
For every input–language combination (Text / Image / Text+Figure $\times$ Korean / English), we compute the normalized score for each problem area.  
These results allow a finer examination of whether a model’s strengths and weaknesses appear consistently across modalities, or whether the patterns shift depending on input format.

\begin{figure}[!th]
    \centering
    \includegraphics[width=\linewidth]{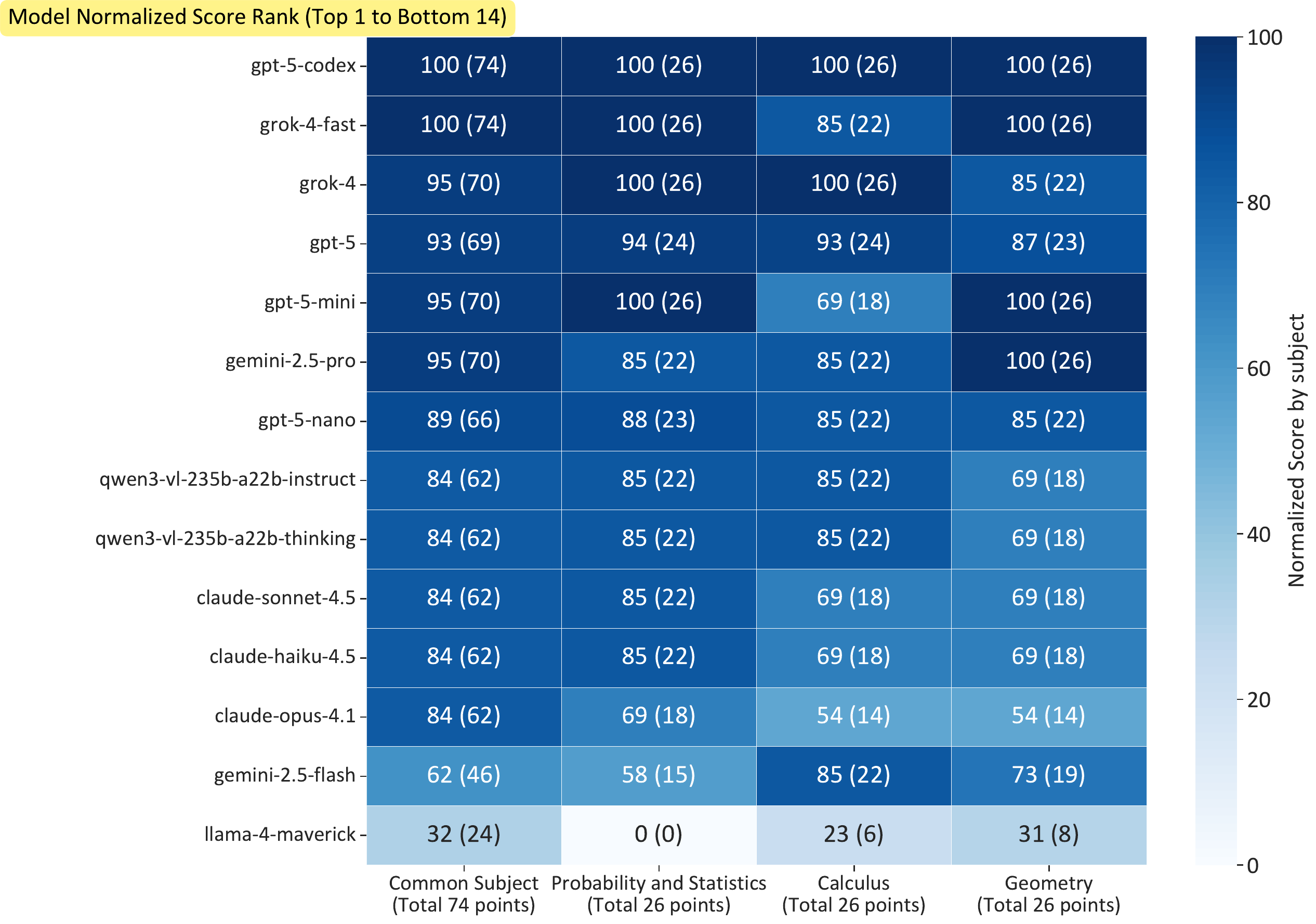}
    \caption{
    Comparison of LLM performance by problem area \textbf{(input modality: Text+Figure, prompt language: Korean)}.  
    Normalized scores represent the ratio of points earned to the maximum points for each area.  
    Both the normalized score and raw points are shown.  
    Models are ordered by overall Normalized Score in descending order.
    }
    \label{fig:RQ2_problem_area_scre_ko_ti}
\end{figure}

\begin{figure}[!th]
    \centering
    \includegraphics[width=\linewidth]{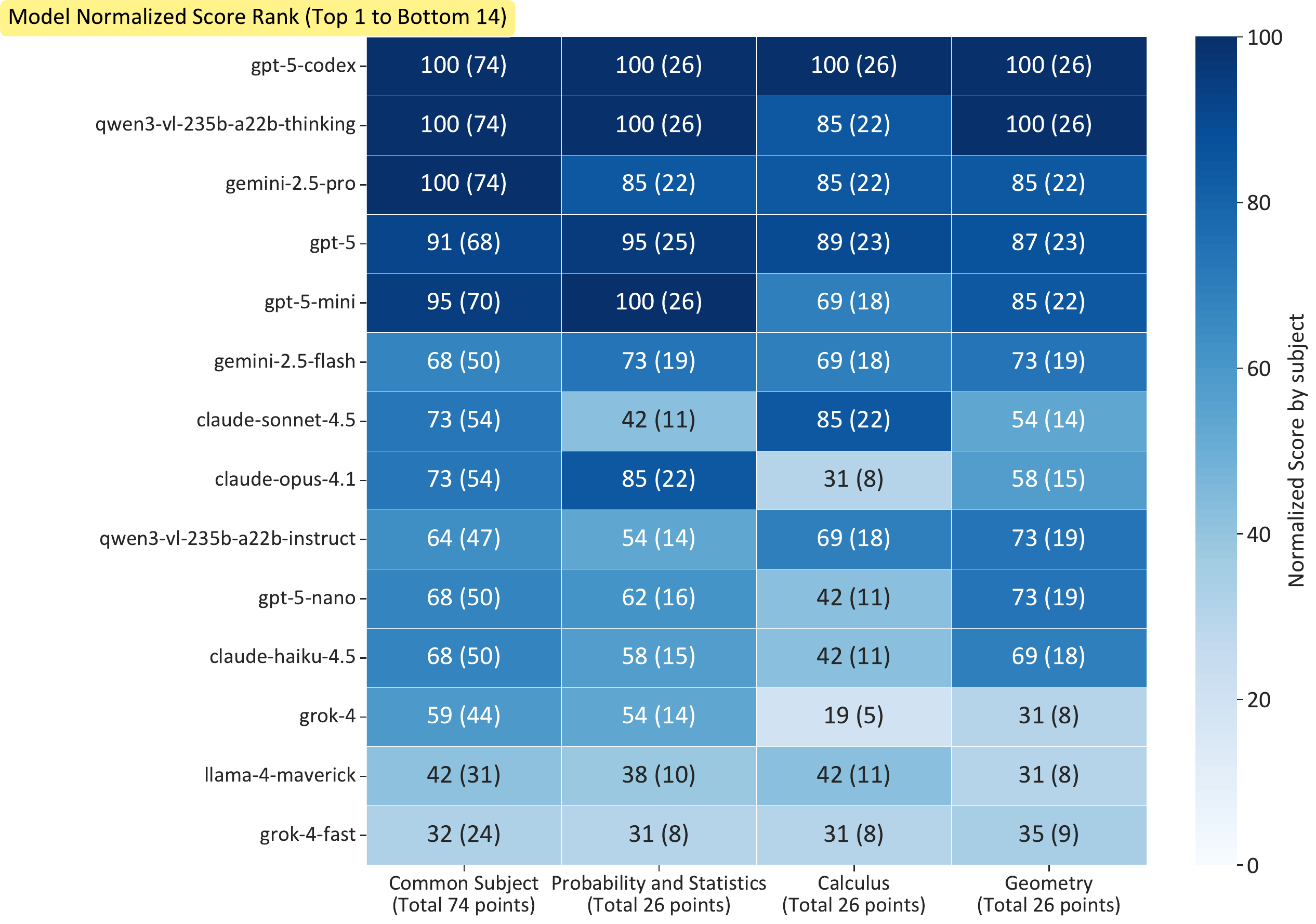}
    \caption{
    Comparison of LLM performance by problem area \textbf{(input modality: Image-only, prompt language: Korean)}.  
    Normalized scores represent the ratio of points earned to the maximum points for each area, with raw points displayed.  
    Models appear in descending order of overall Normalized Score.
    }
    \label{fig:RQ2_problem_area_scre_ko_i}
\end{figure}

\begin{figure}[!th]
    \centering
    \includegraphics[width=\linewidth]{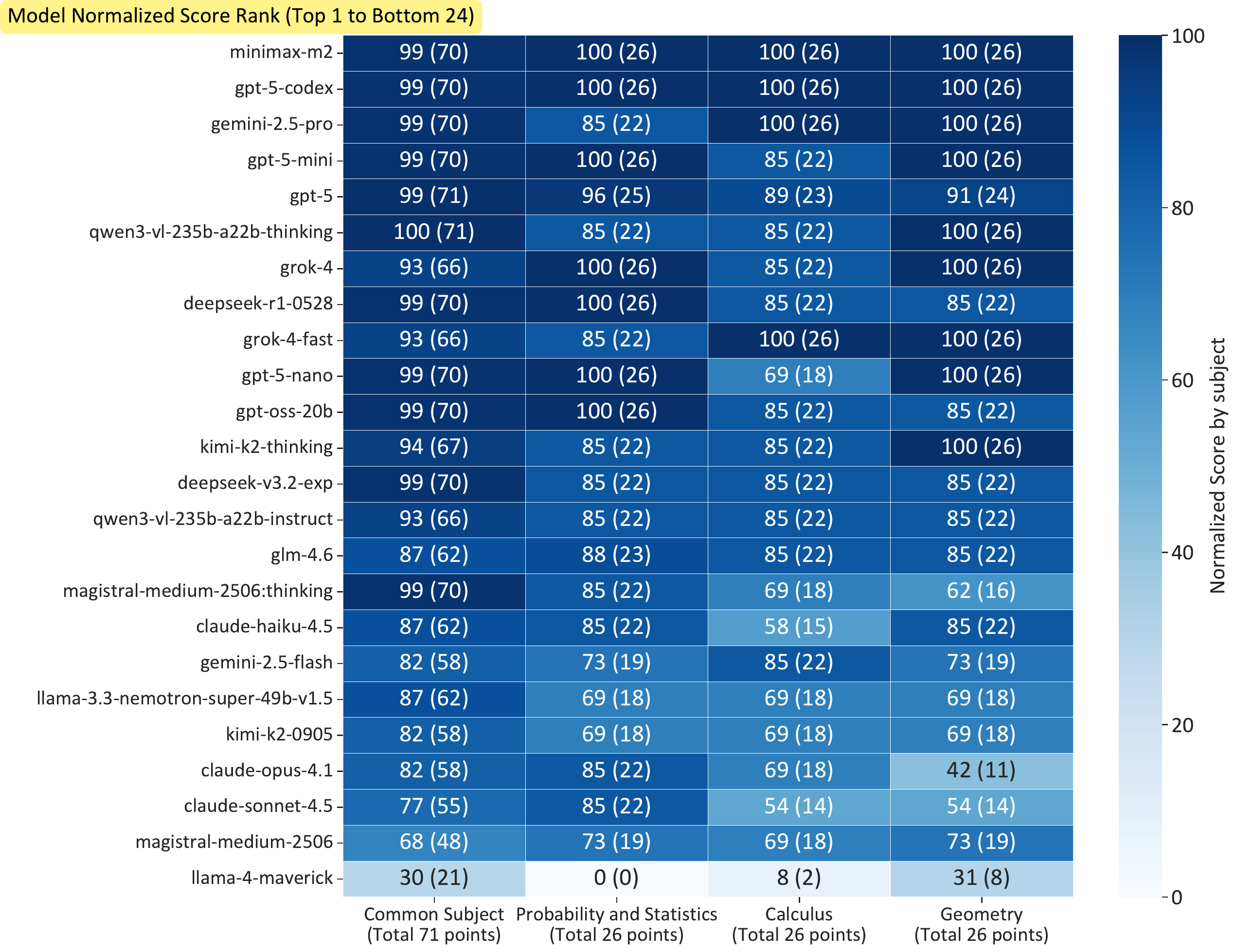}
    \caption{
    Comparison of LLM performance by problem area \textbf{(input modality: Text-only, prompt language: English)}.  
    Normalized scores are computed using area-specific maximum scores; raw points are shown together.  
    Models are sorted by overall Normalized Score.
    }
    \label{fig:RQ2_problem_area_scre_en_t}
\end{figure}

\begin{figure}[!th]
    \centering
    \includegraphics[width=\linewidth]{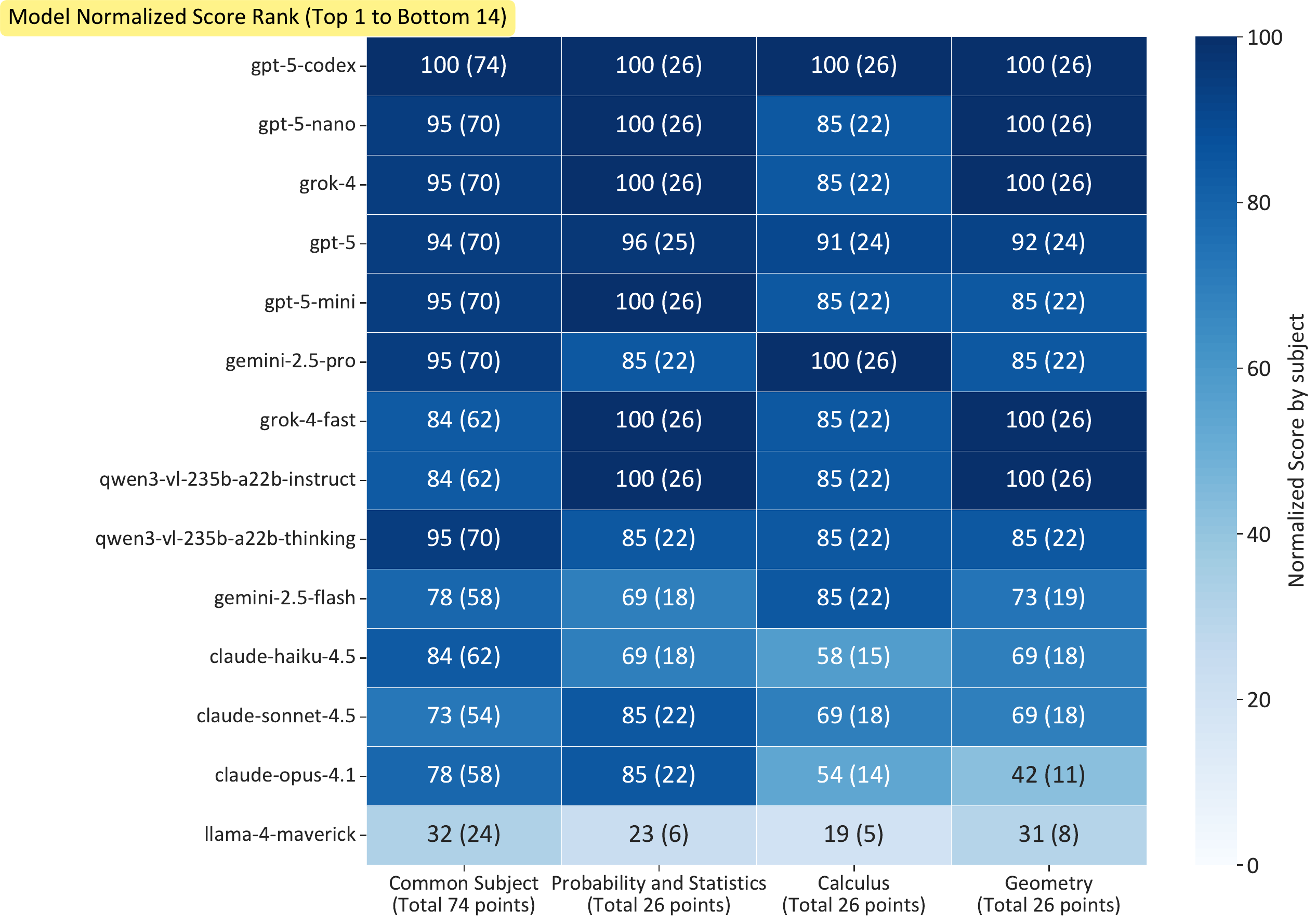}
    \caption{
    Comparison of LLM performance by problem area \textbf{(input modality: Text+Figure, prompt language: English)}.  
    Normalized scores represent the ratio of earned to maximum points, with raw points included.  
    Models are ordered by overall Normalized Score.
    }
    \label{fig:RQ2_problem_area_scre_en_ti}
\end{figure}

\begin{figure}[!th]
    \centering
    \includegraphics[width=\linewidth]{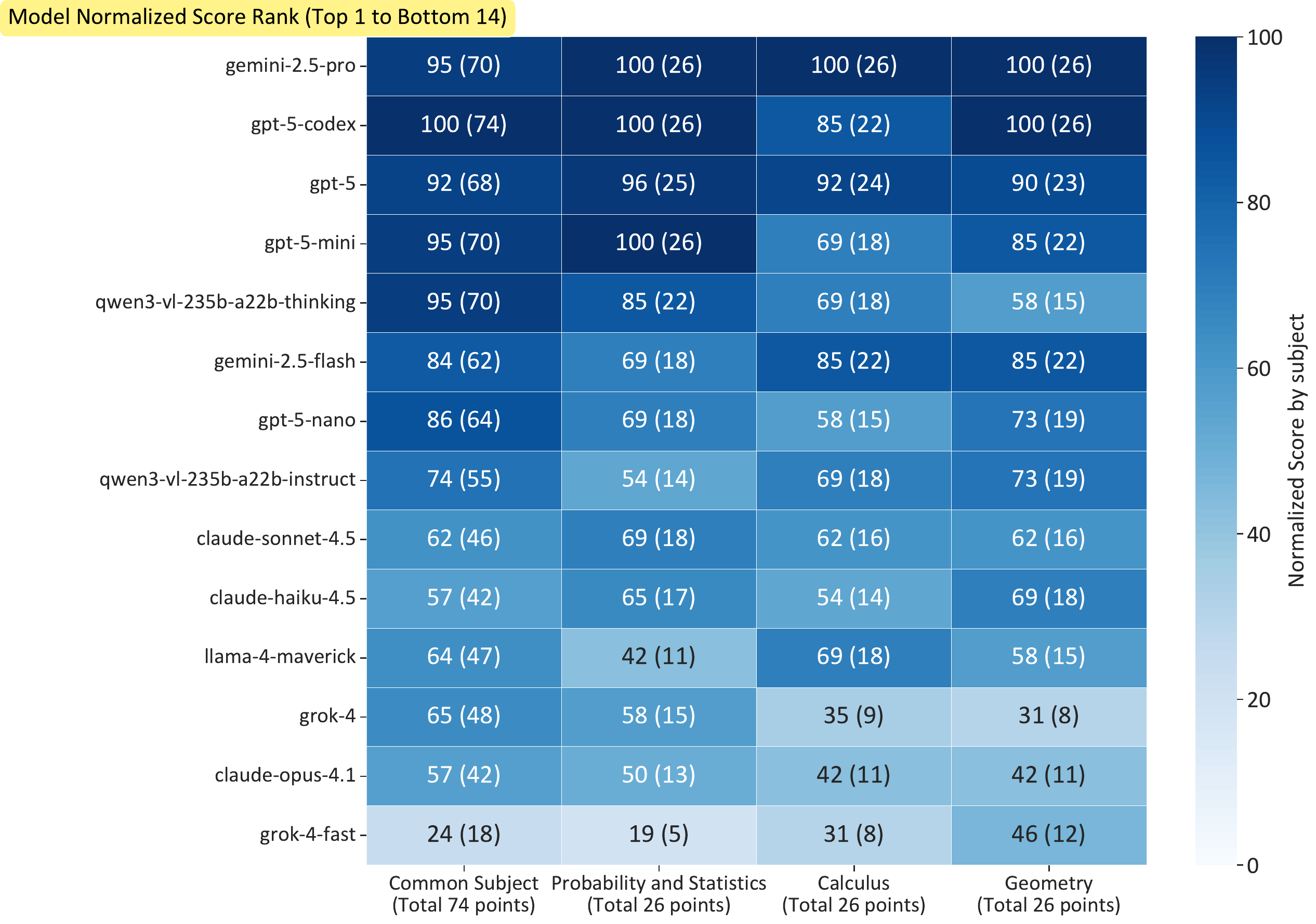}
    \caption{
    Comparison of LLM performance by problem area \textbf{(input modality: Image-only, prompt language: English)}.  
    Normalized scores and raw points are shown.  
    Models appear in descending order of overall Normalized Score.
    }
    \label{fig:RQ2_problem_area_scre_en_i}
\end{figure}

Figure~\ref{fig:RQ2_problem_area_scre_ko_ti} presents area-wise results for the (Text+Figure, Korean) condition;  
Figure~\ref{fig:RQ2_problem_area_scre_ko_i} for the (Image-only, Korean) condition;  
Figure~\ref{fig:RQ2_problem_area_scre_en_t} for the (Text-only, English) condition;  
Figure~\ref{fig:RQ2_problem_area_scre_en_ti} for the (Text+Figure, English) condition; and  
Figure~\ref{fig:RQ2_problem_area_scre_en_i} for the (Image-only, English) condition.  
Models without image-input capability are excluded from Image-only and Text+Figure evaluations.  
Results for the (Text-only, Korean) condition appear in Figure~\ref{fig:RQ2_problem_area_scre} in the main text.

\FloatBarrier

\section{Full Results by Problem Attributes}
\label{sec:appendix_item_attributes}

This appendix extends the analysis in Section~\ref{sec:paper_attribute} by reporting full results across problem attributes (problem format and point value) and all input–language combinations (Text / Image / Text+Figure $\times$ Korean / English).  

\begin{figure}[!th]
    \centering
    % ---- Left: Score / Type ----
    \begin{subfigure}[t]{0.48\linewidth}
        \centering
        \includegraphics[width=\linewidth]{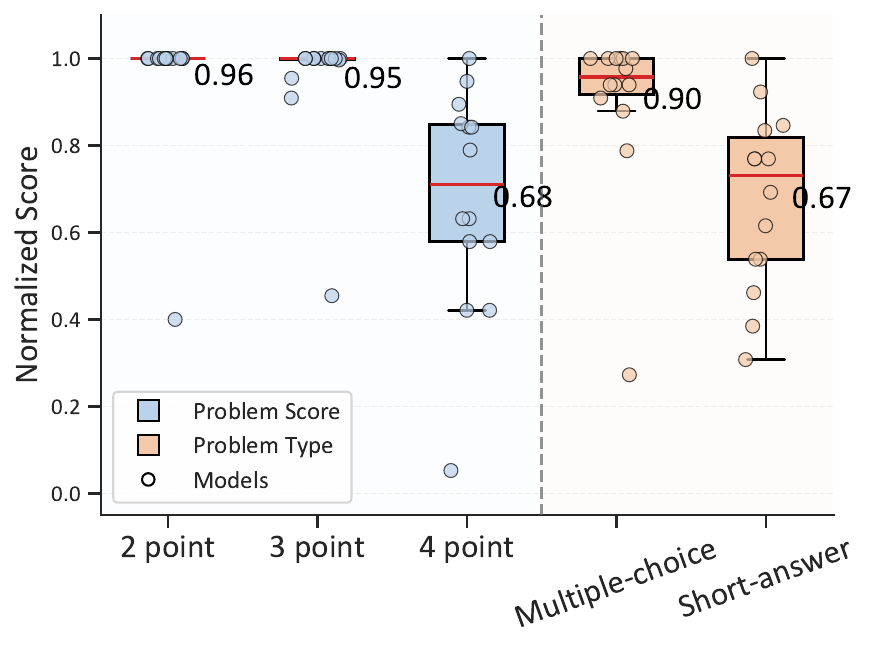}
        \caption{
        Distribution of LLM performance by point value (2/3/4 points) and problem format (multiple-choice / short-answer).
        }
        \label{fig:appendix_dist7}
    \end{subfigure}
    \hfill
    % ---- Right: Score × Type combination ----
    \begin{subfigure}[t]{0.48\linewidth}
        \centering
        \includegraphics[width=\linewidth]{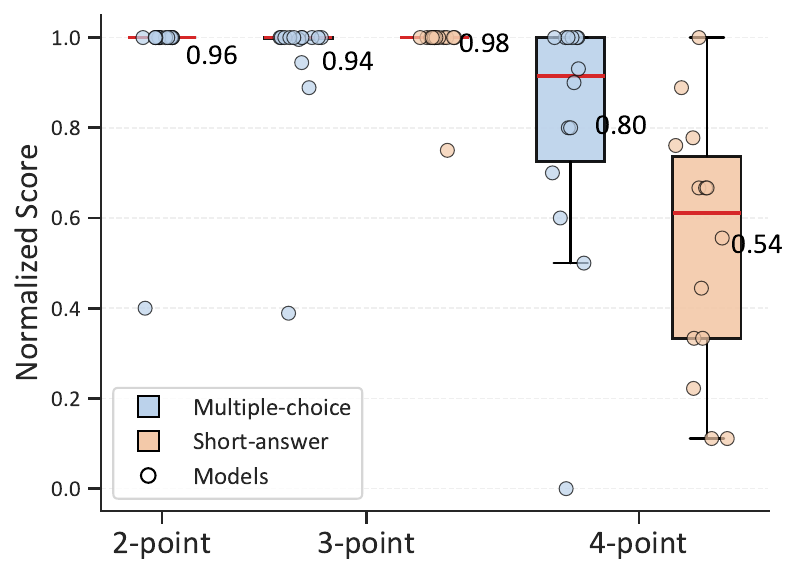}
        \caption{
        LLM performance by combined problem attributes (point value × problem format).  
        (The 2026 CSAT contains no 2-point multiple-choice problems.)
        }
        \label{fig:appendix_dist8}
    \end{subfigure}

    \caption{
    Comparison of LLM performance by problem attributes \textbf{(input modality: Text+Figure, prompt language: Korean)}.  
    Panel (a) examines point value and problem format independently, while  
    panel (b) analyzes performance by combined problem categories (point value × format).
    }
    \label{fig:appendix_dist_ko_ti}
\end{figure}

\begin{figure}[!th]
    \centering
    % ---- Left: Score / Type ----
    \begin{subfigure}[t]{0.48\linewidth}
        \centering
        \includegraphics[width=\linewidth]{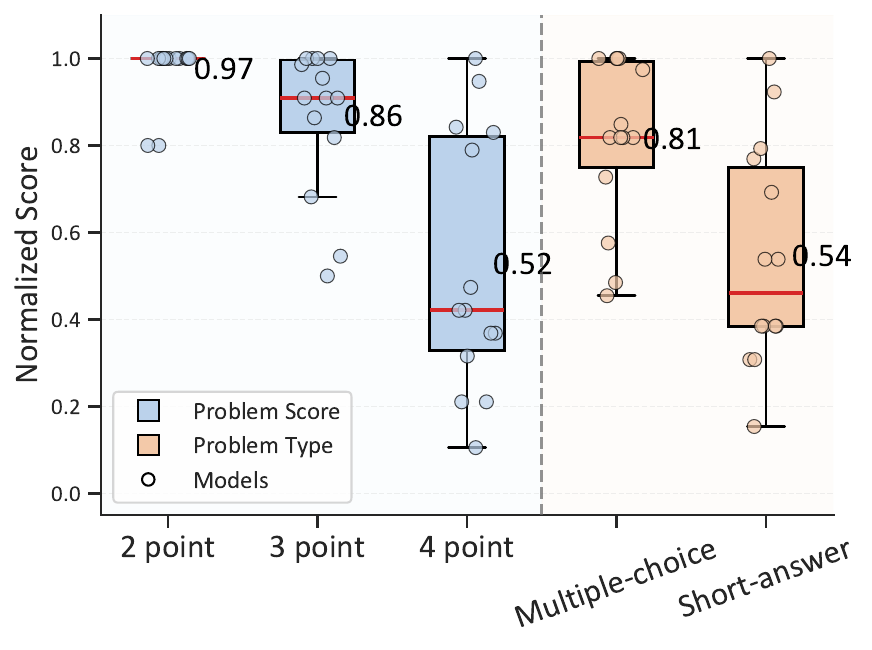}
        \caption{
        Distribution of LLM performance by point value and problem format.
        }
        \label{fig:appendix_dist3}
    \end{subfigure}
    \hfill
    % ---- Right: Score × Type combination ----
    \begin{subfigure}[t]{0.48\linewidth}
        \centering
        \includegraphics[width=\linewidth]{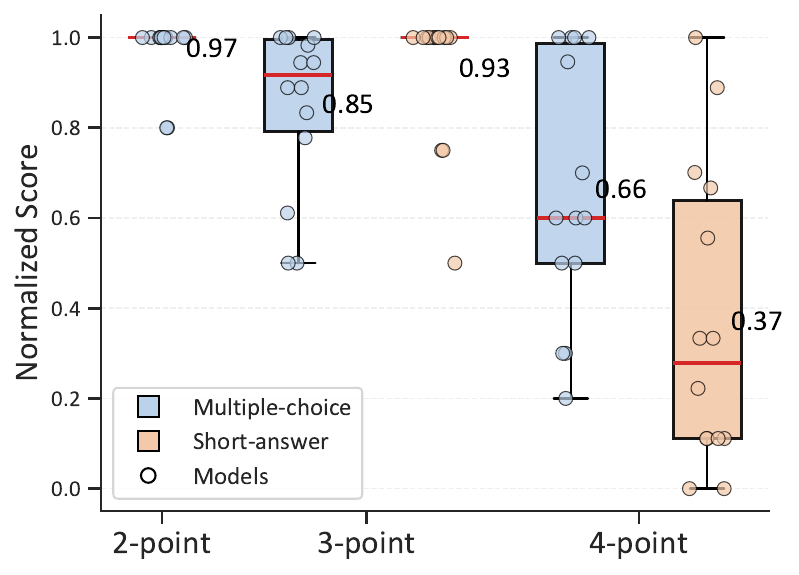}
        \caption{
        LLM performance by combined problem attributes (point value × format).  
        (No 2-point multiple-choice problems in the 2026 CSAT.)
        }
        \label{fig:appendix_dist4}
    \end{subfigure}

    \caption{
    Comparison of LLM performance by problem attributes \textbf{(input modality: Image-only, prompt language: Korean)}.  
    Panel (a) isolates the two attributes, while panel (b) evaluates combined problem categories.
    }
    \label{fig:appendix_dist_ko_i}
\end{figure}

\begin{figure}[!th]
    \centering
    % ---- Left: Score / Type ----
    \begin{subfigure}[t]{0.48\linewidth}
        \centering
        \includegraphics[width=\linewidth]{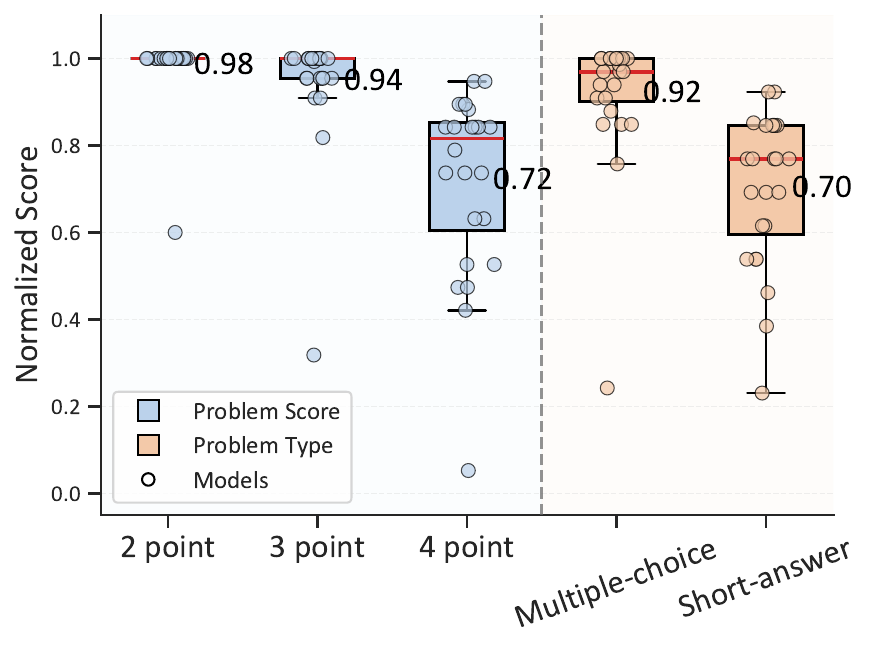}
        \caption{
        Distribution of LLM performance by point value and problem format.
        }
        \label{fig:appendix_dist}
    \end{subfigure}
    \hfill
    % ---- Right: Score × Type combination ----
    \begin{subfigure}[t]{0.48\linewidth}
        \centering
        \includegraphics[width=\linewidth]{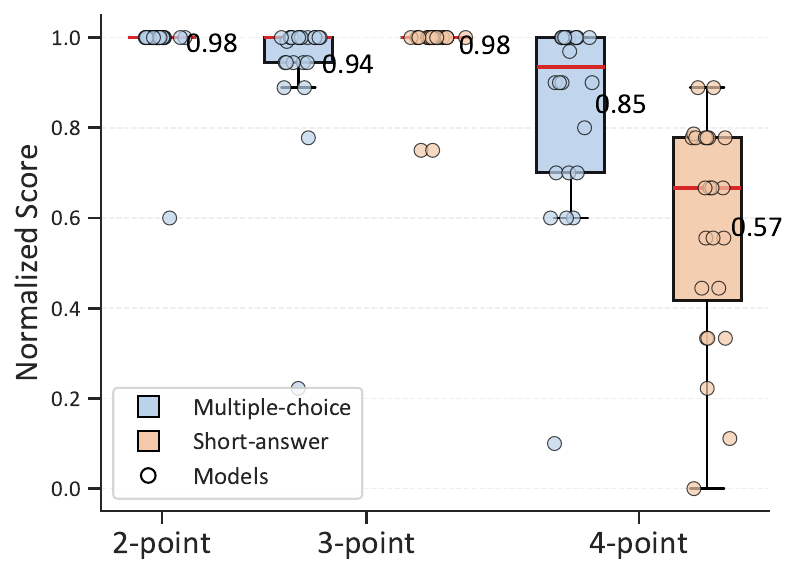}
        \caption{
        LLM performance by combined problem attributes (point value × format).  
        (No 2-point multiple-choice problems in the 2026 CSAT.)
        }
        \label{fig:appendix_dist2}
    \end{subfigure}

    \caption{
    Comparison of LLM performance by problem attributes \textbf{(input modality: Text-only, prompt language: English)}.  
    Panel (a) examines problem format and point value separately, whereas  
    panel (b) analyzes their combined categories.
    }
    \label{fig:appendix_dist_en_t}
\end{figure}

\begin{figure}[!th]
    \centering
    % ---- Left: Score / Type ----
    \begin{subfigure}[t]{0.48\linewidth}
        \centering
        \includegraphics[width=\linewidth]{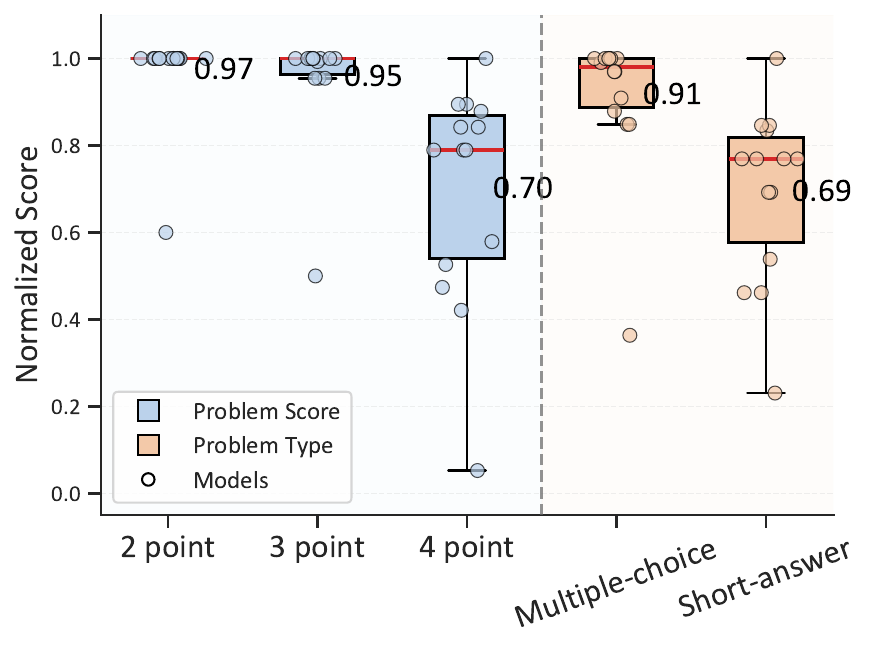}
        \caption{
        Distribution of LLM performance by point value and problem format.
        }
        \label{fig:appendix_dist9}
    \end{subfigure}
    \hfill
    % ---- Right: Score × Type combination ----
    \begin{subfigure}[t]{0.48\linewidth}
        \centering
        \includegraphics[width=\linewidth]{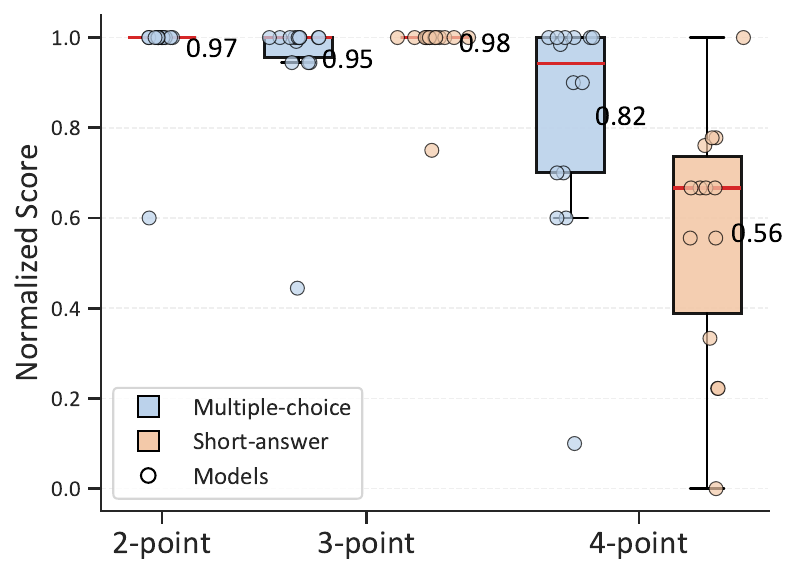}
        \caption{
        LLM performance by combined problem attributes (point value × format).  
        (No 2-point multiple-choice problems in the 2026 CSAT.)
        }
        \label{fig:appendix_dist10}
    \end{subfigure}

    \caption{
    Comparison of LLM performance by problem attributes \textbf{(input modality: Text+Figure, prompt language: English)}.  
    Panel (a) analyzes attributes independently, while panel (b) examines combined problem categories.
    }
    \label{fig:appendix_dist_en_ti}
\end{figure}

\begin{figure}[!th]
    \centering
    % ---- Left: Score / Type ----
    \begin{subfigure}[t]{0.48\linewidth}
        \centering
        \includegraphics[width=\linewidth]{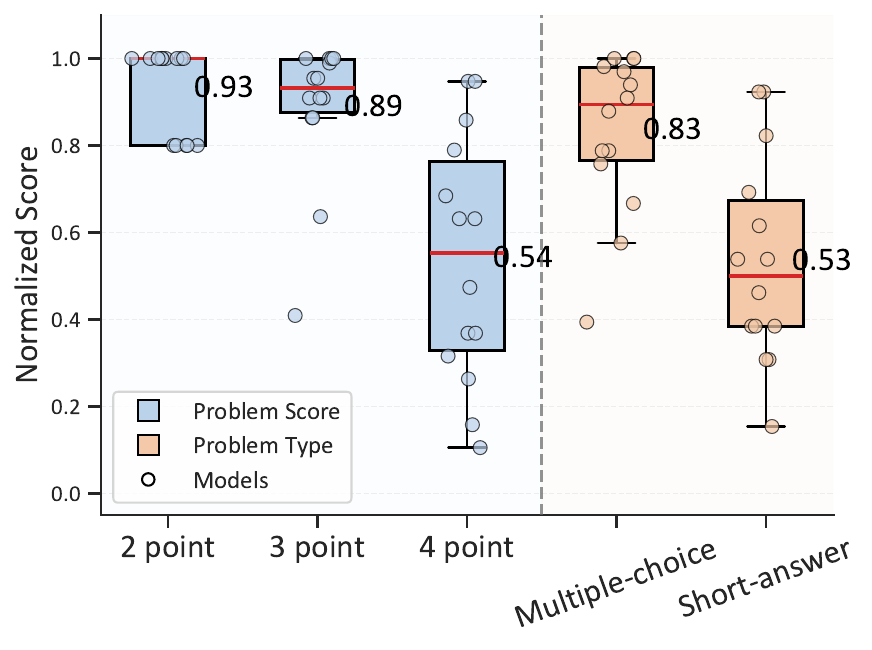}
        \caption{
        Distribution of LLM performance by point value and problem format.
        }
        \label{fig:appendix_dist5}
    \end{subfigure}
    \hfill
    % ---- Right: Score × Type combination ----
    \begin{subfigure}[t]{0.48\linewidth}
        \centering
        \includegraphics[width=\linewidth]{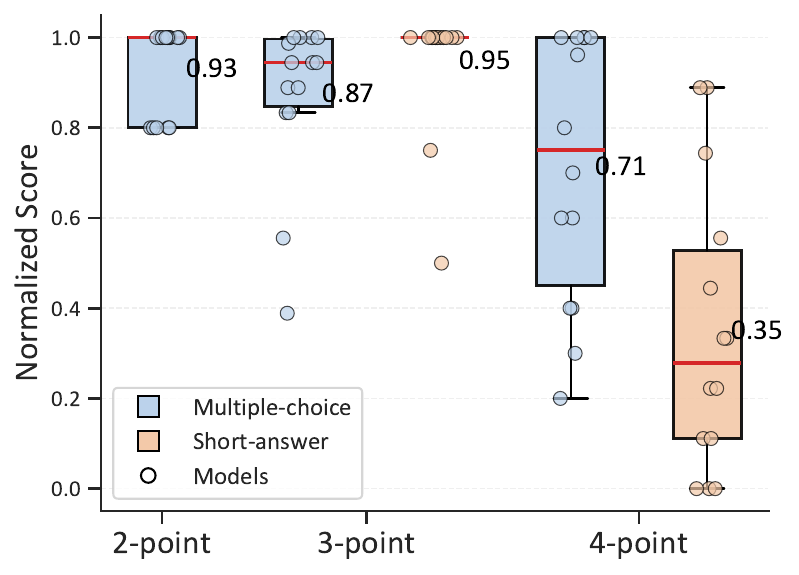}
        \caption{
        LLM performance by combined problem attributes (point value × format).  
        (No 2-point multiple-choice problems in the 2026 CSAT.)
        }
        \label{fig:appendix_dist6}
    \end{subfigure}

    \caption{
    Comparison of LLM performance by problem attributes \textbf{(input modality: Image-only, prompt language: English)}.  
    Panel (a) shows separate attribute-wise performance, while  
    panel (b) presents performance for combined problem categories.
    }
    \label{fig:appendix_dist_en_i}
\end{figure}

Figure~\ref{fig:appendix_dist_ko_ti} presents the results for (Text+Figure, Korean);  
Figure~\ref{fig:appendix_dist_ko_i} for (Image-only, Korean);  
Figure~\ref{fig:appendix_dist_en_t} for (Text-only, English);  
Figure~\ref{fig:appendix_dist_en_ti} for (Text+Figure, English); and  
Figure~\ref{fig:appendix_dist_en_i} for (Image-only, English).  
The corresponding (Text-only, Korean) figure appears in Figure~\ref{fig:RQ2_combined} in the main text.

\end{document}